\documentclass{article}

\usepackage{microtype}
\usepackage{graphicx}
\usepackage{subfig}
\usepackage{booktabs} 
\usepackage{multirow}
\usepackage{wrapfig}
\usepackage{tikz}
\usetikzlibrary{arrows.meta}
\usetikzlibrary{positioning, shapes.geometric, arrows}
\usetikzlibrary{shapes, arrows}

\usepackage{hyperref}

\usepackage[accepted]{icml2024}

\usepackage{amsmath}
\usepackage{amssymb}
\usepackage{mathtools}
\usepackage{amsthm}

\usepackage{amsmath,amsfonts,bm, physics}

\def\eqref#1{equation~\ref{#1}}

\def\1{\bm{1}}

\newcommand{\ie}{{\em i.e., }}

\newcommand{\ba}{\boldsymbol{a}}

\newcommand{\bu}{\boldsymbol{u}}
\newcommand{\bv}{\boldsymbol{v}}
\newcommand{\bw}{\boldsymbol{w}}
\newcommand{\bx}{\boldsymbol{x}}
\newcommand{\by}{\boldsymbol{y}}
\newcommand{\bz}{\boldsymbol{z}}

\newcommand{\BPhi}{\boldsymbol{\Phi}}

\newcommand{\btheta}{\boldsymbol{\theta}}

\newcommand{\bzero}{\boldsymbol{0}}

\newcommand{\reals}{{\mathbb{R}}}

\newcommand{\lag}{\langle}
\newcommand{\rag}{\rangle}

\newcommand{\lp}{\left(} 
\newcommand{\rp}{\right)} 
\newcommand{\ls}{\left[} 
\newcommand{\rs}{\right]} 
\newcommand{\lc}{\left\{} 
\newcommand{\rc}{\right\}} 

\DeclareMathOperator{\E}{\mathbb{E}}

\DeclareMathOperator{\argmin}{argmin}

\theoremstyle{plain}
\newtheorem{theorem}{Theorem}[section]

\newtheorem{lemma}[theorem]{Lemma}

\theoremstyle{definition}
\newtheorem{definition}[theorem]{Definition}
\newtheorem{assumption}[theorem]{Assumption}
\theoremstyle{remark}
\newtheorem{remark}[theorem]{Remark}
\newtheorem{claim}[theorem]{Claim}

\usepackage[capitalize,noabbrev]{cleveref}

\usepackage[]{todonotes}

\usepackage{xcolor}
\definecolor{darkpink}{rgb}{0.91, 0.33, 0.5}
\definecolor{ycolor}{named}{darkpink}
\definecolor{scolor}{rgb}{0.93, 0.53, 0.18}
\definecolor{yscolor}{rgb}{0.92, 0.43, 0.34}
\definecolor{ecolor}{rgb}{0.4, 0.6, 0.8}

\tikzset{
    neuron/.style={circle, draw, minimum size=0.1cm},
    input/.style={neuron, fill=green!50},
    output/.style={neuron, fill=red!50},
    hidden/.style={neuron, fill=scolor!50},
    basefeature/.style={neuron, fill=orange!50},
    eneuron/.style={neuron, fill=ecolor!50},
    yneuron/.style={neuron, fill=ycolor!50},
    annotation/.style={text width=4em, text centered}
}

\icmltitlerunning{Bridging Domains with Approximately Shared Features}

\begin{document}

\onecolumn
\icmltitle{{Bridging Domains with Approximately Shared Features}}

\icmlsetsymbol{equal}{*}

\begin{icmlauthorlist}
\icmlauthor{Ziliang Samuel Zhong}{equal,nyush,cds}
\icmlauthor{Xiang Pan}{equal,cds}
\icmlauthor{Qi Lei}{cds}
\end{icmlauthorlist}

\icmlaffiliation{nyush}{Shanghai Frontiers Science Center of Artificial Intelligence and Deep Learning, New York University Shanghai}
\icmlaffiliation{cds}{Center for Data Science, New York University}

 \icmlcorrespondingauthor{Ziliang Samuel Zhong}{\texttt{zz1706@nyu.edu}}
 \icmlcorrespondingauthor{Xiang Pan}{\texttt{xp2030@nyu.edu}}
 \icmlcorrespondingauthor{Qi Lei}{\texttt{ql518@nyu.edu}}

\icmlkeywords{Machine Learning, ICML}

\vskip 0.3in

\printAffiliationsAndNotice{\icmlEqualContribution} 

\begin{abstract}
Multi-source domain adaptation aims to reduce performance degradation when applying machine learning models to unseen domains. A fundamental challenge is devising the optimal strategy for feature selection. Existing literature is somewhat paradoxical: some advocate for learning invariant features from source domains, while others favor more diverse features. To address the challenge, we propose a statistical framework that distinguishes the utilities of features based on the variance of their correlation to label $y$ across domains. Under our framework, we design and analyze a learning procedure consisting of learning approximately shared feature representation from source tasks and fine-tuning it on the target task. Our theoretical analysis necessitates the importance of learning approximately shared features instead of only the strictly invariant features and yields an improved population risk compared to previous results on both source and target tasks, thus partly resolving the paradox mentioned above. Inspired by our theory, we proposed a more practical way to isolate the content (invariant+approximately shared) from environmental features and further consolidate our theoretical findings.
\end{abstract}

\section{Introduction}
Machine learning models suffer from performance degradation when adapting to new domains~\citep{koh2021wilds}.  To mitigate this issue, (multi-source) domain adaptation has become a pivotal strategy, enabling the rapid transfer of knowledge from source domains to the target domain \citep{quinonero2008dataset,saenko2010adapting}. Despite its empirical success~\citep{wang2018deep,daume2007frustratingly}, a systematic theoretical understanding is still lacking. 

A fundamental problem is devising the optimal strategy for feature selection in source domains. Most previous theoretical works pivot around specific domain adaptation algorithms and select features based on causality, which requires strong assumptions on the statistical relationship between source and target distributions. These assumptions include but are not limited to $(1)$ the source domain covering the target domain~\citep{shimodaira2000improving,heckman1979sample,cortes2010learning,zadrozny2004learning}; $(2)$ the source and target domain having overlapping subgroups~\citep{wei2020theoretical,cai2021theory}; $(3)$ content features being exactly shared across domains~\citep{arjovsky2019invariant,ben2010theory,ahuja2020invariant}. 
Besides the limitation of learning causal features, existing literature also has paradoxical opinions on selecting non-causal features. For instance, some works suggest that we should learn invariant features,~\citep{ahuja2020invariant,ben2010theory,ajakan2014domain} while others suggest that we should learn diverse (even possibly spurious) features~\citep{shen2022data,wortsman2022model} to get a more distributionally robust model. A more detailed discussion is deferred to the related work section. 

\begin{figure}[!hpt]
    \centering
    \scalebox{0.5}{


\begin{tikzpicture}[every node/.style={font=\Large}]

\draw[fill=ecolor!20] (0,-0.5) ellipse (5cm and 3cm) node[below=1.5cm, align=center] {Environmental/Spurious Features};

\draw[fill=scolor!30] (0,0) ellipse (4.2cm and 2cm) node[right=-1cm,align=center] {Approximately Shared\\ Features};

\draw[fill=ycolor!30] (-2,0) circle (1.1 cm) node[align=center] {Invariant \\ Features};

\draw[->,double,line width=1pt, -{Stealth[scale=1.5]}] (5,0) -- (7,0);
\draw[fill=ycolor!30] (9,1) ellipse (2cm and 2cm) node[right=-1cm,align=center, yshift=0.5cm
] {Y-Specific};

\draw[fill=ecolor!30] (9,-1) ellipse (2cm and 2cm) node[right=-1cm,align=center, yshift=-0.5cm] {E-Specific};

\clip (9,1) ellipse (2cm and 2cm);
\clip (9,-1) ellipse (2cm and 2cm);
\fill[scolor!50] (9,0) ellipse (3cm and 2cm);

\node at (9,0) {Shared Features};

\end{tikzpicture}
    }
    \caption{Diagram for different feature types, mathematically defined in~\autoref{eqn: data generation component}. Our work indicates that in addition to invariant features, we should utilize approximately shared features to fully transfer the knowledge from the source to the target domain. The practical way to learn the \textbf{approximately shared features} is learning features are correlated to both $y$ and the environment $e$, which is the \textbf{$y$-$e$ shared features}}
    \label{fig:diagram}
    \vspace{-0.2cm}
\end{figure}
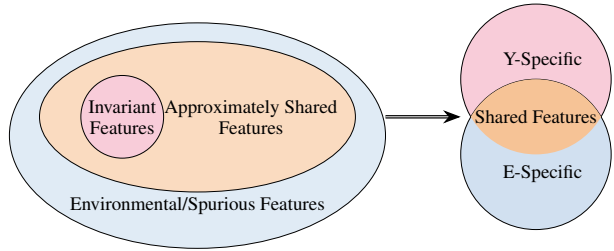

To resolve the limitations and paradoxes in previous works, we will answer the following questions:
{\em What features are robust to the natural distributional shifts in (multi-source) domain adaptation, and how can we learn them from purely observational data?}

\begin{figure}[H]
    \centering
    \includegraphics[width=0.7\linewidth]{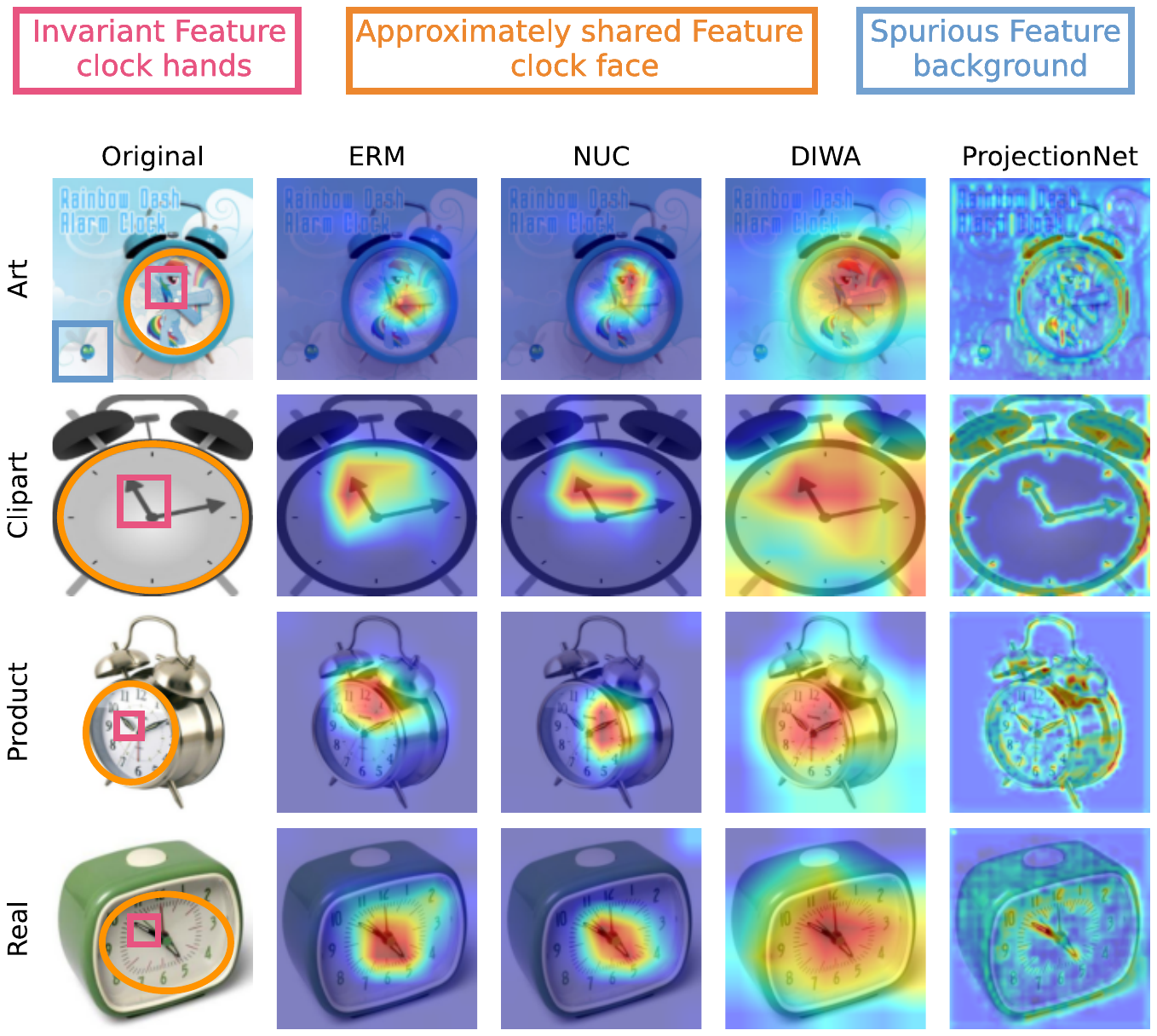}
    \caption{Feature Space Visualization: We show the GradCAM++ of the OfficeHome dataset with source pre-trained models feature space. We can see that ERM and NUC focus more locally; The DiWA feature is more globally distributed, while the feature space of ProjectionNet (ours) is more semantically meaningful.}\label{fig:feature_space}
\end{figure}

\paragraph{Our contribution.} 
In this work, we propose a novel theoretical framework that distinguishes \textcolor{yscolor}{content} (\textcolor{scolor}{approximately shared} $+$ \textcolor{ycolor}{invariant}) vs.\ \textcolor{ecolor}{environmental features} solely from observational data (based on how differently the features are utilized across tasks); see illustration diagram in \autoref{fig:diagram}. Under our framework, we design and analyze a learning procedure consisting of $(1)$ learning the content features via meta-representation learning on source domains and $(2)$ fine-tuning the learned representation on the target domain. Our theoretical analysis yields a smaller and more interpretable population risk bound by removing the irreducible term in the previous works such as~\citet{tripuraneni2021provable, DHK21}. It necessitates learning \textcolor{ycolor}{invariant} and \textcolor{scolor}{approximately shared} features instead of only \textcolor{ycolor}{invariant features} for a quick adaption to the target domain. Inspired by our theory, we proposed ProjectionNet, a more practical way to isolate the content (invariant+approximately shared) from environmental features. As illustrated in \autoref{fig:feature_space}, the feature space of ProjectionNet is more semantically meaningful than existing methods. Our findings effectively bridge the gap between different opinions in previous works mentioned above.

\subsection{Related Works}
We discuss relevant work in (multi-source) domain adaptation categorized by the types of distributional shifts that cause performance degradation.  
\paragraph{Selection bias} describes the phenomenon of data collected in separate routines to present as if drawn from different distributions. It can be introduced by selecting individuals, groups, or data for analysis so that proper randomization is not achieved. For selection bias on individual samples, it can represent general covariate shift, and prior works accordingly proposed relevant methods including importance or hard-sample reweighting~\citep{shimodaira2000improving,heckman1979sample,cortes2010learning,zadrozny2004learning, missing1,liu2021just,nam2020learning}, distribution matching/discrepancy minimization~\citep{cortes2015adaptation,ben2010theory,berthelot2021adamatch}, and domain-adversarial algorithms~\citep{ajakan2014domain,ganin2016domain,long2018conditional} between source and target domains in their feature representation space. 

Subpopulation shift or dataset imbalance comes from selection bias on groups or subpopulations. People have investigated label propagation ~\citep{cai2021theory,berthelot2021adamatch} or other consistency regularization~\citep{miyato2018virtual,xie2020unsupervised,yang2023sample} that migrate the predictions from source to target. Some more studies go beyond semi-supervised learning settings to contrastive representation learning~\citep{haochen2022beyond,shen2022connect,liu2021self} or self-training~\citep{wei2020theoretical,kumar2020understanding}. Under small covariate shift, learning algorithms have been investigated to achieve near minimax risks~\citep{lei2021near,pathak2022new}. 
\paragraph{Spurious correlation} corresponds to the dependence between features and labels that is not fundamental or not consistent across domains. It poses a significant challenge in deploying machine learning models, as they can lead to reliance on irrelevant or environmental features and poor generalization. 
Specifically, with spurious or environmental features accompanying the invariant/content features, studies show fine-tuning can distort pretraining features~\citep{kumar2022fine}, hurting out-of-distribution generalization. The existence of spurious features also brings about a trade-off between in-distribution and out-of-distribution performances.  

Model ensemble~\citep{kumar2022calibrated} or model soups~\citep{wortsman2022model} are introduced to learn diverse features to compensate for the effect of spurious features. 
More approaches include introducing auxiliary information through human annotation \citep{srivastava2020robustness} 
 or from multiple sources~\citep{xie2020n}, through invariant feature representation learning~\citep{arjovsky2019invariant,chen2022iterative,ahuja2020invariant}, through self-training~\citep{chen2020self} or overparametrization~\citep{sagawa2020investigation}, feature manipulation \citep{shen2022data}, adding regularizations~\citep{missing2,shi2023domain}, or through causal approaches~\citep{lu2021nonlinear}.
 
Finally, we discuss relevant results in representation learning that bear some resemblances but are fundamentally distinct in the presenting purposes.

\textbf{Theoretical understanding on meta representation learning.} 
Prior work on meta-representation learning~\citep{DHK21,CLL21,tripuraneni2020theory,tripuraneni2021provable} focused on when and how one can identify the ground-truth representation from multiple training tasks. All their results (as well as~\cite{missing3} that is more closely to our work) suffer from an irreducible term stemming from source representation error. Besides that, we are more interested in handling the spurious correlation and adapting to the new task sample efficiently.

\subsection{Notations}
Let $[n] = \{1,\ldots,n\}$. We use $\norm{\cdot}$ to denote the $\ell_2$ norm of a vector or the spectral norm of a matrix. We denote the Frobenius norm of a matrix as $\norm{\cdot}_F$. Let $\lag \cdot, \cdot \rag$ be the Euclidean inner product between two vectors or matrices. The $d \times d $ identity matrix is denoted as $I_d$.

For a vector $\bv \in \reals^{m}$  $v_k$ is the $k$-th entry and $\bv_{[k:\ell]}$ is the vector formed by $\bv$'s $k$-th to $\ell$-th entries, $1\le k < \ell \le m$.
For a matrix $A \in \reals^{m \times n}$, $m \ge n$, $A_k$ is the $k$-th column and $A_{[k:\ell]}$ is the matrix formed by $A$'s $k$-th to $\ell$-th columns, $1\le k < \ell \le n$. Denote $\mathcal{P}_{A} = A(A^\top A)^\dagger A^\top \in \reals^{m \times m}$, which is the projection matrix onto ${\rm span}(A) = \{A \bv \mid \bv \in \reals^n\}$. Here $\dagger$ stands for the Moore-Penrose pseudo-inverse. We define $\mathcal{P}^\perp_A = I_m - \mathcal{P}_A$ which is the projection matrix onto the orthogonal complement of ${\rm span}(A)$. The $k$-th smallest eigenvalue and $k$-th smallest singular value of $A$ are denoted by $\lambda_k (A)$ and $\sigma_k(A)$ respectively. We denote $\mathcal{O}(n)$ as the $n-$dimensional orthogonal group. 

We use $\gtrsim, \lesssim, \asymp$ to denote greater than, less than, equal to up to some constant. Our use of $O,\Omega,\Theta$ is standard.

\section{Methodology}
This paper focuses on multi-source domain adaptation. Suppose we have $E$ source environments indexed by $1,\ldots, E$ and a target environment indexed by $E+1$. For $e \in [E]$, each source environment provides a dataset with $n_1$ samples denoted by $\{(\bx^e_i, y^e_i)\}_{i=1}^{n_1}$. Similarly, for $e=E+1$, the target environment provides a dataset with $n_2$ samples denoted by $\{(\bx^e_i, y^e_i)\}_{i=1}^{n_2}$. For $e \in [E+1]$, each tuple $(\bx^e_i, y^e_i)$ is an i.i.d. sample drawn from an environment-specific data distribution $\mu_e$ over the joint data space $\mathcal{X} \times \mathcal{Y}$ where $\mathcal{X} \subseteq \reals^d$ is the input space and $\mathcal{Y} \subseteq \reals$ is the output space. We are interested in the few-shot setting where the number of training samples from the target environment is much smaller than that from the source environments, \ie $n_2 \ll n_1$.
\subsection{Data Generation Process }
\label{sec: data_generation}
As illustrated in Figure~\ref{fig:diagram}, there are two types of features: \textcolor{yscolor}{content} (\textcolor{scolor}{approximately shared} $+$ \textcolor{ycolor}{invariant}) and \textcolor{ecolor}{environmental (spurious) features}, which are distinguished based on how differently they are utilized across environments, \ie the variance of their correlation to $y$. To theoretically model the features of these two types, we introduce the following data generation process. 

Let $\phi_{D}: \mathcal{X} \rightarrow \mathcal{Z}$ be a representation function from the input space $\mathcal{X} \subseteq \reals^{d}$ to some latent space $\mathcal{Z} \subseteq \reals^{D}$. Let $\BPhi_{D}$ denote the function class to which $\phi_{D}$ belongs. Here $d$ is the ambient input dimension and $D$ is the representation dimension. 

For $k < D$, we define the following low-dimensional representation function classes induced by some fixed $\BPhi_D$
\begin{equation} \label{eqn: induced low-dimensional function class}
    \begin{aligned}
    \BPhi_{k} & := \{f: f(\bx) = \phi_{D[1:k]}(\bx), \phi_D \in \BPhi_D \},\\
     \BPhi_{D-k} & := \{f: f(\bx) = \phi_{D[k+1:D]}(\bx), \phi_D \in \BPhi_D \},
\end{aligned}
\end{equation}
where $\phi_{D[1:k]}(\bx)$ and $\phi_{D[k+1:D]}(\bx)$ are defined as 
$\phi_D(\bx) = [\phi_{D[1:k]}(\bx)^{\top}, \phi_{D[k+1:D]}(\bx)^{\top}]^\top$,
the splitting of $\phi_D \in \BPhi_D$. 

We use different linear predictors on top of a common representation function $\phi^{*}_{D}$ to model the input-output relations in each environment. For $e \in [E+1]$, let $\phi^*_{D} \in \BPhi_{D}$ be some ground-truth representation shared by every environment and $\btheta^{* 1}, \ldots, \btheta^{* E+1} \in \reals^{D}$ the environment-specific parameters characterizing the correlation of the features to the response $y$. The data generation process $(\bx ,y ) \sim \mu_e$ can be described as
\begin{equation}
\label{eqn: data generation}
    y = \langle \phi^*_D(\bx), \btheta^{*e} \rangle  + z, \quad \bx \sim p_{e}, \quad z \sim \mathcal{N}(0, \sigma^2),
\end{equation}
where $\bx$ and $z$ are statistically independent, and $p_e$ is some environment-specific input distribution. Furthermore, the environment-specific parameters follow the multivariate Gaussian meta-distribution\footnote{The assumption of Gaussian distribution is for simplicity. The fundamental requirement is for the meta distribution to be light-tail. }:
\begin{equation}
\label{eqn: meta distribution}
    \btheta^{*e} \overset{{\rm i.i.d}}{\sim} \mathcal{N}\left( \begin{bmatrix} \btheta^*\\ \bzero \end{bmatrix}, \begin{bmatrix} \Lambda_{11} & 0\\ 0 & \Lambda_{22} \end{bmatrix} \right),
\end{equation}
where $\btheta^* \in \mathbb{R}^k$, and $\Lambda_{11} \in \mathbb{R}^{k \times k}, \Lambda_{22} \in \mathbb{R}^{(D-k) \times (D-k)}$ are diagonal matrices with ascending entries, and $\| \Lambda_{11} \| < \| \Lambda_{22} \|$. 
Then \eqref{eqn: data generation} can be naturally split into
 \begin{align}
 \label{eqn: data generation component}
     y = {\color{yscolor}\lag \phi^*_{D[1:k]}(\bx), \btheta_{[1:k]}^{*e} \rag} + {\color{ecolor}\lag \phi^*_{D[k+1:D]}(\bx), \btheta_{[k+1:D]}^{*e} \rag}  + z.
 \end{align}
 In this setting, the first $k$ entries of $\phi^*_D(\bx)$ are the low-dimensional content features that are (approximately) shared by all environments. The correlation of these features to $y$ is relatively stable across environments because $\btheta^{*e}_{[1:k]}$ distributes around $\btheta^*$ with a small variance $\Lambda_{11}$. In particular, the \textcolor{scolor}{approximately shared features} are those corresponding to larger entries in $\Lambda_{11}$.  On the other hand, the remaining $D-k$ entries are the \textcolor{ecolor}{environmental features} that are uncorrelated with $y$ on average but their correlation to $y$ varies largely as the environment changes, characterized by $\| \Lambda_{22} \| > \| \Lambda_{11} \|$.
 \vspace{3mm}

\begin{remark}[Importance of approximately shared features]
The inclusion of approximately shared features in the low-dimensional representation solves some prior paradoxes. On one hand, intuitively one would like to prioritize learning invariant features as shown in ~\citet{arjovsky2019invariant}; on the other hand, the reason model soups~\cite{wortsman2022model} and model ensembles learn richer features but also are more robust to domain shifts is because the effect of environmental features is evened out during the model average (modeled by the zero-mean part in~\eqref{eqn: meta distribution}).
\end{remark}
\vspace{3mm}
\begin{remark}[Distinction to prior work]
Compared with the state-of-the-art works on meta-learning theory, our data generation process is more reasonable and closer to real data. First, some existing works~\citep{arjovsky2019invariant,ahuja2020invariant} consider that $y$ is generated only from the strictly invariant features and ignore the effect of environmental features, \ie for $e\in[E+1]$, 
\[
y = \langle \phi^*_{k}(\bx), \btheta^{*e} \rangle  + z,~\bx \sim p_{e}, \quad\text{where }\phi^*_{k} \in \BPhi_k, k < d.
\]
However, for real data, the distinction between invariant and environmental features is not a clear-cut binary division. Second, the representation learning works~\citep{DHK21,tripuraneni2020theory} make no distinction to the utilities of features (variance of correlation); they learn all features correlated to $y$ and have a different focus from us. Their risk bound also has an irreducible term from source tasks without investigating how the representation error can be eliminated through fine-tuning.
\end{remark}

\subsection{The Meta-Representation Learning Algorithm}
\label{sec: meta-representation learning algorithm}
Given the data generation process above, a natural question to ask is how we can learn the content features from the source environments, \ie a low-dimensional representation $\phi_k \in \BPhi_k$ and use the learned features on the target environment as a fine-tuning phase so that the performance is better than directly learning the target environment.

Before delving into the details of the algorithm, we introduce some auxiliary notations for clarity. For $e \in [E]$, we concatenate $n_1$ i.i.d. samples $\{(\bx^e_i, y^e_i)\}_{i=1}^{n_1}$ into a data matrix $X^e \in \reals^{n_1 \times d}$ and a response vector $\by^e \in \reals^n$, \ie
$X^e = \ls \bx^{e}_1, \ldots, \bx^{e}_{n_1} \rs^{\top}, \quad \text{and} \quad \by^e = \ls y^{e}_{1}, \ldots, y^{e}_{n_1} \rs^{\top}.$ Similarly, for the target environment, we have $X^{E+1} \in \reals^{n_2 \times d}$ and $\by^{E+1} \in \reals^{n_2}$.

Given some representation function $\phi_D \in \BPhi_D$ and data matrix $X^e$, $e\in [E]$, we overload the notation to allow $\phi_D$ to apply to all the samples in a data matrix simultaneously, \ie 
$
\phi_D(X^e) = \ls
    \phi_D(\bx^{e}_1), \ldots, 
    \phi_D(\bx^{e}_{n_1}) \rs^{\top} \in \reals^{n_1 \times D}.$ Similarly, for the target environment, we have $\phi_D(X^{E+1}) \in \reals^{n_2 \times D}$. 

Then our meta-representation learning algorithm can be described as follows.\\
\textbf{Source Pretraining:} For each source environment $e \in [E]$, we set the prediction function to be $\bx \mapsto \lag \phi_{k}(\bx), \btheta^{e} \rag$ $(\btheta^{e} \in \reals^{k})$. Therefore, using the training data from $E$ environments we can learn the content features represented by $\widehat{\phi}^{E}_{k} \in \BPhi_{k}$ via the following optimization:
\begin{equation}
\label{eqn: source vectorized}
    \begin{aligned}
   & \lp {\color{yscolor}\widehat{\phi}^{E}_k}, \widehat{\btheta}^1,\ldots,\widehat{\btheta}^E \rp\\
   & = \underset{\phi_k \in \BPhi_k, \btheta^e \in \mathbb{R}^k} {\argmin} \frac{1}{2n_1E} \sum_{e=1}^E \norm{\by^{e} - \phi_k(X^e)\btheta^e}^2.
    \end{aligned}
\end{equation}
\textbf{Target Finetuning:}
For the target environment, we solve the following optimization to learn the final predictor $\bx \mapsto \lag \phi_{D}(\bx), \btheta^{E+1} \rag$ $(\btheta^{E+1} \in \reals^{D})$ that uses $\widehat{\phi}^{E}_{k}$ obtained from~\autoref{eqn: source vectorized}:
\begin{equation}
\label{eqn: target}
    \begin{aligned}
    &\lp \widehat{\phi}^{E+1}_D, \widehat{\btheta}^{E+1} \rp\\ = &\underset{\phi_D \in \BPhi_{D}, \btheta^{E+1} \in \mathbb{R}^D} {\argmin} \frac{1}{2n_2}  \norm{\by^{E+1} - \phi_D(X^{E+1})\btheta^{E+1}}^2\\ &+ \frac{\lambda_1}{2n_2}\norm{\mathcal{P}_{\widehat{\phi}^{E}_k(X^{E+1})}^{\perp} \phi_D(X^{E+1})\btheta^{E+1}}^2 +\frac{\lambda_2}{2} \norm{\btheta^{E+1}}^2.
    \end{aligned}
\end{equation}
\autoref{eqn: target} is the empirical risk minimization on the target environment with two regularizing terms. The first one penalizes the prediction on the directions \textit{perpendicular} to the content features learned from source environments, which discourages the learning dynamics from capturing the environmental (spurious) features since we assume they are uncorrelated to $y$ on average. The second one is the standard $\ell_2$ regularization.
\begin{remark}
    The above learning procedure enables us to learn and utilize the content features purely from the observational data even without any causal information on the features.
\end{remark}

For $e \in [E+1]$, to evaluate whether a predictor $\bx \mapsto \lag \phi(\bx), \btheta \rag$ works well on some unseen data $(\bx,y)\sim \mu^{e}$ sampled from that environment, we are interested in the excess risk
${\rm ER}_{e}(\phi, \btheta)
     = \frac{1}{2}\E_{\bx\sim p^e} \lp \ls\lag \phi^{*}_D(\bx), \btheta^{*e} \rag - \lag \phi(\bx), \btheta \rag\rs^2\rp$
and its expectation over the meta distribution $\E_{\btheta^{*e}}[{\rm ER}_{e}(\phi, \btheta)]$. For the target environment, we abbreviate $\text{ER}_{E+1}$ as $\text{ER}$.

\subsection{Isolating Content Features in Practice}

The optimizations in~\eqref{eqn: source vectorized} and~\eqref{eqn: target} are intractable since searching through the function class and finding the optimal representation function is generally expensive. In this section, we use the following two methods to approximate them in practice. 

\subsubsection{Nuclear Norm Regularization}

\textbf{Source Pretraining:} To learn a low-dimensional representation, we choose a model $\phi_{\btheta_{\text{rep}}}(\cdot)$ with trainable weights $\btheta_{\text{rep}}$ (such as ResNet) and control the complexity of its output instead of searching the whole function class for the optimal representation function. For $e \in [E]$, we learn the predictor $\bx \mapsto \lag \phi_{\btheta_{\text{rep}}}(\bx), \btheta \rag$ where $\btheta$ is the weight of the last layer shared by all source environments via
\vspace{-0.1in}
\begin{equation}
\vspace{-0.1in}
    \begin{aligned}
    \min_{\btheta_{\text{rep}}, \btheta}
    \frac{1}{E} \sum_{e=1}^{E}
    \mathcal{L}( \lag \phi_{\btheta_{\text{rep}}}(X^{e}), \btheta \rag, \by^{e})+ \lambda_{\rm nuc}\cdot\|\phi_{\btheta_{\text{rep}}}(X^{e})\|_*
    \label{eqn: nuc}
\end{aligned}
\end{equation}

where $\mathcal{L}(\cdot,\cdot)$ is the standard Cross-Entropy (CE) loss. The nuclear norm regularization~\citep{shi2023domain} helps us learn a low-rank representation that captures the \textcolor{yscolor}{content features} from the source environments. The learned $\phi_{\btheta_{\text{rep}}}$ will be utilized in the target environment for finetunning. We vary regularization strength $\lambda_{\rm nuc} \in \{0.01, 0.1\}$ to control the complexity of the feature space, larger $\lambda_{\rm nuc}$ will lead to an more invariant feature space. However, tuning the $\lambda$ requires prior over the feature space or training multiple times.

\textbf{Target Finetuning:} We use the source pretrained $\phi_{\btheta_{\text{rep}}}$ and $\btheta$ as the initialization and use ERM in the finetuning stage.

\subsubsection{ProjectionNet (ours)}
The nuclear norm regularization has a fundamental limitation: we need to tune the regularization strength $\lambda_{\rm nuc}$ to control the complexity of the feature space. However, in practice, we may have no prior and it is unrealistic to tune the regularizer multiple times for large datasets source pretraining. To address this issue, we propose \textbf{ProjectionNet}, a training method that naturally distinguishes \textcolor{ycolor}{invariant features $\phi_y$}, \textcolor{scolor}{approximately shared features $\phi_s$} and \textcolor{ecolor}{environment-specific features $\phi_e$}.

\begin{table*}[t]
    \centering
    \scalebox{.58}{\begin{tabular}{lllllllllllllllllllll}
\toprule
 & \multicolumn{5}{c}{OfficeHome} & \multicolumn{5}{c}{PACS} & \multicolumn{5}{c}{TerrainCognita} & \multicolumn{5}{c}{VLCS} \\
 & A & C & P & R & \textbf{Mean} 
 & A & C & P & S & \textbf{Mean}  
 & L100 & L38 & L43 & L46 & \textbf{Mean} 
 & C & L & S & V & \textbf{Mean}  \\
\midrule
DANN & 0.6091 & 0.0297 & 0.6599 & 0.7408 & 0.5099 & 0.1463 & 0.1026 & 0.6707 & 0.2112 & 0.2827 & 0.0890 & 0.0770 & 0.2040 & 0.2806 & 0.1627 & 0.9437 & 0.3636 & 0.4006 & 0.4438 & 0.5379 \\
DIWA & 0.6914 & 0.5767 & 0.7838 & 0.7982 & 0.7125 & 0.9034 & 0.8154 & 0.9880 & 0.8244 & \textbf{0.8828} & 0.5776 & 0.4517 & 0.5995 & 0.4354 & 0.5161 & 0.9718 & 0.6014 & 0.7350 & 0.8047 & 0.7782 \\
ERM & 0.6914 & 0.5240 & 0.7793 & 0.8211 & 0.7039 & 0.8244 & 0.7906 & 0.9641 & 0.8270 & 0.8515 & 0.5190 & 0.4415 & 0.5693 & 0.3861 & 0.4789 & 0.9718 & 0.6503 & 0.7098 & 0.8136 & 0.7864 \\
NUC-0.01 & 0.6996 & 0.5446 & 0.8018 & 0.8119 & 0.7145 & 0.8244 & 0.7521 & 0.9820 & 0.7888 & 0.8368 & 0.2025 & 0.4138 & 0.5516 & 0.2177 & 0.3464 & 0.3169 & 0.6692 & 0.7652 & 0.1391 & 0.4726 \\
NUC-0.1 & 0.7284 & 0.5423 & 0.7995 & 0.8188 & 0.7223 & 0.8195 & 0.8120 & 0.9880 & 0.7990 & 0.8546 & 0.6181 & 0.4969 & 0.6121 & 0.2177 & 0.4862 & 0.9789 & 0.6842 & 0.7287 & 0.8077 & \textbf{0.7999} \\
PN-Y & 0.7243 & 0.4920 & 0.8063 & 0.8417 & 0.7161 & 0.7707 & 0.8128 & 0.9641 & 0.7150 & 0.8157 & 0.5561 & 0.5277 & 0.5340 & 0.4762 & \textbf{0.5235} & 0.9789 & 0.5594 & 0.7382 & 0.8166 & 0.7733 \\
PN-Y-S & 0.7284 & 0.5057 & 0.8108 & 0.8486 & \textbf{0.7234} & 0.7668 & 0.8111 & 0.9641 & 0.6997 & 0.8104 & 0.5612 & 0.5277 & 0.5290 & 0.4728 & 0.5227 & 0.9859 & 0.5594 & 0.7382 & 0.8343 & 0.7795 \\
\bottomrule
\end{tabular}}
    \caption{Domain Generalization Results: We use the source pretrained model to directly test the target domain to evaluate the domain generalization ability of the pretrained model. NUC-0.1 and NUC-0.01 stand for different regularization strength $\lambda_\text{NUC}$ in \autoref{eqn: nuc} ProjectionNet (PN) can achieve similar or better results than the baseline methods.}\label{table:uda}
\end{table*}

\begin{figure}[H]
    \begin{center}
        \scalebox{.8}{
                
\begin{tikzpicture}[x=0.8cm, y=1.1cm]
\foreach \m/\l [count=\y] in {1,2}
    \node [neuron] (input-\m) at (0,2.5-\y) {};

\foreach \m [count=\y] in {1,2}
    \node [basefeature] (basefeature-\m) at (2,2.5-\y) {};

\foreach \m [count=\y] in {1,2,3}
    \node [hidden] (hidden-\m) at (4,3-\y) {};

\node [yneuron] (hidden-1) at (4,3-1) {$\phi_{y}$};
\node [hidden] (hidden-2) at (4,3-2) {$\phi_{s}$};
\node [eneuron] (hidden-3) at (4,3-3) {$\phi_{e}$};


\node [yneuron] (output-y) at (6,1.5) {$y$};

\node [eneuron] (output-e) at (6,0.5) {$e$};

\foreach \i in {1,...,2}
    \foreach \j in {1,...,2}
        \draw [->] (input-\i) -- (basefeature-\j);

\foreach \i in {1,...,2}
    \foreach \j in {1,...,3}
        \draw [->] (basefeature-\i) -- (hidden-\j);

\foreach \i in {1,...,2}
    \draw [->] (hidden-\i) -- (output-y);

\foreach \i in {2,...,3}
    \draw [->] (hidden-\i) -- (output-e);
\foreach \i in {3,...,3}
    \draw [->] (hidden-\i) -- (output-e);

\node [annotation] at (0.8,-0.8) {Feature Encoder ($\phi$)};

\node [annotation] at (2.8,-1) {Projector ($\mathcal{P}$)};

\node [annotation] at (5.3,-1) {Classifier ($\btheta_y, \btheta_e$)};

\node [annotation] at (0,3) {Input};
\node [annotation] at (2,3) {Base Feature};
\node [annotation] at (4,3) {\small{Disentangled Feature}};
\end{tikzpicture}
        }
    \end{center}
    \caption{
    ProjectionNet: We disentangle the base representation $\phi$ into \textcolor{ycolor}{target-specific feature $\phi_y$}, \textcolor{scolor}{approximately shared feature $\phi_s$} and \textcolor{ecolor}{environment-specific feature $\Phi_e$}.\ (\textcolor{ycolor}{$\mathcal{P}_y$}, \textcolor{scolor}{$\mathcal{P}_s$}, \textcolor{ecolor}{$\mathcal{P}_e$}) are the projection heads applied to the base feature. The [\textcolor{ycolor}{$\phi_y$}, \textcolor{scolor}{$\phi_s$}] are used in the target label prediction. [\textcolor{scolor}{$\phi_s$}, \textcolor{ecolor}{$\phi_e$}] are used in the environment label prediction. 
    }\label{fig:pn}
\end{figure}

\textbf{Source Pretraining:} For each training sample we consider the triplet $(\bx, y, e)$ where $e$ is the environment $(\bx, y)$ belongs to. Let $\phi_{\btheta_\text{rep}}$ with trainable weights $\btheta_\text{rep}$ be some feature map. Besides the model weights, we would like to train 3 projections $\mathcal{P}_y$, $\mathcal{P}_{s}$, $\mathcal{P}_e$ such that $\phi_{y} = \phi_{\btheta_\text{rep}}(\bx)\mathcal{P}_y$, $\phi_{s} =\phi_{\btheta_\text{rep}}, (\bx)\mathcal{P}_s$, $\phi_{e} =\phi_{\btheta_\text{rep}}(\bx)\mathcal{P}_e$. The algorithm is summarized in~\eqref{eqn: pn}. 

\begin{equation} \label{eqn: pn}
    \min_{\btheta_\text{rep},\btheta_y, \btheta_e, \mathcal{P}_e, \mathcal{P}_y, \mathcal{P}_s} \mathcal{L}_{y} + \mathcal{L}_{e} + \mathcal{L}_{\text{disentangle}} + \lambda \mathcal{L}_{\text{reg}}
\end{equation}
where 
\[
\begin{aligned}
    \mathcal{L}_{y} &= 
    \frac{1}{n_1} \sum_{i=1}^{n_1}
    \mathcal{L}\left( 
        \begin{bmatrix}
            \phi_{\btheta_\text{rep}}(\bx_i) \textcolor{ycolor}{\mathcal{P}_y} \\
            \phi_{\btheta_\text{rep}}(\bx_i) \textcolor{scolor}{\mathcal{P}_s} 
        \end{bmatrix} \btheta_y, y_i \right) , \\
    \mathcal{L}_{e} &= \frac{1}{n_1} \sum_{i=1}^{n_1} \text{CE} \left(
        \begin{bmatrix}
            \phi_{\btheta_\text{rep}}(\bx_i) \textcolor{scolor}{\mathcal{P}_s} \\ 
            \phi_{\btheta_\text{rep}}(\bx_i) \textcolor{ecolor}{\mathcal{P}_e} 
        \end{bmatrix} \btheta_e, e_i \right) , \\
    \mathcal{L}_{\text{disentangle}} &= \| \mathcal{P}_y^{\top} \mathcal{P}_e \|_{F}^{2} + \| \mathcal{P}_y^{\top} \mathcal{P}_s \|_{F}^{2} + \| \mathcal{P}_e^{\top} \mathcal{P}_s \|_{F}^{2}, \\
    \mathcal{L}_{\text{reg}} &= \left\| \mathcal{P}_e^{\top} \mathcal{P}_e - I \right\|_{F}^{2} + \left\| \mathcal{P}_y^{\top} \mathcal{P}_y - I \right\|_{F}^{2} + \left\| \mathcal{P}_s^{\top} \mathcal{P}_s - I \right\|_{F}^{2}.
\end{aligned}
\]
This method is based on the assumption that the approximately shared features are correlated to both $y$ and the environment. We jointly minimize the 4 loss functions above. $\mathcal{L}_{y}$ is the standard empirical minimization, it extracts the features that are correlated to the response $y$, \ie \textcolor{ycolor}{invariant features} and \textcolor{scolor}{approximately shared features}. $\mathcal{L}_{e}$ extracts the features that determine the environment of a training sample $(\bx,y)$, \ie  \textcolor{scolor}{approximately shared features} and \textcolor{ecolor}{environmental features}. $ \mathcal{L}_{\text{disentangle}}$ ensures that the learned learned $\mathcal{P}$'s are almost orthogonal to each other and $\mathcal{L}_{\text{reg}}$ ensures the columns of the learned $\mathcal{P}$'s are almost orthogonal. The method is visualized in Figure~\ref{fig:pn}.

\textbf{Target Finetuning:} We can flexibly choose to utilize $\phi_{y}$ or $\left[\phi_{y}, \phi_{s}\right]$ as the initialization for the target finetuning phase, which depends on whether we want to use invariant features or content features.

\section{Theoretical Analysis}
\label{sec: theoretical analysis}
In this section, we will present the statistical analysis of the learning framework in Section~\ref{sec: meta-representation learning algorithm}. Before stating the main theorem, we make the following assumptions.
\begin{assumption}[Homogeneous input distribution]
\label{assumption: homogeneous input distribution}
    We assume that all the tasks follow the same input distribution, \ie for $e\in [E+1]$, $p_1=\cdots = p_{E+1} = p$.
\end{assumption}
This assumption is for the simplicity of theoretical analysis since otherwise we may need stronger assumptions on the representation function class. When the representation function is linear, we show that this assumption can be relaxed to moment boundedness conditions (see Section~\ref{sec: linear}).

We will use the covariance between two representations to measure the distance between representation functions. Our notion of representation covariance is generalized from~\citet{DHK21} to measure the distance from different function classes. We defer the detailed properties of representation covariance to Section~\ref{sec: nonlinear detail}.
\begin{definition}[Covariance between representations]
\label{def: representation covariance informal}
    Let $p$ be a distribution over $\reals^d$ and $\BPhi_D$ be some function class with $\BPhi_{k} / \BPhi_{D-k}$ defined in~\eqref{eqn: induced low-dimensional function class}. For two representation functions $\phi \in \BPhi_{d_1}$, $\phi' \in \BPhi_{d_2}$, $d_1,d_2 \in \{D, k, D-k\}$, we define the covariance between $\phi$ and $\phi'$ with respect to $p$ as 
    $\Sigma_p(\phi, \phi') = \E_{\bx \sim p} [\phi(\bx)\phi'(\bx)^\top] \in \reals^{d_1 \times d_2}$, where $d_1$ and $d_2$ are the output dimension of $\phi$ and $\phi'$.We also define the symmetric covariance as 
\[
 S_{p}(\phi, \phi') = \begin{bmatrix}
      \Sigma_p(\phi, \phi) &  \Sigma_p(\phi, \phi')\\  \Sigma_p(\phi', \phi) &  \Sigma_p(\phi', \phi') 
 \end{bmatrix} \in \reals^{(d_1+d_2) \times (d_1+d_2)}.
\]  
If $\phi := \phi'$, we abbreviate $\Sigma_p(\phi, \phi) := \Sigma_p(\phi), \quad S_p(\phi, \phi) = S_p(\phi)$.
\end{definition}

We make Assumption~\ref{assumption: nonlinear point} and~\ref{assumption: nonlinear uniform} on the ground-truth representation function class $\BPhi_D$ and the input distribution $p$, which ensure the point-wise and uniform concentration of empirical covariance $S_{\widehat{p}}$ to their population counterpart $S_{p}$. In our main theorem, we assume $n_1$ is large enough to guarantee uniform concentration and $n_2$ is large enough to guarantee point-wise concentration (this implies $n_1 \gtrsim n_2$). 

The following assumption guarantees $ \lag \phi^*_{D[1:k]}(\bx), \btheta_{[1:k]}^{*e} \rag $ dominants in~\eqref{eqn: data generation component}, that is, the contribution of content features to $y$ is stronger than that of environmental features.
\begin{assumption}[Dominance of content features]
\label{assumption: trace assumption nonlinear}
    \[
    \begin{aligned}
        \frac{\norm{\Sigma_{p} (  \phi^*_{D[1:k]} )}}{\norm{\Sigma_{p} (   \phi^*_{D[k+1:D]} )}} &\gtrsim \frac{\Tr(\Lambda_{22})}{\norm{\btheta^*}^2 + \Tr(\Lambda_{11})}\\
    \end{aligned}
    \]
\end{assumption}

\begin{assumption}[Diverse source tasks]
\label{assumption: diverse source tasks}
Let $\Theta^* = [\btheta^{*1},\ldots, \btheta^{*E}]$.
The smallest singular value of $\Theta^*$ satisfies $\sigma^2_{\min}(\Theta^*) \gtrsim \frac{E}{\norm{\btheta^*}^2+ \Tr(\Lambda_{11})}$.
\end{assumption}
This assumption requires source tasks to utilize all directions of the representation function, or otherwise the weakest direction will be hard to learn. 

Our main results are the following theorems.
\begin{theorem}
[Source environment guarantee, informal version of Theorem~\ref{lemma: guarantee of source nonlinear}]
\label{lemma: guarantee of source nonlinear short}  
Under Assumption~\ref{assumption: homogeneous input distribution},~\ref{assumption: nonlinear uniform},~\ref{assumption: trace assumption nonlinear}, and~\ref{assumption: diverse source tasks}, if $n_1$ is large enough, the average excess risk across the source environments with probability $1-o(1)$ satisfies
\[
\begin{aligned}
    \overline{\rm ER}(\widehat{\phi}^E_k, \widehat{\btheta}^{1},\ldots,\widehat{\btheta}^{E} )&:= \frac{1}{E} \sum_{e=1}^{E} {\rm ER}_{e}(\widehat{\phi}^E_k, \widehat{\btheta}^{e}) \\&\lesssim \underbrace{\sigma\sqrt{\frac{\mathcal{C}_{\text{cont}}}{n_1 E}}+  \mathcal{C}_{\text{env}}}_{:= \text{RE}}    
\end{aligned}
\]
where $\mathcal{C}_{\text{cont}} \asymp O(k)$ measures the complexity of content features and $\mathcal{C}_{\text{env}} \asymp O(D-k)$ measures the complexity of environmental features.

\end{theorem}

 In this result, the first term indicates we are able to learn the (approximately) shared representation with all $n_1E$ samples and the second term is the non-vanishing error caused by environmental features since we only learn content features. This result is vacuous but will be useful for the target environment. 
\begin{theorem}[Target environment guarantee, informal version of Theorem~\ref{thm: nonlinear main}]
    \label{thm: nonlinear main short}
    Under Assumption~\ref{assumption: homogeneous input distribution},~\ref{assumption: nonlinear point},~\ref{assumption: nonlinear uniform},~\ref{assumption: trace assumption nonlinear}, and~\ref{assumption: diverse source tasks}, we further assume that $n_1$ and $n_2$ are large enough but still $n_1 \lesssim n_2$. Under proper choice of $\lambda_1$ and $\lambda_2$, the excess risk of the learned predictor $\bx \mapsto \lag \widehat{\phi}^{E+1}_D (\bx), \widehat{\btheta}^{E+1} \rag$ in~\eqref{eqn: target} on the target domain with probability $1-o(1)$ satisfies
    \[
    \begin{aligned}
        &\E_{\btheta^{*E+1}}\ls{\rm ER}(\widehat{\phi}^{E+1}_D, \widehat{\btheta}^{E+1})\rs\\ &\lesssim \underbrace{\sigma\sqrt{\frac{\overline{\mathcal{C}'}_{\text{env}}}{n_2}}  + \sigma\sqrt{\frac{ k}{n_2}}}_{\text{adaptation error}} +\underbrace{\sigma\sqrt{\frac{{\rm RE} }{n_2}}}_{\text{representation error}},
    \end{aligned}
    \]
where $\overline{\mathcal{C}'}_{\text{env}} \asymp O(D-k)$ measures the complexity of the environmental representation function class.
\end{theorem}
In this excess risk, the first two terms are the adaptation error: it roughly is what the result will look like if we fine-tune the model with the perfect $\phi_{D[1:k]}^*$. The third term is the representation error caused by learning $\phi_{D[1:k]}^*$ with finite samples from the source tasks. It can be further reduced by target samples in the fine-tuning phase. Compared to previous work~\citet{CLL21,DHK21,tripuraneni2020theory,tripuraneni2021provable} that presents irreducible representation error (which does not go to zero as $n_2 \rightarrow \infty$) plus adaptation/estimation error, we no longer have the irreducible term from source tasks. 
\vspace{3mm}
\begin{remark}[Linear representation]
    When the ground-truth representation function class $\BPhi_{D}$ is linear, we propose a spectral algorithm inspired by~\citet{tripuraneni2021provable} along with its statistical analysis in Section~\ref{sec: linear}. The idea is similar to the general case discussed above but is tractable in terms of optimization and our fine-tuning phase on the target environment removes the irreducible term from the risk bound in~\citet{tripuraneni2021provable}. 
\end{remark}
\section{Experiments}

In Section~\ref{sec: synthetic}, we verify our results with simulations under our exact theoretical framework in Section~\ref{sec: linear}, which is the case when the representation function is linear. In Section~\ref{sec: real}, we approximate the theoretical framework controlling the feature space implicitly by using the \textbf{Nuclear Norm (NUC)} regularization and explicitly by learning the \textit{Disentangled Representation} via \textbf{ProjectionNet}. We show that controlling the feature space to learn the approximately shared feature will benefit the downstream domain adaptation.


\subsection{Synthetic Data}\label{sec: synthetic}
\vspace{-0.2cm}
\begin{figure}[H]
    \begin{center}
    \scalebox{0.8}{
    \includegraphics[width=0.8\linewidth]{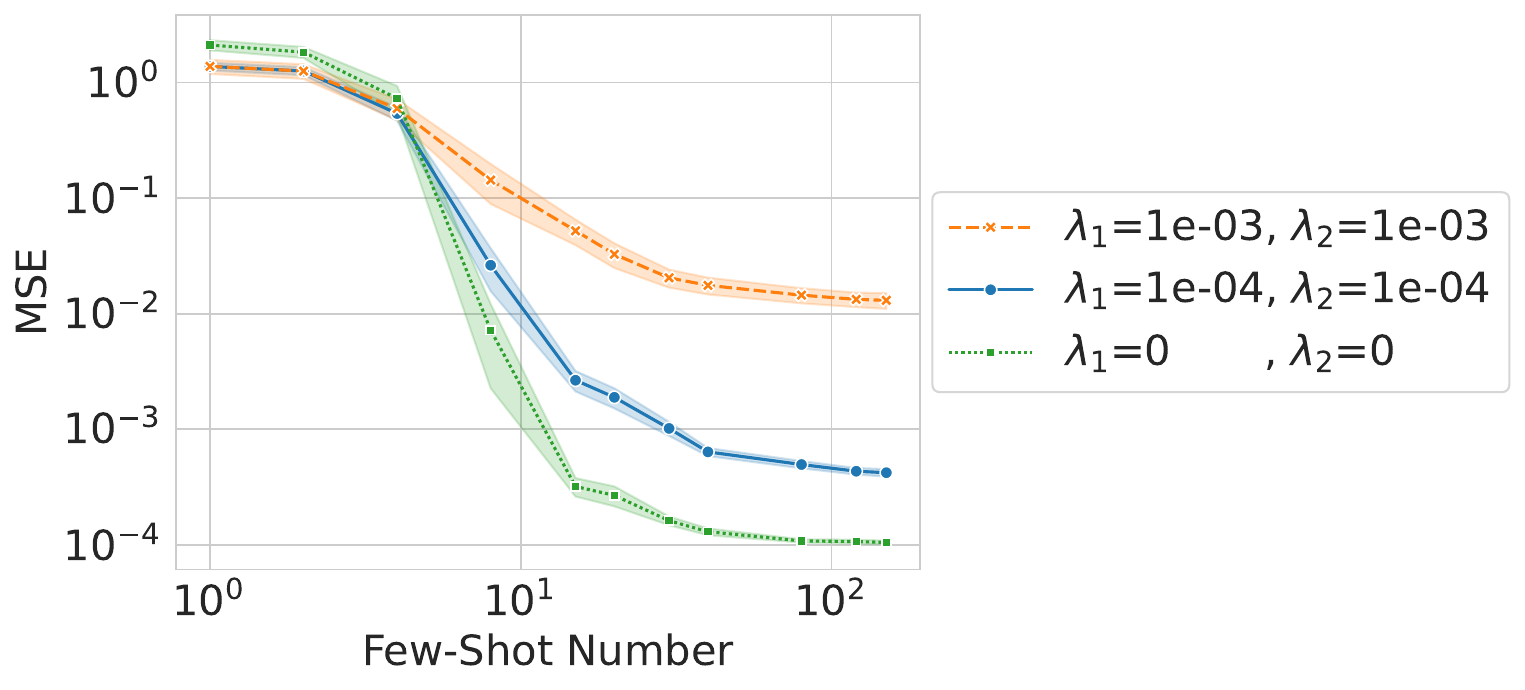}
    }
    \caption{Target Loss v.s.few-shot number plot for linear synthetic data. Both ${\rm Reg}_1$ and ${\rm Reg}_2$ in~\autoref{eqn: linear fine-tune} help the domain adaptation performance.}\label{fig:linear}
    \end{center}
\end{figure}
\vspace{-0.5cm}
We use synthetic data to validate the linear representation.
The ground-truth representation $R^*$ is a randomly generated orthogonal matrix\footnote{Details in \autoref{subsec:linear_synthetic}}, and for $e \in [E+1]$, the environment-specific parameters $\btheta^{*e}$ are sampled according to the meta distribution~\autoref{eqn: meta distribution}. In \textbf{Few-shot domain adaptation}, we test the linear model via the close form solution of \autoref{eqn: linear fine-tune}. 
We present the loss v.s.\ few-shot number plot in \autoref{fig:linear}. 

\subsection{Real Data}\label{sec: real}

We conducted an empirical evaluation of learning behaviors on VLCS~\citep{fang2013unbiased}, OfficeHome~\citep{venkateswara2017deep}, TerraIncognita~\citep{beery2018recognition} subset from DomainBed benchmark~\citep{gulrajani2020search} for the real dataset experiment. We use the same training setting as DeepDG~\citep{deepdg}. We use three domains as source environments and one as the target environment. In the source domain training stage, we train the model 60 epochs with the source data in the source domain training stage. The model that performed best on the source data validation set was selected for further testing. This model was then applied to the target domain to assess its \textbf{domain generalization} capabilities. To test whether the representation is adaptive to the target domain, we do \textbf{linear probing} on the target domain with different sizes of target data; the tuning of the hyperparameter and selecting the best model are based on the target data validation set. 


\subsection{Training Details}

In the target finetuning stage, we initialize the classifier from source pretrained $\btheta$ over different feature parts: For example, PN-Y is the backbone with the classifier only using the target-specific feature. \footnote{We have different training hyperparameters space with DomainBed. The absolute performance may differ from the DomainBed benchmark, but the relative rank is consistent. All the runs are averaged over 5 runs seeds. We show the figure with the shaded area representing the Standard Error.  We describe the ablation study, full results and more detailed parameter settings in the appendix. Our code is available at \url{https://github.com/Xiang-Pan/ProjectionNet}.}.

\textbf{Domain Generalization}\label{subsec:domain_generalization}
We show the results of Domain Generalization in \autoref{table:uda}. The model is trained on the $D_{S}$ and directly tested on the $D_{T}$. We show that ProjectionNet can achieve similar or better performance than DiWA with \textbf{20 times} less training time and without hyperparameter tuning headaches like NUC.\@

\begin{figure}[H]
    \begin{center}
        \scalebox{0.66}{
            \includegraphics{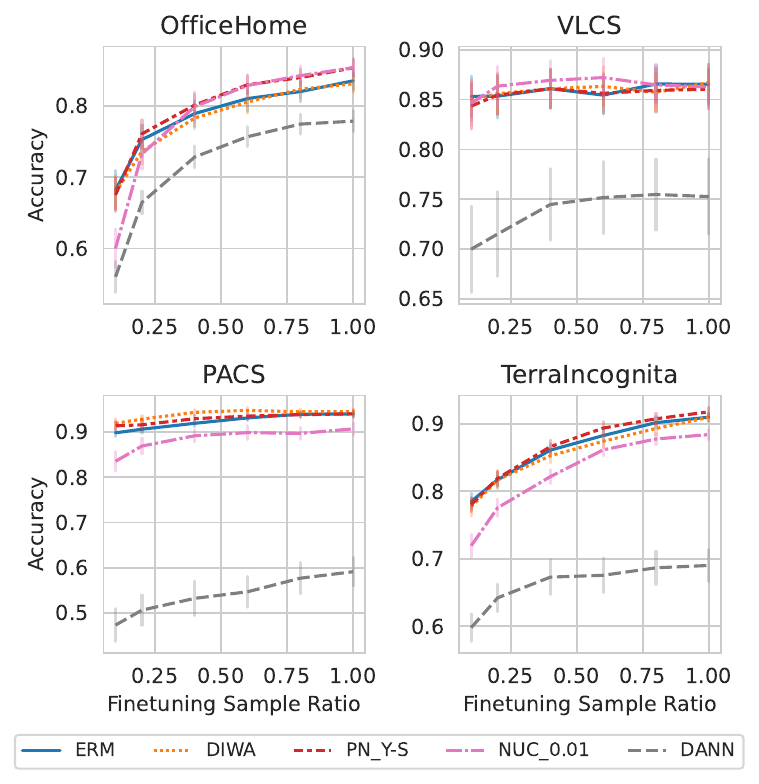}
        }
    \end{center}
    \vspace{-0.7cm}
    \caption{Linear Probing Results: We show the linear probing results, ProjectionNet share similar adaptive performance as DiWA}\label{fig:linear_probing}
\end{figure}

\textbf{Linear Probing}\label{subsec:linear_probing} We show the linear probing results in \autoref{fig:linear_probing}. With the target data, the linear probing performance increases, which suggests the feature from source pretraining is adaptive. However, there is a gap between the linear probing and target finetuning results. This suggests that we need to adjust the feature space to learn the target domain-specific feature.


\begin{figure}[H]
    \centering
    \scalebox{0.66}
    {\includegraphics{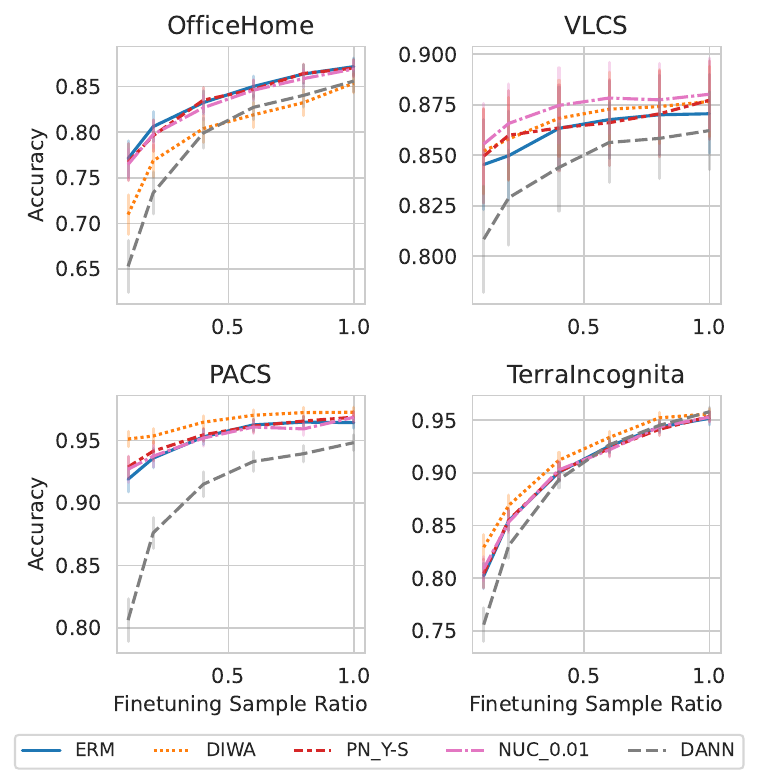}
    }
    \vspace{-0.7cm}
    \caption{Target Finetuning Results: We show the target finetuning results with different feature parts. We can see that PN-YS and PN-YSE can achieve similar or better results than the traditional diverse feature learning method (DiWA) and strict invariant feature learning method (DANN).
   }\label{fig:target_finetune}
\end{figure}

\textbf{Target Finetuning}\label{subsec:domain_adaptation} We tested the fixed backbone and feature encoder learning rate decay setting in domain adaptation; we show that for finetuning, we also need to adjust the feature encoder to learn the target domain-specific feature to achieve good performance.


\textbf{Feature Space Visualization} To visualize the feature space and valid our hypothesis that the ProjectionNet learns approximately shared features helps, we visualize OfficeHome dataset alarm clock class using the GradCAM++~\citep{chattopadhay2018grad} with source pretrained models feature space in \autoref{fig:feature_space}.

\subsection{Empircal Takeaways}

We found that in unsupervised domain generalization \autoref{table:uda} or small target samples, methods with more invariant features (PN-Y, NUC-0.1) may achieve good performance; with large target fine-tuning samples, more diverse feature space will perform better. However, achieving invariant and diverse feature space is hard. ProjectionNet mitigates the paradox by disentangling the feature space into pieces with different invariance, which can be optionally used in downstream tasks.

Besides, we observe the following phenomena in our experiments.
\textbf{Approximately Shared feature is not always helpful}: For datasets that domain generalization can achieve good performance (e.g., VLCS), invariant feature learning methods (DANN, PN-Y) will achieve good adaptation performance with finetuning.  
\textbf{Feature space adjustment is helpful with large target dataset size}: We see a clear gap between the target finetuning and the linear probing results, which suggests it is hard to learn a linear classifier as well as the target finetuning results. This process is comprehended as integrating the specific feature of the target domain into the representation. We show an ablation study that targets finetuning with feature learning rate decay, which decreases the performance most of the time (except for the VLCS dataset), which suggests that we need to adjust the feature space to learn the target domain-specific feature.
\textbf{Disentangled Feature is more adaptive and semantical meaningful}: In the domain generalization setting, the PN-Y feature can achieve better performance than strict invariant feature learning (DANN, Large NUC Penalization) in most cases. In the target finetuning setting, we can see that PN-Y-S and PN-YSE can efficiently get similar or better results than the traditional diverse feature learning method (DiWA) and strict invariant feature learning method (DANN).











\section{Conclusion}
In this work, we present a theoretical framework for multi-source domain adaptation to incorporate different distributional shifts. Our theoretical framework naturally distinguishes content features from environmental features. We investigate the algorithms that can reduce the spurious correlation under our framework and derive generalization bound on the target distribution. We prove how adding regularization in both pre-training and fine-tuning phases reduces the impact of spurious features and quickly adapts to the target task with few samples. Our simulations and real-data experiments also support our findings.

\clearpage
\bibliography{icml2024}
\bibliographystyle{icml2024}

\newpage
\appendix
\onecolumn

\section{General nonlinear representations}\label{sec: nonlinear detail}
In this section, we will present the omitted details in Section~\ref{sec: theoretical analysis} including Assumption~\ref{assumption: nonlinear point} and~\ref{assumption: nonlinear uniform}, the complete version of Theorem~\ref{lemma: guarantee of source nonlinear short} and~\ref{thm: nonlinear main short}, and the proof of them.

We make the following assumptions on the representation function class $\BPhi_D$ and the input distribution $p$, which ensures the concentration properties of the representation covariances.

\begin{assumption}[Point-wise concentration]
    \label{assumption: nonlinear point}
    For some failure probability $\delta \in (0,1)$, there exists a positive integer $N_{\rm point}(\BPhi_D, p, \delta)$ such that if $n \ge N_{\rm point}(\BPhi_D, p, \delta)$, then for any two given representation functions $\phi \in \BPhi_{d_1}$, $\phi' \in \BPhi_{d_2}$, $d_1,d_2 \in \{D, k, D-k\}$, $n$ i.i.d. samples of $p$ will with probability $1-\delta$ satisfy 
    \[
    0.9 S_{p}(\phi, \phi') \preceq S_{\widehat{p}}(\phi, \phi') \preceq 1.1 S_{p}(\phi, \phi')
    \]
    where $\widehat{p}$ is the empirical distribution over the $n$ samples.
\end{assumption}

\begin{assumption}[Uniform concentration]
    \label{assumption: nonlinear uniform}
    For some failure probability $\delta \in (0,1)$, there exists a positive integer $N_{\rm unif}(\BPhi_D, q, \delta)$ such that if $n \ge N_{\rm unif}(\BPhi_D, q, \delta)$, then $n$ i.i.d. samples of $p$ will with probability $1-\delta$ satisfy 
    \[
    \begin{aligned}
        0.9 S_{p}(\phi, \phi') \preceq S_{\widehat{p}}(\phi, \phi') \preceq 1.1 S_{p}(\phi, \phi'),\\ \forall \phi \in \BPhi_{d_1}, \phi' \in \BPhi_{d_2},
    \end{aligned}
    \]
    where $d_1,d_2 \in \{D, k, D-k\}$ and $\widehat{p}$ is the empirical distribution over the $n$ samples.
\end{assumption}

Assumption~\ref{assumption: nonlinear point} and~\ref{assumption: nonlinear uniform} are the conditions of the representation function class and the input distribution that ensure the concentration of empirical covariances to their population counterpart. Since $\BPhi_k / \BPhi_{D-k}$ are function classes induced by some fixed $\BPhi_D$, the concentration of $S_{\widehat{p}}$ in $\BPhi_{D}$ automatically guarantees its concentration in $\BPhi_{k}$, $\BPhi_{D-k}$ , and their union.

Since uniform concentration is stronger than point-wise concentration, we expect $N_{\rm unif}(\BPhi_D, q, \delta) \gtrsim N_{\rm point}(\BPhi_D, q, \delta)$. In particular, in the linear case discussed in Section~\ref{sec: main linear} where $\BPhi_{D} = \{f: f(\bx) = R^\top \bx, R \in \mathcal{O}(d)\}$ and $\BPhi_{k} = \{f: f(\bx) = R^\top \bx, R^\top R = I_k, R \in \reals^{d\times k}\}$, we show that $N_{\rm unif}(\BPhi_D, q, \delta) = O(d)$ and $  N_{\rm point}(\BPhi_D, q, \delta) = O(k)$ in Claim~\ref{claim: linear uniform} and~\ref{claim: linear point}.

We present the complete version of Theorem~\ref{lemma: guarantee of source nonlinear short} and~\ref{thm: nonlinear main short} as follows. 

\begin{theorem}[Full version of Theorem~\ref{lemma: guarantee of source nonlinear short}]
\label{lemma: guarantee of source nonlinear}  
Under Assumption~\ref{assumption: homogeneous input distribution},~\ref{assumption: nonlinear uniform},~\ref{assumption: trace assumption nonlinear}, and~\ref{assumption: diverse source tasks}, for some failure probability $\delta = o(1)$, we further assume $n_1 \gtrsim N_{\rm unif}(\BPhi_D, q, \delta)$, then the average excess risk across the source environments with probability $1-o(1)$ satisfies
\[
\begin{aligned}
    \overline{\rm ER}(\widehat{\phi}^E_k, \widehat{\btheta}^{1},\ldots,\widehat{\btheta}^{E} )&:= \frac{1}{E} \sum_{e=1}^{E} {\rm ER}_{e}(\widehat{\phi}^E_k, \widehat{\btheta}^{e})\lesssim \sigma\sqrt{\frac{\mathcal{C}_{\text{cont}}}{n_1 E}}+  \mathcal{C}_{\text{env}}    
\end{aligned}
\]
where
\[
\mathcal{C}_{\text{cont}} := \norm{\Sigma_{p}(\phi^*_{D[1:k]})}\lp\norm{\btheta^*}^2 + \Tr(\Lambda_{11})\rp, \quad
    \mathcal{C}_{\text{env}}:=\norm{\Sigma_{p}(\phi^{*e}_{D[k+1:D]})}\Tr(\Lambda_{22}).
\]
\end{theorem}

\begin{theorem}[Full version of Theorem~\ref{thm: nonlinear main short}]
    \label{thm: nonlinear main}
    Under Assumption~\ref{assumption: homogeneous input distribution},~\ref{assumption: nonlinear point},~\ref{assumption: nonlinear uniform},~\ref{assumption: trace assumption nonlinear}, and~\ref{assumption: diverse source tasks}, for some failure probability $\delta = o(1)$, we further assume $n_1 \gtrsim N_{\rm unif}(\BPhi_D, q, \delta)$, $n_2 \gtrsim N_{\rm point}(\BPhi_D, q, \delta)$. Under the choice of $\lambda_1$ and $\lambda_2$ in~\ref{eqn: choice of lambda nonlinear}, the excess risk of the learned predictor $\bx \mapsto \lag \widehat{\phi}^{E+1}_D (\bx), \widehat{\btheta}^{E+1} \rag$ in~\eqref{eqn: target} on the target domain with probability $1-o(1)$ satisfies
    \[
        \E_{\btheta^{*E+1}}\ls{\rm ER}_{E+1}(\widehat{\phi}^{E+1}_D, \widehat{\btheta}^{E+1})\rs\lesssim \sigma\sqrt{\frac{\overline{\mathcal{C}'}_{\text{env}}}{n_2}}  + \sigma\sqrt{\frac{ k}{n_2}} + \sigma\sqrt{\frac{ \text{RE} }{n_2}},
    \]
where 
\[
\overline{\mathcal{C}'}_{\text{env}} := \lp\norm{\btheta^*}^2 + \Tr(\Lambda_{11})+\Tr(\Lambda_{22}) \rp \max_{\phi \in \BPhi_{D-k}}\Tr(\Sigma_{p}(\phi)), \quad {\rm RE}:=\sigma\sqrt{\frac{\mathcal{C}_{\text{cont}}}{n_1 E}} + \mathcal{C}_{\text{env}}.
\]

\end{theorem}

\subsection{Representation divergence}
First, we introduce the definition of representation divergence, which is helpful in the analysis of the excess risk. Note that it is a generalization of the representation divergence defined in~\citep{DHK21} where the authors only consider the divergence between representations in the same function class.

\begin{definition}
    Let $q$ be a distribution over $\reals^d$ and $\BPhi_D$ be some function class with $\BPhi_{k} / \BPhi_{D-k}$ defined in~\eqref{eqn: induced low-dimensional function class}. For two representation functions $\phi \in \BPhi_{d_1}$, $\phi' \in \BPhi_{d_2}$, $d_1,d_2 \in \{D, k, D-k\}$, we define the divergence between $\phi$ and $\phi'$ with respect to $q$ as
    \[
    \begin{aligned}
        D_q(\phi,\phi') &= \Sigma_q(\phi',\phi') - \Sigma_q(\phi',\phi)(\Sigma_q(\phi,\phi))^{\dagger}\Sigma_q(\phi,\phi') \in \reals^{d_2 \times d_2}.
    \end{aligned}
    \]
\end{definition}
It can be verified that $D_q(\phi,\phi') \succeq 0$, $D_q(\phi,\phi')=0$ for any $\phi,\phi'$ and $q$. The following lemma states the relation between covariance and divergence between representations. Note that in~\citep{DHK21}, it holds for $\phi,\phi' \in \BPhi$ for some function class $\Phi$. It can be generalized to $\phi \in \BPhi_{d_1}$, $\phi' \in \BPhi_{d_2}$, $d_1,d_2 \in \{D, k, D-k\}$ using the same proof in~\citep{DHK21} due to the fact that $\BPhi_k$ and $\BPhi_{D-k}$ are induced by $\BPhi_D$. 

\begin{lemma}
\label{lemma: divergence and covariance}
    Given two representation functions $\phi \in \BPhi_{d_1}$, $\phi' \in \BPhi_{d_2}$, $d_1,d_2 \in \{D, k, D-k\}$, and two distributions $q,q'$ over $\reals^d$. If 
    \[
    S_q(\phi,\phi') \succeq \alpha \cdot S_{q'}(\phi,\phi') 
    \]
    then 
    \[
    D_q(\phi,\phi') \succeq \alpha \cdot D_{q'}(\phi,\phi'). 
    \]
\end{lemma}

\noindent \textit{Proof.  }    We will only show the case when $\phi \in \BPhi_k$ and $\phi' \in \BPhi_{D-k}$ since the technique is the same. 
For any $\bv \in \reals^{D-k}$, we will show that $\bv^\top D_q(\phi,\phi') \bv \ge \alpha \cdot \bv^\top D_{q'}(\phi,\phi') \bv$. We define the quadratic function $f(\bw):= [\bw^\top , -\bv^\top] S_q(\phi,\phi')[\bw^\top , -\bv^\top]^\top$. Using the definition of $S_q(\phi,\phi')$, we get 
\[
\begin{aligned}
    f(\bw) &= \bw^\top \Sigma_{q}(\phi,\phi)\bw - \bv^\top \Sigma_{q}(\phi',\phi)\bw - \bw^\top \Sigma_{q}(\phi,\phi')\bv + \bv^\top \Sigma_{q}(\phi',\phi')\bv\\
    &= \E_{\bx \sim q} \ls (\lag \phi(\bx), \bw \rag - \lag \phi'(\bx), \bv \rag)^2 \rs \ge 0.
\end{aligned}
\]
Note that $f(\bw)$ has maximized at $\bw^* = (\Sigma_q(\phi,\phi))^{\dagger}\Sigma_q(\phi,\phi') \bv$ 
\[
\min_{\bw \in \reals^{k}} f(\bw) = f(\bw^*) = \bv^\top D_{q}(\phi,\phi')\bv.
\]

Similarly, let $g(\bw):= [\bw^\top , -\bv^\top] S_{q'}(\phi,\phi')[\bw^\top , -\bv^\top]^\top$. We have 
\[
\min_{\bw \in \reals^{k}} g(\bw) = \bv^\top D_{q'}(\phi,\phi')\bv.
\]
Note that
\[
S_q(\phi,\phi') \succeq \alpha \cdot S_{q'}(\phi,\phi')
\]
implies 
$f(\bw) \ge \alpha g(\bw)$ for any $\bw \in \reals^k$. Recall that $f(\bw)$ is minimized at $\bw^*$, we have 
\[
\begin{aligned}
    \alpha \bv^\top D_{q'}(\phi,\phi')\bv = \alpha \min_{\bw \in \reals^{k}} g(\bw) \le \alpha  g(\bw^*) \le f(\bw^*) = \bv^\top D_{q}(\phi,\phi')\bv,
\end{aligned}
\]
which finishes the proof.
\hfill $\square$
\subsection{Proof of Theorem~\ref{lemma: guarantee of source nonlinear}}
By the definition of average excess risk across source environment,
\begin{equation}
\label{eqn: average excess risk}
   \overline{\rm ER}(\widehat{\phi}^E_k, \widehat{\btheta}^{1},\ldots,\widehat{\btheta}^{E} )= \frac{1}{E} \sum_{e=1}^{E} {\rm ER}_e(\widehat{\phi}^E_k, \widehat{\btheta}^{e}). 
\end{equation}
For $e \in [E]$, we define the empirical excess risk on $n_1$ i.i.d. samples $(\bx^{e}_1, y^e_1), \ldots, (\bx^{e}_{n_1}, y^e_{n_1})\sim \mu^{e}$ as 
\begin{equation}
\label{eqn: empirical excess risk}
    \widehat{\rm ER}(\widehat{\phi}^E_k, \widehat{\btheta}^{e}):= \frac{1}{2n_1}  \norm{\phi_D^*(X^e)\btheta^{*e} - \widehat{\phi}^{E}_k(X^e)\widehat{\btheta}^{e}}^2.
\end{equation}

Then each ${\rm ER}_e(\widehat{\phi}^E_k, \widehat{\btheta}^{e})$ can be upper bounded by its empirical counterpart via
\[
\begin{aligned}
    {\rm ER}_e(\widehat{\phi}^E_k, \widehat{\btheta}^{e}) 
      &= \frac{1}{2}\E_{\bx\sim p} [(\lag \phi^{*}_D(\bx), \btheta^{*E+1} \rag - \lag \widehat{\phi}^{E}_k(\bx), \widehat{\btheta}^{e} \rag)^2]\\
         &=\frac{1}{2} \begin{bmatrix}
             \widehat{\btheta}^{e}\\
             -\btheta^{*E+1} 
         \end{bmatrix}^\top S_{p} \lp\widehat{\phi}^{E}_k , \phi^{*}_D \rp \begin{bmatrix}
             \widehat{\btheta}^{e}\\
             -\btheta^{*E+1} 
         \end{bmatrix}\\
         &\lesssim \begin{bmatrix}
             \widehat{\btheta}^{e}\\
             -\btheta^{*E+1} 
         \end{bmatrix}^\top S_{\widehat{p}^e} \lp\widehat{\phi}^{E}_k , \phi^{*}_D \rp \begin{bmatrix}
             \widehat{\btheta}^{e}\\
             -\btheta^{*E+1} 
         \end{bmatrix}\\
         &= 2\widehat{\rm ER}_{e}(\widehat{\phi}^E_k, \widehat{\btheta}^{e}).
\end{aligned}
\]
Then to estimate~\eqref{eqn: average excess risk}, it suffices to upper bound the average empirical excess risk
\begin{equation}
\label{eqn: source empirical risk}
    \frac{1}{E} \sum_{e}^{E}  \widehat{\rm ER}_{e}(\widehat{\phi}^E_k, \widehat{\btheta}^{e}) = \frac{1}{2n_1 E}\sum_{e = 1}^E  \norm{\phi_D^*(X^e)\btheta^{*e} - \widehat{\phi}^{E}_k(X^e)\widehat{\btheta}^{e}}^2.
\end{equation}

By the optimality of $\widehat{\phi}_k, \widehat{\btheta}_1,\ldots, \widehat{\btheta}_E$, the empirical risk of the source environments satisfies
\[
\begin{aligned}
    &\frac{1}{2 n_1 E} \sum_{e=1}^E \norm{\by^e - \widehat{\phi}^{E}_k(X^e)\widehat{\btheta}^e}^2 \le \frac{1}{2 n_1 E} \sum_{e=1}^E \norm{\by^e - \phi^*_{D[1:k]}(X^e)\widehat{\btheta}_{[1:k]}^e}^2,
\end{aligned}
\]
which implies that \eqref{eqn: source empirical risk} can be decomposed into two terms
\begin{equation}
    \begin{aligned}
   \frac{1}{2n_1 E}\sum_{e = 1}^E  \norm{\phi_D^*(X^e)\btheta^{*e} - \widehat{\phi}^{E}_k(X^e)\widehat{\btheta}^{e}}^2 &\le  \underbrace{\frac{1}{n_1 E}\sum_{e = 1}^E \lag \widehat{\phi}^{E}_k(X^e)\widehat{\btheta}^{e} -   \phi_{D[1:k]}^*(X^e)\btheta^{*e}_{[1:k]}, \bz^e \rag}_{T_1}\\  &+  \underbrace{\frac{1}{2n_1 E} \sum_{e = 1}^E  \norm{\phi_{D[k+1:D]}^*(X^e)\btheta^{*e}_{D[k+1:D]}}^2}_{T_2}
\end{aligned}
\end{equation}
where we use 
\[
\by^e = \phi^*_{D[1:k]}(X^e)\btheta^{*e}_{[1:k]} + \phi^*_{D[k+1:D]}(X^e)\btheta^{*e}_{[k+1:D]} + \bz^e.
\]
\paragraph{Estimation of $T_1.$}

Using $\widehat{\phi}^{E}_k(X^e)\widehat{\btheta}^{e} = \mathcal{P}_{\widehat{\phi}^{E}_k(X^e)} \by^e$, each summand of $T_1$ can be decomposed into 
\[
\begin{aligned}
  \lag \widehat{\phi}^{E}_k(X^e)\widehat{\btheta}^{e} -   \phi_{D[1:k]}^*(X^e)\btheta^{*e}_{[1:k]}, \bz^e \rag &= \underbrace{ \lag -\mathcal{P}^{\perp}_{\widehat{\phi}^{E}_k(X^e)}\phi_{D[1:k]}^*(X^e)\btheta^{*e}_{[1:k]}, \bz^e \rag}_{T_{1,1,e}}\\
   &+\underbrace{\lag \mathcal{P}_{\widehat{\phi}^{E}_k(X^e)}\phi_{D[k+1:D]}^*(X^e)\btheta^{*e}_{[k+1:D]}, \bz^e \rag}_{T_{1,2,e}} \\
   &+ \underbrace{\lag \mathcal{P}_{\widehat{\phi}^{E}_k(X^e)}\bz^e, \bz^e \rag}_{T_{1,3,e}}
\end{aligned}
\]

Let $\bv^e = n_1^{-1/2}\phi_{D[1:k]}^*(X^e)\btheta^{*e}_{[1:k]}$ which is independent of $\bz^e$.
Then 
\[
\begin{aligned}
\frac{1}{n_1 E}\sum_{e = 1}^E T_{1,1,e} &=
      \frac{1}{n_1 E}\sum_{e = 1}^E\lag -\mathcal{P}^{\perp}_{\widehat{\phi}^{E}_k(X^e)}\phi_{D[1:k]}^*(X^e)\btheta^{*e}_{[1:k]}, \bz^e \rag \\ &\le \frac{1}{\sqrt{n_1 E}}\sum_{e = 1}^E\lag \frac{1}{\sqrt{n_1}}\phi_{D[1:k]}^*(X^e)\btheta^{*e}_{[1:k]}, \frac{1}{\sqrt{E}}\bz^e \rag\\
    &= \frac{1}{\sqrt{n_1 E}}\sum_{e = 1}^E  \lag \bv^e, \frac{1}{\sqrt{E}}\bz^e \rag \\
    &\lesssim  \frac{1}{\sqrt{n_1 E}}\sqrt{\frac{\sigma^2 \sum_{e=1}^E \norm{\bv^{e}}^2}{E}}\\
\end{aligned}
\]
where the last inequality is obtained via Bernstein's inequality. Note that we estimate $\norm{\bv^e}^2$ via 
\[
\begin{aligned}
    \norm{\bv^e}^2 &= \norm{n_1^{-1/2}\phi_{D[1:k]}^*(X^e)\btheta^{*e}_{[1:k]}}^2\\
    &= \norm{\Sigma^{1/2}_{\widehat{p}}(\phi^*_{D[1:k]})\btheta^{*e}_{[1:k]}}^2\\
    &\le \norm{\Sigma_{\widehat{p}}(\phi^*_{D[1:k]})}\lp\norm{\btheta^*}^2 + \Tr(\Lambda_{11})\rp\\
    & \lesssim \norm{\Sigma_{p}(\phi^*_{D[1:k]})}\lp\norm{\btheta^*}^2 + \Tr(\Lambda_{11})\rp
\end{aligned}
\]
where we use $\btheta^{*e}_{[1:k]} \sim \mathcal{N}(\btheta^*, \Lambda_{11})$ and Assumption~\ref{assumption: nonlinear uniform}.
Thus,
\[
 \frac{1}{n_1 E}\sum_{e = 1}^E T_{1,1,e} \lesssim \sigma\sqrt{\frac{\norm{\Sigma_{p}(\phi^*_{D[1:k]})}\lp\norm{\btheta^*}^2 + \Tr(\Lambda_{11})\rp}{n_1 E}}.
\]

Similarly, 
\[
\begin{aligned}
    \frac{1}{n_1 E}\sum_{e = 1}^E T_{1,2,e} \lesssim \sigma\sqrt{\frac{\norm{\Sigma_{p}(\phi^*_{D[k+1:D]})} \sum_{j=D-k+1}^D \lp \Lambda_{22} \rp_{j}}{n_1 E}}
\end{aligned}
\]
where $\sum_{j=D-k+1}^D \lp \Lambda_{22} \rp_{j}$ is the sum of top $k$ entries in $\Lambda_{22}$,
and 
\[
\frac{1}{n_1 E}\sum_{e = 1}^E T_{1,3,e} \lesssim \frac{\sigma^2 k}{n_1}.
\]
Then we obtain the estimation of $T_1$
\begin{equation}
    \begin{aligned}
        T_1 &= \frac{1}{n_1 E}\sum_{e = 1}^E \lp T_{1,1,e} + T_{1,2,e} + T_{1,3,e} \rp\\
        &\lesssim \sigma\sqrt{\frac{\norm{\Sigma_{p}(\phi^*_{D[1:k]})}\lp\norm{\btheta^*}^2 + \Tr(\Lambda_{11})\rp}{n_1 E}}+ \sigma\sqrt{\frac{\norm{\Sigma_{p}(\phi^*_{D[k+1:D]})} \sum_{j=D-k+1}^D \lp \Lambda_{22} \rp_{j}}{n_1 E}} + \frac{\sigma^2 k}{n_1}\\
        &\lesssim \sigma\sqrt{\frac{\norm{\Sigma_{p}(\phi^*_{D[1:k]})}\lp\norm{\btheta^*}^2 + \Tr(\Lambda_{11})\rp}{n_1 E}}\\
        &= \sigma\sqrt{\frac{\mathcal{C}_{\text{cont}}}{n_1 E}}
    \end{aligned}
\end{equation}
where we use Assumption~\ref{assumption: trace assumption nonlinear} and omit the third term which is of small order in the last inequality. 

\paragraph{Estimation of $T_2.$} Note that $T_2$ is the non-vanishing term as $n_1, E \rightarrow \infty$ because we only learn the content features from source environments. This term is the approximation error due to the environmental features, thus it can be bounded by the complexity of the environmental features.
\begin{equation}
\begin{aligned}
T_2 &=  \frac{1}{2n_1 E} \sum_{e = 1}^E \norm{\phi_{D[k+1:D]}^*(X^e)^{*e}\btheta^{*e}_{[k+1:D]}}^2 \\
&=\frac{1}{E} \sum_{e = 1}^E \norm{\Sigma^{1/2}_{\widehat{p}}(\phi^{*e}_{D[k+1:D]})\btheta^{*e}_{[k+1:D]}}^2\\
&\lesssim \norm{\Sigma_{p}(\phi^{*e}_{D[k+1:D]})}\Tr(\Lambda_{22})\\
&= \mathcal{C}_{\text{env}}
\end{aligned}
\end{equation}

Thus, with probability $1-o(1)$, the average excess risk across source environments satisfies
\[
\begin{aligned}
    \overline{\rm ER}(\widehat{\phi}^E_k, \widehat{\btheta}^{1},\ldots,\widehat{\btheta}^{E} )= \frac{1}{E} \sum_{e=1}^{E} {\rm ER}_e(\widehat{\phi}^E_k, \widehat{\btheta}^{e})
    \lesssim T_1 + T_2 \lesssim \sigma\sqrt{\frac{\mathcal{C}_{\text{cont}}}{n_1 E}} + \mathcal{C}_{\text{env}}.
\end{aligned}      
\]

\subsection{Proof of Theorem~\ref{thm: nonlinear main}}
In this section, we abbreviate $ {\rm ER}_{E+1}(\widehat{\phi}^{E+1}_D, \widehat{\btheta}^{E+1})$ as $ {\rm ER}(\widehat{\phi}^{E+1}_D, \widehat{\btheta}^{E+1})$ and $ \widehat{\rm ER}_{E+1}(\widehat{\phi}^{E+1}_D, \widehat{\btheta}^{E+1})$ as $ \widehat{\rm ER}(\widehat{\phi}^{E+1}_D, \widehat{\btheta}^{E+1})$.
We will first bound the empirical excess risk of $\bx \mapsto \lag \widehat{\phi}^{E+1}_D(\bx),\widehat{\btheta}^{E+1} \rag$ and then prove that it is close to its population counterpart.  By the optimality of $(\widehat{\phi}^{E+1}_D, \widehat{\btheta}^{E+1})$ for~\eqref{eqn: target}, the empirical risk satisfies
\begin{equation}
\label{eqn: target basic inequality}
    \begin{aligned}  
    & \frac{1}{2n_2}  \norm{\by^{E+1} - \widehat{\phi}^{E+1}_D(X^{E+1})\widehat{\btheta}^{E+1}}^2\\
    & \le \frac{1}{2n_2}  \norm{\by^{E+1} - \widehat{\phi}^{E+1}_D(X^{E+1})\widehat{\btheta}^{E+1}}^2 + \frac{\lambda_1}{n_2}\norm{\mathcal{P}_{\widehat{\phi}^{E}_k(X^{E+1})}^{\perp} \widehat{\phi}^{E+1}_D(X^{E+1})\widehat{\btheta}^{E+1}}^2 +\frac{\lambda_2}{2} \norm{\widehat{\btheta}^{E+1}}^2\\ & \le \frac{1}{2n_2}  \norm{\by^{E+1} - \phi^*_D(X^{E+1})\theta^{*E+1}}^2 + \frac{\lambda_1}{n_2}\norm{\mathcal{P}_{\widehat{\phi}^{E}_k(X^{E+1})}^{\perp} \phi^*_D(X^{E+1})\btheta^{*E+1}}^2 +\frac{\lambda_2}{2} \norm{\btheta^{*E+1}}^2.
\end{aligned}
\end{equation}
Let $A = \ls \phi^*_{D[1:k]}(X^{E+1}),  \widehat{\phi}^{E}_k(X^{E+1}) \rs$. By plugging $\by^{E+1} = \phi^*_D(X^{E+1})\btheta^{*E+1} + \bz^{E+1}$ in~\eqref{eqn: target basic inequality}, the empirical excess risk on the target task satisfies
\begin{equation}
    \begin{aligned}
    \widehat{\rm ER}(\widehat{\phi}^{E+1}_D, \widehat{\btheta}^{E+1})
    &=\frac{1}{2n_2}  \norm{\phi^*_D(X^{E+1})\theta^{*E+1} - \widehat{\phi}^{E+1}_D(X^{E+1})\widehat{\btheta}^{E+1}}^2\\
        &\le -\frac{1}{n_2} \left \langle \bz^{E+1},  \phi^*_D(X^{E+1})\theta^{*E+1} - \widehat{\phi}^{E+1}_D(X^{E+1})\widehat{\btheta}^{E+1}\right \rangle\\
        & + \frac{\lambda_1}{n_2}\norm{\mathcal{P}_{\widehat{\phi}^{E}_k(X^{E+1})}^{\perp} \phi^*_D(X^{E+1})\btheta^{*E+1}}^2 + \frac{\lambda_2}{2} \norm{\btheta^{*E+1}}^2\\
        &= -\frac{1}{n_2} \left \langle \bz^{E+1},  \mathcal{P}_{A}\lp \phi^*_D(X^{E+1})\theta^{*E+1} - \widehat{\phi}^{E+1}_D(X^{E+1})\widehat{\btheta}^{E+1}\rp \right \rangle\\
        &-\frac{1}{n_2} \left \langle \bz^{E+1},  \mathcal{P}^{\perp}_{A}\lp \phi^*_D(X^{E+1})\theta^{*E+1} - \widehat{\phi}^{E+1}_D(X^{E+1})\widehat{\btheta}^{E+1}\rp \right \rangle\\
        & + \frac{\lambda_1}{n_2}\norm{\mathcal{P}_{\widehat{\phi}^{E}_k(X^{E+1})}^{\perp} \phi^*_D(X^{E+1})\btheta^{*E+1}}^2 + \frac{\lambda_2}{2} \norm{\btheta^{*E+1}}^2\\
        &\le \underbrace{\frac{1}{n_2}\norm{\mathcal{P}_{A}\bz^{E+1}}\norm{\phi^*_D(X^{E+1})\theta^{*E+1} - \widehat{\phi}^{E+1}_D(X^{E+1})\widehat{\btheta}^{E+1}}}_{T_1}\\
        &+ \underbrace{\frac{1}{n_2} \left \langle \bz^{E+1},  \mathcal{P}^{\perp}_{A} \phi^*_D(X^{E+1})\theta^{*E+1} \right \rangle}_{T_2}\\ &+ \underbrace{\frac{1}{n_2}\left \langle \bz^{E+1}, \mathcal{P}^{\perp}_{A}\widehat{\phi}^{E+1}_D(X^{E+1})\widehat{\btheta}^{E+1}  \right \rangle}_{T_3}\\
        &+ \underbrace{\frac{\lambda_1}{n_2}\norm{\mathcal{P}_{\widehat{\phi}^{E}_k(X^{E+1})}^{\perp} \phi^*_D(X^{E+1})\btheta^{*E+1}}^2 + \frac{\lambda_2}{2} \norm{\btheta^{*E+1}}^2}_{T_4}\\
    \end{aligned}
\end{equation}

We will bound the $4$ terms one by one. 
\paragraph{Estimation of $T_1.$} Note that 
\[
    \norm{\mathcal{P}_{A}\bz^{E+1}} \lesssim \sigma \sqrt{k},
\]
then 
\[
T_1 \lesssim \frac{\sigma\sqrt{k}}{n_2}\norm{\phi^*_D(X^{E+1})\theta^{*E+1} - \widehat{\phi}^{E+1}_D(X^{E+1})\widehat{\btheta}^{E+1}} = 2\sigma\sqrt{\frac{k\widehat{\rm ER}(\widehat{\phi}^{E+1}_D, \widehat{\btheta}^{E+1})}{n_2}}.
\]
\paragraph{Estimation of $T_2.$}
\[
\begin{aligned}
    T_2 = \frac{1}{n_2} \left \langle \bz^{E+1},  \mathcal{P}^{\perp\top}_{A} \phi^*_D(X^{E+1})\btheta^{*E+1} \right \rangle &= \frac{1}{n_2} \norm{\phi^*_D(X^{E+1})^{\top}\mathcal{P}^{\perp\top}_{A} \bz^{E+1}} \norm{\btheta^{*E+1}}\\
    &= \frac{1}{n_2} \norm{\phi^*_{D[k+1:D]}(X^{E+1})^{\top}\mathcal{P}^{\perp \top}_{A} \bz^{E+1}} \norm{\btheta^{*E+1}}\\
    &\lesssim \sigma\sqrt{\frac{\Tr(\Sigma_{p}(\phi^*_{D[k+1:D]}))\lp \norm{\btheta^*}^2 + \Tr(\Lambda_{11})+\Tr(\Lambda_{22}) \rp}{n_2}}\\
    &= \lesssim \sigma\sqrt{\frac{\mathcal{C}'_{\text{env}}}{n_2}}
\end{aligned}
\] where we use Assumption~\ref{assumption: trace assumption nonlinear} and the fact that $A \perp \phi^*_{D[1:k]}(X^{E+1})$ and define $\mathcal{C}'_{\text{env}} := \Tr(\Sigma_{p}(\phi^*_{D[k+1:D]}))\lp \norm{\btheta^*}^2 + \Tr(\Lambda_{11})+\Tr(\Lambda_{22}) \rp$.
\paragraph{Estimation of $T_3.$} 
Similarly to $T_2$, $T_3$ is bounded via
\[
\begin{aligned}
    T_3 &= \frac{1}{n_2}\left \langle \bz^{E+1}, \mathcal{P}^{\perp}_{A}\widehat{\phi}^{E+1}_D(X^{E+1})\widehat{\btheta}^{E+1}  \right \rangle\\ &\le \frac{1}{n_2} \norm{\widehat{\phi}^{E+1}_D(X^{E+1})^{\top}\mathcal{P}^{\perp\top}_{A}\bz^{E+1}}\norm{\widehat{\btheta}^{E+1}}\\
    &\lesssim \sigma\sqrt{\frac{\Tr(\Sigma_{p}(\widehat{\phi}^{E+1}_D))\lp \norm{\btheta^*}^2 + \Tr(\Lambda_{11})+\Tr(\Lambda_{22}) \rp}{n_2}}\\
    &\le \sigma\sqrt{\frac{\overline{\mathcal{C}'}_{\text{env}}}{n_2}}
\end{aligned}
\]
where $\overline{\mathcal{C}}_{\text{env}}$ is the maximum complexity of the environment features defined as
\[
\overline{\mathcal{C}'}_{\text{env}} := \lp\norm{\btheta^*}^2 + \Tr(\Lambda_{11})+\Tr(\Lambda_{22}) \rp \max_{\phi \in \BPhi_{D-k}}\Tr(\Sigma_{p}(\phi)).
\]
Thus $T3 \gtrsim T_2$ by definition.

\paragraph{Estimation of $T_4.$}
The following lemma estimates $n_2^{-1}\norm{\mathcal{P}_{\widehat{\phi}^{E}_k(X^{E+1})}^{\perp} \phi^*_D(X^{E+1})\btheta^{*E+1}}^2$.
\begin{lemma}
\label{lemma: feasibility of ground-truth}
Under the conditions in Theorem~\ref{thm: nonlinear main}, it holds with probability $1-o(1)$ that
    \[
    \frac{1}{n_2}\norm{\mathcal{P}_{\widehat{\phi}^{E}_k(X^{E+1})}^{\perp} \phi^*_D(X^{E+1})\btheta^{*E+1}}^2 \lesssim \underbrace{\sigma\sqrt{\frac{\mathcal{C}_{\text{cont}}}{n_1 E}} + \mathcal{C}_{\text{env}}}_{:= \text{RE}}.
    \]
\end{lemma}
Then $T_4$ is bounded via
\[
\begin{aligned}
    T_4 &= \frac{\lambda_1}{n_2}\norm{\mathcal{P}_{\widehat{\phi}^{E}_k(X^{E+1})}^{\perp} \phi^*_D(X^{E+1})\btheta^{*E+1}}^2 + \frac{\lambda_2}{2} \norm{\btheta^{*E+1}}^2\\
    &\lesssim \lambda_1 \text{RE} + \lambda_2 (\norm{\btheta^*}^2 + \Tr(\Lambda_{11}) + \Tr(\Lambda_{22}) ).
\end{aligned}
\]

Combining the above $4$ terms, and under the choice 
\begin{align}\label{eqn: choice of lambda nonlinear}
    \lambda_1 = \sqrt{\frac{\sigma^2}{n_2 \text{RE}}}, \quad \lambda_2 = \frac{\sigma\sqrt{\text{RE}}}{\sqrt{n_2}(\norm{\btheta^*}^2 + \Tr(\Lambda_{11}) + \Tr(\Lambda_{22}))},
\end{align}
we have the following quadratic inequality
\begin{equation} \label{eqn: nonlinear target emperical}
    \begin{aligned}
    \widehat{\rm ER}(\widehat{\phi}^{E+1}_D, \widehat{\btheta}^{E+1})& \lesssim 2\sigma\sqrt{\frac{k\widehat{\rm ER}(\widehat{\phi}^{E+1}_D, \widehat{\btheta}^{E+1})}{n_2}} + \sigma\sqrt{\frac{\overline{\mathcal{C}'}_{\text{env}}}{n_2}}\\
    &+ \lambda_1 \text{RE} +  \lambda_2 (\norm{\btheta^*}^2 + \Tr(\Lambda_{11}) + \Tr(\Lambda_{22}) )\\
    &\lesssim 2\sigma\sqrt{\frac{k\widehat{\rm ER}(\widehat{\phi}^{E+1}_D, \widehat{\btheta}^{E+1})}{n_2}} + \sigma\sqrt{\frac{\overline{\mathcal{C}'}_{\text{env}}}{n_2}} + \sigma\sqrt{\frac{ \text{RE} }{n_2}}, 
\end{aligned}
\end{equation}

which gives the solution
\[
\widehat{\rm ER}(\widehat{\phi}^{E+1}_D, \widehat{\btheta}^{E+1})\lesssim \sigma\sqrt{\frac{\overline{\mathcal{C}'}_{\text{env}}}{n_2}}  + \sigma\sqrt{\frac{ k}{n_2}} + \sigma\sqrt{\frac{ \text{RE} }{n_2}}.
\]

Finally, we will prove that the empirical excess risk is close to its population counterpart
\begin{equation}
    \begin{aligned}
         {\rm ER}(\widehat{\phi}^{E+1}_D, \widehat{\btheta}^{E+1}) &= \frac{1}{2}\E_{\bx\sim p} [(\lag \phi^{*}_D(\bx), \btheta^{*E+1} \rag - \lag \widehat{\phi}^{E+1}_D(\bx), \widehat{\btheta}^{E+1} \rag)^2]\\
         &=\frac{1}{2} \begin{bmatrix}
             \widehat{\btheta}^{E+1}\\
             -\btheta^{*E+1} 
         \end{bmatrix}^\top S_{p} \lp\widehat{\phi}^{E+1}_D , \phi^{*}_D \rp \begin{bmatrix}
             \widehat{\btheta}^{E+1}\\
             -\btheta^{*E+1} 
         \end{bmatrix}\\
         &\lesssim\begin{bmatrix}
             \widehat{\btheta}^{E+1}\\
             -\btheta^{*E+1} 
         \end{bmatrix}^\top  S_{\widehat{p}^{E+1}} \lp\widehat{\phi}^{E+1}_D , \phi^{*}_D \rp \begin{bmatrix}
             \widehat{\btheta}^{E+1}\\
             -\btheta^{*E+1} 
         \end{bmatrix}\\
         &= 2 \widehat{\rm ER}(\widehat{\phi}^{E+1}_D, \widehat{\btheta}^{E+1})\\
         &\lesssim \sigma\sqrt{\frac{\overline{\mathcal{C}'}_{\text{env}}}{n_2}}  + \sigma\sqrt{\frac{ k}{n_2}} + \sigma\sqrt{\frac{ \text{RE} }{n_2}}.
    \end{aligned}
\end{equation}

\subsection{Proof of Lemma~\ref{lemma: feasibility of ground-truth}}
Note that we have 
\[
    \frac{1}{n_2}\norm{\mathcal{P}_{\widehat{\phi}^{E}_k(X^{E+1})}^{\perp} \phi^*_D(X^{E+1})\btheta^{*E+1}}^2 
    \le \frac{1}{n_2}\norm{\mathcal{P}_{\widehat{\phi}^{E}_k(X^{E+1})}^{\perp} \phi^*_D(X^{E+1})}^2_F\norm{\btheta^{*E+1}}^2 .
\]
Let $\Theta^* = [\btheta^{*1}, \ldots, \btheta^{*E}]$ and $\sigma_1(\Theta^*)$ be its smallest singular value. We have the following chain of inequalities
\[
\begin{aligned}
    &\frac{1}{n_2}\norm{\mathcal{P}_{\widehat{\phi}^{E}_k(X^{E+1})}^{\perp} \phi^*_D(X^{E+1})}^2_F \frac{\sigma^2_{1}(\Theta^*)}{E}\\ &\le \frac{1}{E} \Tr(D_{\widehat{p}^{E+1}} (\widehat{\phi}^{E}_k,\phi^*_D)) \sigma^2_{1}(\Theta^*)\\ 
    &\lesssim \frac{1}{E}\Tr(D_{p} (\widehat{\phi}^{E}_k,\phi^*_D)) \sigma^2_{1}(\Theta^*)\\
    &\le \frac{1}{E}\norm{\lp D_{p} (\widehat{\phi}^{E}_k,\phi^*_D)\rp^{1/2}\Theta^*}^2_F\\
    &= \frac{1}{E}\sum_{e=1}^E \btheta^{*e\top}D_{p} (\widehat{\phi}^{E}_k,\phi^*_D) \btheta^{*e}\\
    &\lesssim \frac{1}{E}\sum_{e=1}^E \btheta^{*e\top}D_{\widehat{p}^e} (\widehat{\phi}^{E}_k,\phi^*_D) \btheta^{*e}\\
    &= \frac{1}{n_1E}\sum_{e=1}^E \btheta^{*e\top}(\phi^*_D(X^e))^{\top} \lp I_{n_1} - \widehat{\phi}^{E}_k(X^e) \ls(\widehat{\phi}^{E}_k(X^e))^{\top}\widehat{\phi}^{E}_k(X^e)\rs^\dagger \widehat{\phi}^{E}_k(X^e) \rp\phi^*_D(X^e))\btheta^{*e}\\
    &= \frac{1}{n_1 E} \sum_{e=1}^E \norm{\mathcal{P}^{\perp}_{\widehat{\phi}(X^e)}\phi^*_D(X^e)\btheta^{*e}}^2\\
    &\lesssim  \frac{1}{2n_1 E}\sum_{e = 1}^E \norm{\phi_D^*(X^e)\btheta^{*e} - \widehat{\phi}^{E}_k(X^e)\widehat{\btheta}^{e}}^2\\
       &\lesssim \underbrace{\sigma\sqrt{\frac{\mathcal{C}_{\text{cont}}}{n_1 E}} + \mathcal{C}_{\text{env}}}_{\text{RE}}.
\end{aligned}
\]
where we apply Assumption~\ref{assumption: nonlinear point} and~\ref{assumption: nonlinear uniform} and~\eqref{eqn: empirical excess risk}. Then using the gaussianality of $\btheta^{*E+1}$, we get 
\[
\begin{aligned}
    \frac{1}{n_2}\norm{\mathcal{P}_{\widehat{\phi}^{E}_k(X^{E+1})}^{\perp} \phi^*_D(X^{E+1})\btheta^{*E+1}}^2 
    &\le \frac{1}{n_2}\norm{\mathcal{P}_{\widehat{\phi}^{E}_k(X^{E+1})}^{\perp} \phi^*_D(X^{E+1})}^2_F\norm{\btheta^{*E+1}}^2\\
    &\lesssim \frac{E (\norm{\btheta^*}^2 \Tr(\Lambda_{11}) + \Tr(\Lambda_{22}))}{\sigma^2_{1}(\Theta^*)} \text{RE}.
\end{aligned}
\]
Finally, under Assumption~\ref{assumption: trace assumption nonlinear} and~\ref{assumption: diverse source tasks}, 
\[
\begin{aligned}
    \frac{1}{n_2}\norm{\mathcal{P}_{\widehat{\phi}^{E}_k(X^{E+1})}^{\perp} \phi^*_D(X^{E+1})\btheta^{*E+1}}^2  \lesssim \frac{E (\norm{\btheta^*}^2 \Tr(\Lambda_{11}) + \Tr(\Lambda_{22}))}{\sigma^2_{1}(\Theta^*)} \text{RE}\lesssim \text{RE}.
\end{aligned}
\]

\section{Linear representations}
\label{sec: linear}
In the linear case, the ground-truth representation function 
class is 
\[
\BPhi_{d} = \{f: f(\bx) = R^\top \bx, R \in \mathcal{O}(d)\}.
\]
In particular, we let the ground-truth representation function be $\phi^*_d(\bx) = R^{^*\top}\bx$ where $R^{^*} \in \mathcal{O}(d)$. Then the data generation process $(\bx, y) \sim \mu_e$ for $e \in [E+1]$ can be described as 
\[
    y = \lag \bx, R^*\btheta^{e*} \rag + z.
\]
Given the meta distribution in~\eqref{eqn: meta distribution}, $R^*\btheta^{e*}$'s are i.i.d. rotated multivariate Gaussian random variables
\[
\begin{aligned}
    R^*\btheta^{e*} \sim \mathcal{N} \left( R^*_1\btheta^*,\Sigma_{\text{RM}} \right),\\ \Sigma_{\text{RM}} = R^*_1\Lambda_{11}R^{*\top}_1 +  R^*_2\Lambda_{22}R^{*\top}_2,
\end{aligned}
\]
where $R^* = [R^*_1 , R^*_2]$, $R^*_1$ is the first $k$ columns of $R^*$ and $R^*_2$ is the rest $(d-k)$ columns. The ``RM'' here stands for ``rotated meta distribution''.

\subsection{The Meta-Representation Learning Algorithm}

Our goal is to learn a low-dimensional representation 
\[
\widehat{\phi}^E_k \in \BPhi_k = \{f: f(\bx) = R^\top \bx, R^\top R = I_k, R \in \reals^{d\times k}\}
\]
from the source environments that capture the content features. In the linear case, it is equivalent to finding a $d \times k$ matrix with orthogonal columns to approximate $R^*_1$. If $E$ is large enough, \ie we have enough source environments, a natural way of estimating $R^*_1$ is to use the least $k$ eigenvectors of the sample covariance matrix of $\widehat{\btheta}^{e}$'s. This motivates the following learning process.

For each source environment $e \in [E]$, we obtain $\widehat{\btheta}^{e}$ via empirical risk minimization 
\begin{equation}
    \widehat{\btheta}^{e} =  \underset{\btheta^{e} \in \reals^{d}}{\argmin} \frac{1}{2n_1} \norm{\by^e - X^e \btheta^e}^2.
\end{equation}
Let $\bar{\btheta}^E = E^{-1}\sum_{e=1}^E \widehat{\btheta}^e$
 be the ``sample mean'' of these learned parameters and consider the ``sample covariance'' with its eigendecomposition
\begin{equation}
    \begin{aligned}
\label{eqn: eigendecomposition of sample cocariance}
    \Sigma_{\widehat{\btheta}} &= \frac{1}{E} \sum_{e=1}^E  \lp\widehat{\btheta}^e - \bar{\btheta}^E\rp \lp\widehat{\btheta}^e - \bar{\btheta}^E\rp^\top\\
    &= \widehat{R}_1 \widehat{\Lambda}_{11} \widehat{R}_1^\top + \widehat{R}_2\widehat{\Lambda}_{22}\widehat{R}_2^\top
\end{aligned} 
\end{equation}
where $\widehat{\Lambda}_{11} \in \reals^{k \times k}$ and $\widehat{\Lambda}_{22} \in \reals^{(d-k) \times (d-k)}$ are diagonal matrices with ascending entries, $\widehat{\Lambda}_{11}$ consists of the least $k$ eigenvalues of $\Sigma_{\widehat{\btheta}}$ with eigenvectors $\widehat{R}_1 \in \reals^{d\times k}$ and $\widehat{\Lambda}_{22}$ consists of the remaining $d-k$ eigenvalues with eigenvectors $\widehat{R}_2 \in \reals^{d\times (d-k)}$.
 
The learned representation $\widehat{R}_1$ and the average parameter $\bar{\btheta}^E$ will be applied to the fine-tunning phase on the target environment via
\begin{equation}
\begin{aligned}
    \label{eqn: linear fine-tune}
    \widehat{\btheta}^{E+1} &= \underset{\theta^{E+1} \in \mathbb{R}^d}{\argmin} \frac{1}{2n_2} \norm{\by^{E+1} - X^{E+1} \btheta^{E+1}}^2\\ &+ \underbrace{\frac{\lambda_1}{2} \norm{\mathcal{P}_{\widehat{R}_1}(\btheta^{E+1} - \bar{\btheta}^E)}^2}_{{\rm Reg}_1} + \underbrace{\frac{\lambda_2}{2} \norm{\mathcal{P}^\perp_{\widehat{R}_1}(\btheta^{E+1}) }^2}_{{\rm Reg}_2}.
\end{aligned}
\end{equation}
This optimization is the empirical risk minimization on the target domain with two regularizing terms. ${\rm Reg}_1$ penalizes the difference between $\btheta - \bar{\btheta}^E$ on the subspace spanned by the column space of the learned representation $\widehat{R}_1$. This regularization encourages the learning dynamics to capture more content features approximately shared through all the environments. On the other hand, ${\rm Reg}_2$ penalizes the weight in the directions that are perpendicular to the column space of $\widehat{R}_1$, which discourages the learning dynamics from capturing the environmental features. This regularization-based fine-tuning process is motivated by the biased regularization~\citep{DPC18, DPC20} which has been widely applied to the theoretical analysis of transfer learning. 


We are interested in the excess risk of the learned predictor $\bx \mapsto \lag \bx, \widehat{\btheta}^{E+1} \rag$ on the target environment, \ie how much our learned model performs worse than the optimal model on the target task:
\begin{align}
    {\rm ER}(\widehat{\btheta}^{E+1}) = \frac{1}{2}\E_{(\bx,y)\sim \mu_{E+1}} [\lag \bx, R^*\btheta^{*E+1} -  \widehat{\btheta}^{E+1}\rag^2].
\end{align}
We often calculate the expected excess risk with respect to the meta distribution, \ie $\E_{\btheta^{*E+1}}[{\rm ER}(\widehat{\btheta}^{E+1})]$.
\subsection{Theoretical Analysis}
\label{sec: main linear}
Before stating the main theorem, we first make some statistical assumptions on the input data. For $e \in [E+1]$, we assume $\E_{\bx \sim p_e} [\bx] = \bzero$ and let $ \Sigma_{X^e} = \E_{\bx \sim p_e} [\bx \bx^\top] $. Note that a sample $\bx \sim p_e$ can be generated from $\bx = \Sigma^{1/2}_{X^e} \bar{\bx}$ where $\bar{\bx} \sim \bar{p}_e$ where $\E_{\bx \sim \bar{p}_e} [\bar{\bx}] = \bzero$ and $\E_{\bx \sim \bar{p}_e} [\bar{\bx}\bar{\bx}^\top] = I_d$ ($p_e$ is called the whitening of $p_e$). We make the following assumptions on the input distribution $p_1, \ldots, p_{E+1}$.
\begin{assumption}[Subgaussian input]
\label{assumption: subgaussian input}
    There exists $\rho >0$ such that for $e \in [E+1]$, $\bar{\bx} \sim \bar{p}_e$ is $\rho^2$-subgaussian.
\end{assumption}
\begin{assumption}[Covariance dominance]
\label{assumption: covariance dominance}
    There exists $c>0, c_2 > c_1 >0$ and $\Sigma_X \succeq 0$ such that for $e\in[E]$, $c_1 \cdot \Sigma_X \preceq c \cdot \Sigma_{X^{E+1}} \preceq \Sigma_{X^{e}} \preceq c_2 \cdot \Sigma_X$.
\end{assumption}

Assumption~\ref{assumption: subgaussian input} is a standard assumption in statistical learning to obtain probabilistic tail bounds used in the proof. It might be replaced with other moment or boundedness conditions if we use different tail bounds in the analysis.

Assumption~\ref{assumption: covariance dominance} says that every direction spanned by $\Sigma_{E+1}$ should be spanned by $\Sigma_e$, $e \in [E]$ and the parameter $c$ quantifies how ``easy'' it is for $\Sigma_e$ to cover $\Sigma_{E+1}$. We remark that instead of having $c \cdot \Sigma_{X^{E+1}} \preceq \Sigma_{X^{e}}$ for all $e\in[E]$, as long as this holds for a constant fraction of $[E]$, our result is valid. We also assume that $\Sigma_{X^e}$'s are uniformly bounded by some PSD matrix $\Sigma_X$ up to some constant, which facilitates our theoretical analysis.

Recall that we distinguish the content and environmental features based on the variation of their correlation with $y$. The following assumption guarantees the well-separation between these two features. Let $\Delta_{\Lambda}:= \min(\Lambda_{22}) - \norm{\Lambda_{11}}$ be the eigengap between environmental and content features.

\begin{assumption}[Well-speration]
\label{assumption: eigengap}
\[
\norm{\widehat{\Sigma}_{\widehat{\btheta}}-\Sigma_{{\rm RM}}} \lesssim \Delta_{\Lambda}  .
\]
\end{assumption}
This assumption indicates the eigengap between the content feature space and environmental feature space in the covariance matrix $\Sigma_{{\rm RM}}$ is large enough so that the Davis-Kahan bound $\min_{O \in \mathcal{O}(k)}\norm{\widehat{R}_1 - R_1^* O}$ is meaningful. One sufficient condition for the separation assumption is $E$ and $n_1$ being sufficiently large: $    \sqrt{\frac{d}{E}}\norm{\Lambda_{22}} + \frac{\sigma^2}{n_1\lambda_{1}(\Sigma_X)} \lesssim \Delta_{\Lambda} $.

\begin{assumption}[Content features dominance]
\label{assumption: trace assumption}

\[
 \frac{\Tr\lp R^{*\top}_1  \Sigma_X R^*_1  \rp}{\Tr\lp  R^{*\top}_2\Sigma_X R^*_2 \rp} \gtrsim \frac{\norm{\Lambda_{22}}}{\norm{\Lambda_{11}}}.
\] 
\end{assumption}
\begin{remark}
    In the liner case,~\eqref{eqn: data generation component} can be written as
    \[
    y^e = \bx^\top R^*_1\btheta^{*e}_{[1:k]} + \bx^\top R^*_2\btheta^{*e}_{[k+1:D]} + z.
    \]

    Assumption~\ref{assumption: trace assumption} gaurentees 
    \[
    \E_{\theta^{*e},(\bx,y)} \left \lvert   \bx^\top R^*_1\btheta^{*e}_{[1:k]} \right \rvert  \gtrsim  \E_{\theta^{*e},(\bx,y)} \left \lvert \bx^\top R^*_2\btheta^{*e}_{[k+1:D]} \right \rvert .
    \]
\end{remark}

The following theorem gives a high probability bound of the excess risk for linear representations on the target domain. 

\begin{theorem}
\label{thm: linear main full}
Under Assumption~\ref{assumption: subgaussian input},~\ref{assumption: covariance dominance},~\ref{assumption: eigengap} and~\ref{assumption: trace assumption}, we further assume that $k \le d \le E$ and the sample size in source and target environments satisfies $n_1 \gtrsim \rho^4d, n_2 \gtrsim \rho^4k$ and $n_1 \gtrsim n_2$. Under the choice of $\lambda_1$ and $\lambda_2$ in~\eqref{eqn: lambda_1}, with probability $1-o(1)$, the excess risk of the learned predictor $\bx \mapsto \lag \bx ,\widehat{\btheta}^{E+1}\rag $ in~\eqref{eqn: linear fine-tune} is 
\[
\begin{aligned}
    \E_{\btheta^{*E+1}}[{\rm ER}(\widehat{\btheta}^{E+1})]
&\lesssim \frac{\sigma^2 \norm{\Lambda_{11}}\Tr\lp  R^{*\top}_1\Sigma_X R^*_1 \rp }{n_2} + \frac{\sigma^2 \norm{\Lambda_{22}}\Tr\lp  R^{*\top}_2\Sigma_X R^*_2 \rp }{n_2}\\ &+ \frac{\sigma \norm{\Lambda_{22}}\norm{\btheta^*}}{n_2(\min(\Lambda_{22}) - \norm{\Lambda_{11}} ) )}\sqrt{\frac{d\Tr(R^{*\top}_2\Sigma_X R^*_2)}{n_2 E}}  + \xi(\sigma,n_1,n_2,d,k,E)
\end{aligned}
\]
where
\[
\begin{aligned}
   \xi(\sigma,n_1,n_2,d,k,E) &=  \frac{\sigma^2 \norm{\Sigma_X}\Tr(\Lambda_{11})}{n_2E} + \frac{\sigma^4\norm{\Sigma_X}\Tr(R^{*\top}_1\Sigma^{-1}_XR^{*}_1)}{n_1n_2E} + \frac{\sigma}{n_2}\sqrt{\frac{\Tr(R^{*\top}_1\Sigma_XR^{*}_1)\Tr(\Lambda_{11})}{n_2 E}}\\ &+ \frac{\sigma^2}{n_2}\sqrt{\frac{\Tr(R^{*\top}_1\Sigma_XR^{*}_1)\Tr(R^{*\top}_1\Sigma^{-1}_XR^{*}_1)}{n_1 n_2E}}
\end{aligned}
\]
is the lower order terms.
\end{theorem}  
\begin{remark}
In our setting, traditional ridge regression will yield a $\sqrt{\frac{\sigma^2\|\btheta^{*E+1}\|^2\Tr(\Sigma_{X})}{n_2}}$ (and $\norm{\btheta^{*E+1}}^2$ is bounded by $\|\btheta^*\|^2+\|\Lambda_{11}\|+\|\Lambda_{22}\|$) rate. In comparison, our rate not only achieves a fast rate, but also manages to 1) quickly eliminate the error caused by the variation in utilizing content and environmental features (reflected by the $\Lambda_{11}$ and $\Lambda_{22}$ in the first two terms), and 2) fully utilize the shared part learned jointly from all $E$ environments (reflected by the third term that involves $\|\btheta^*\|$ and is lower-order).  
\end{remark}

\subsection{Proof of Theorem~\ref{thm: linear main full}}
We first prove two claims on the covariance concentration for both source and target tasks.
\begin{claim}[covariance concentration of source tasks]
\label{claim: linear uniform}
    Suppose $n_1 \gtrsim \rho^4 d$. Then it holds with probability $1-o(1)$ that for $e\in [E]$,
    \[
    0.9\Sigma_{X^e} \preceq \frac{1}{n_1} X^{e\top}X^e \preceq 1.1\Sigma_{X^e}.
    \]
\end{claim}
\noindent \textit{Proof.  }For $e \in [E]$, we write $X^{e} = \bar{X}^{e}\Sigma^{1/2}_{X^{e}}$. Lemma~\ref{lemma: covariance concentration of subgaussian rv} gives
\[
0.9 I_d \preceq \frac{1}{n_1} \bar{X}^{e\top} \bar{X}^e \preceq 1.1 I_d,
\]
which implies that 
\[
    0.9\Sigma_{X^e} \preceq \frac{1}{n_2} \Sigma^{1/2}_{X^{e}} \bar{X}^{e\top}\bar{X}^{e}\Sigma^{1/2}_{X^{e}} \preceq 1.1\Sigma_{X^e}.
    \]
Taking a union bound over all $e\in[E]$ finishes the proof.
\hfill $\square$
\begin{claim}[covariance concentration of target tasks]
\label{claim: linear point}
    Suppose $n_2 \gtrsim \rho^4 k$. Then for any $B \in \reals^{d \times k}$, $k \le d$, independent of $X^{E+1}$, it holds with probability $1-o(1)$ that 
    \[
    0.9B^\top\Sigma_{X^{E+1}}B \preceq \frac{1}{n_1} B^\top X^{E+1 \top}X^{E+1} B \preceq 1.1B^\top\Sigma_{X^{E+1}}B.
    \]
\end{claim}
\noindent \textit{Proof.  } Let $X^{E+1} = \bar{X}^{E+1}\Sigma^{1/2}_{X^{E+1}}$. Let the SVD of $\Sigma^{1/2}_{X^{E+1}}B$ be $UDV^\top$ where $U \in \reals^{d\times k}$, $D, V \in \reals^{k \times k}$. Then it can be verified that the rows of $X^{E+1}U$ are $k$-dimensional i.i.d. $\rho^2$-subgaussian random vectors with zero mean and identity covariance. Then similarly to the proof of the source covariance concentration, Lemma~\ref{lemma: covariance concentration of subgaussian rv} together with some algebra operations finish the proof.
\hfill $\square$

Then we will analyze the expected excess risk with respect to the meta distribution
\begin{equation}
\label{eqn: linear excess risk}
    \begin{aligned}
        \E_{\btheta^{*E+1}}[{\rm ER}(\widehat{\btheta}^{E+1})] & = \frac{1}{2}\E_{\btheta^{*E+1}} \E_{(\bx,y) \sim \mu_{E+1}} [ \lag \bx, R^*\btheta^{*E+1} -  \widehat{\btheta}^{E+1}\rag^2 ].
    \end{aligned}
\end{equation}
The learned parameter of the target task from~\eqref{eqn: linear fine-tune} has the closed form
\[
\begin{aligned}
   \widehat{\btheta}^{E+1} &= \lp\frac{1}{n_2}X^{E+1 \top}X^{E+1} + \lambda_1 \mathcal{P}_{\widehat{R}_1} + \lambda_2 \mathcal{P}^\perp_{\widehat{R}_1} \rp^{-1} \lp \frac{1}{n_2}X^{E+1 \top}\by^{E+1}  + \lambda_1 \mathcal{P}_{\widehat{R}_1} \bar{\btheta}^E\rp\\ &= C^{-1}_{n_2, \lambda_1, \lambda_2}\lp \frac{1}{n_2}X^{E+1 \top}\by^{E+1}  + \lambda_1 \mathcal{P}_{\widehat{R}_1} \bar{\btheta}^E\rp.
\end{aligned}
\] where 
\[
C_{n_2, \lambda_1,\lambda_2} = \lp\frac{1}{n_2}X^{E+1 \top}X^{E+1} + \lambda_1 \mathcal{P}_{\widehat{R}_1} + \lambda_2 \mathcal{P}^\perp_{\widehat{R}_1} \rp.
\]
Recall that for $e \in [E+1]$, the rotated meta distribution is 
\[
R^*\btheta^{e*} \overset{\rm i.i.d.}{\sim}\mathcal{N} \left( R^*_1\btheta^*, \Sigma_{\text{RM}} \right), \quad \Sigma_{\text{RM}} = R^*_1\Lambda_{11}R^{*\top}_1 +  R^*_2\Lambda_{22}R^{*\top}_2.
\]

Then it can be verified that 
\begin{equation}
\label{eqn: difference fact}
    \begin{aligned}
    \widehat{\btheta}^{E+1} - R^*\btheta^{*E+1} &= C^{-1}_{n_2, \lambda_1, \lambda_2}\lp \frac{1}{n_2} X^{E+1 \top}\by^{E+1}  + \lambda_1 \mathcal{P}_{\widehat{R}_1} \bar{\btheta}^E\rp - R^*\btheta^{*E+1}\\
    &= C^{-1}_{n_2, \lambda_1, \lambda_2}\lp  \frac{1}{n_2} X^{E+1 \top}\lp X^{E+1}R^*\btheta^{*E+1} + \bz_{n_2} \rp + \lambda_1 \mathcal{P}_{\widehat{R}_1} \bar{\btheta}^E  \rp - R^*\btheta^{*E+1}\\
    &= \lambda_1 C^{-1}_{n_2, \lambda_1,\lambda_2} \mathcal{P}_{\widehat{R}_1} \lp \bar{\btheta}^E - R^*\btheta^{*E+1} \rp - \lambda_2 C^{-1}_{n_2, \lambda_1,\lambda_2} \mathcal{P}^\perp_{\widehat{R}_1}  R^*\btheta^{*E+1}\\  &+  \frac{1}{n_2}C^{-1}_{n_2, \lambda_1,\lambda_2}X^{E+1 \top}\bz_{n_2}
\end{aligned}
\end{equation}
where in the second equality we use $\by^{E+1} = X^{E+1}R^*\btheta^{*E+1} + \bz_{n_2}$. Plugging~\eqref{eqn: difference fact} into~\eqref{eqn: linear excess risk}, the excess risk can be decomposed into $5$ terms and we will bound each of them:
\begin{equation}
    \begin{aligned}
         \E_{\btheta^{*E+1}}[{\rm ER}(\widehat{\btheta}^{E+1})]  
&= \frac{1}{2}\E_{\btheta^{*E+1}} \ls \lp \widehat{\btheta}^{E+1} - R^*\btheta^{*E+1}  \rp^\top \Sigma_{X^{E+1}}\lp \widehat{\btheta}^{E+1} - R^*\btheta^{*E+1}  \rp \rs\\
&= \frac{1}{2}(T_1+T_2+T_3 + T_4 +T_5), 
    \end{aligned}
\end{equation}
where 
\[
\begin{aligned}
    T_1 &= \E_{\btheta^{*E+1}}\ls\lambda_1^2\lp \bar{\btheta}^E - \btheta^{*E+1}  \rp^\top \mathcal{P}_{\widehat{R}_1} C^{-1}_{n_2, \lambda_1,\lambda_2} \Sigma_{X^{E+1}} C^{-1}_{n_2, \lambda_1,\lambda_2} \mathcal{P}_{\widehat{R}_1}  \lp \bar{\btheta}^E - \btheta^{*E+1}  \rp  \rs,\\
    T_2 &= \E_{\btheta^{*E+1}}\ls\lambda_2^2\btheta^{*E+1 \top} \mathcal{P}^\perp_{\widehat{R}_1} C^{-1}_{n_2, \lambda_1,\lambda_2} \Sigma_{X^{E+1}} C^{-1}_{n_2, \lambda_1,\lambda_2} \mathcal{P}^\perp_{\widehat{R}_1}   \btheta^{*E+1}\rs,\\
    T_3 &= \frac{1}{n_2^2} \bz_{n_2}^\top X^{E+1 }C^{-1}_{n_2, \lambda_1,\lambda_2} \Sigma_{X^{E+1}} C^{-1}_{n_2, \lambda_1,\lambda_2} X^{E+1 \top}\bz_{n_2},\\
     T_4 &= \E_{\btheta^{*E+1}}\ls\lag \lambda_1 C^{-1}_{n_2, \lambda_1,\lambda_2} \mathcal{P}_{\widehat{R}_1} \lp \bar{\btheta}^E - R^*\btheta^{*E+1} \rp,   \frac{1}{n_2}C^{-1}_{n_2, \lambda_1,\lambda_2}X^{E+1 \top}\bz_{n_2}\rag \rs, \\
      T_5 &= \E_{\btheta^{*E+1}}\ls \lag -\lambda_2 C^{-1}_{n_2, \lambda_1,\lambda_2} \mathcal{P}^\perp_{\widehat{R}_1}  R^*\btheta^{*E+1},\frac{1}{n_2}C^{-1}_{n_2 \lambda_1,\lambda_2}X^{E+1 \top}\bz_{n_2} \rag \rs.
\end{aligned}
\] 

\paragraph{Estimation of $T_1$.} Let $\bv = R^*\btheta^{*E+1} - R_1^* \btheta^*$. We have $\E_{\btheta^{*E+1}} \ls \bv \rs =0$ and $\E_{\btheta^{*E+1}} \ls \bv\bv^\top \rs =\Sigma_{{\rm RM}}$. Then the first term $T_1$ can be written as 
\begin{align*}
    T_1 &= \E_{\btheta^{*E+1}}\ls\lambda_1^2\lp \bar{\btheta}^E -  R_1^* \btheta^*+ \bv  \rp^\top \mathcal{P}_{\widehat{R}_1} C^{-1}_{n_2, \lambda_1,\lambda_2} \Sigma_{X^{E+1}} C^{-1}_{n_2, \lambda_1,\lambda_2} \mathcal{P}_{\widehat{R}_1}  \lp \bar{\btheta}^E - R_1^* \btheta^* + \bv  \rp  \rs\\
&= \lambda_1^2 \lp  \bar{\btheta}^E - R_1^* \btheta^*    \rp^\top \mathcal{P}_{\widehat{R}_1} C^{-1}_{n_2, \lambda_1,\lambda_2} \Sigma_{X^{E+1}} C^{-1}_{n_2, \lambda_1,\lambda_2} \mathcal{P}_{\widehat{R}_1}\lp  \bar{\btheta}^E- R_1^* \btheta^*   \rp\\ &+  \lambda_1^2 \E_{\btheta^{*E+1}}  \ls \bv^\top  \mathcal{P}_{\widehat{R}_1} C^{-1}_{n_2, \lambda_1,\lambda_2} \Sigma_{X^{E+1}}  C^{-1}_{n_2, \lambda_1,\lambda_2} \mathcal{P}_{\widehat{R}_1} \bv \rs\\
&= \lambda_1 ^2 \lp  \bar{\btheta}^E-R_1^* \btheta^*    \rp^\top \mathcal{P}_{\widehat{R}_1} C^{-1}_{n_2, \lambda_1,\lambda_2} \Sigma_{X^{E+1}} C^{-1}_{n_2, \lambda_1,\lambda_2} \mathcal{P}_{\widehat{R}_1}\lp  \bar{\btheta}^E- R_1^* \btheta^*    \rp\\ 
&+ \Tr \lp \E_{\btheta^{*E+1}}  \ls \widehat{R}_1^\top \bv \bv^\top \widehat{R}_1 \rs \lambda_1^2 \widehat{R}_1^\top C^{-1}_{n_2, \lambda_1,\lambda_2} \Sigma_{X^{E+1}} C^{-1}_{n_2, \lambda_1,\lambda_2}\widehat{R}_1 \rp\\
&\le  \underbrace{\norm{\lambda_1 \Sigma_X^{1/2} C^{-1}_{n_2,\lambda_1, \lambda_2} \widehat{R}_1 \widehat{R}_1^\top \lp  \bar{\btheta}^E- R_1^* \btheta^*   \rp }^2}_{T_{1,1}} + \underbrace{\lambda_1^2 \Tr  \lp 
     \widehat{R}_1^\top C^{-1}_{n_2, \lambda_1,\lambda_2} \Sigma_X C^{-1}_{n_2, \lambda_1,\lambda_2}\widehat{R}_1  \rp  \norm{\E_{\btheta^{*E+1}} \ls  \widehat{R}_1^\top \bv \bv^\top \widehat{R}_1 \rs}}_{T_{1,2}} .
\end{align*}
To bound $T_{1,1}$ we need the performance guarantee of $\bar{\btheta}^E$ on the direction of the content features, which is given by the following lemma.
\begin{lemma}[source task guarantee on the content feature space]
\label{lemma: guarantee of source linear}
Under the conditions of Theorem~\ref{thm: linear main full}, with probability $1-o(1)$, 
\[
 \norm{\mathcal{P}_{\widehat{R}_1} \lp  \bar{\btheta}^E  - R_1\btheta^*  \rp}  \lesssim \sqrt{\frac{1}{E}\Tr(\Lambda_{11})} + \sqrt{\frac{\sigma^2\Tr(R^{*\top}_1\Sigma^{-1}_XR^{*}_1)}{n_1 E}}+ \zeta(n_1,E,d)
\] and $\zeta(n_1,E,d)$ is the lower terms
\[
\begin{aligned}
    \zeta(n_1,E,d)= (\min(\Lambda_{22}) - \norm{\Lambda_{11}} ) )^{-1} \lp \sqrt{\frac{d}{E}}\norm{\Lambda_{22}} + \frac{\sigma^2}{n_1\lambda_{1}(\Sigma_X)} \rp \lp\sqrt{\frac{\Tr(\Lambda_{11}+\Lambda_{22})}{E}} + \sqrt{\frac{\Tr(\Sigma^{-1})}{n_1 E}} \rp
\end{aligned}
\]
where $\lambda_{i}(\Sigma_X)$ denotes the $i$-th smallest eigenvalue of $\Sigma_X$.
\end{lemma}
Under the choice
\begin{align}
    \label{eqn: lambda_1}
    \lambda_1 = \lambda_2 = \frac{\lambda_{1}(\Sigma_X)\sigma}{\sqrt{n_2}-\sigma},
\end{align}
we have
\[
\begin{aligned}
    T_{1,1} &\le \lambda_1^2 \norm{\Sigma_X^{1/2} C^{-1}_{n_2,\lambda_1, \lambda_2}  \widehat{R}_1 }^2 \norm{\mathcal{P}_{\widehat{R}_1} \lp  \bar{\btheta}^E  - R_1\btheta^*  \rp}^2\\
    &\lesssim \lambda_1^2 \norm{\Sigma_X^{1/2} C^{-1}_{n_2,\lambda_1, \lambda_2}  \widehat{R}_1 }^2 \lp \frac{\Tr(\Lambda_{11})}{E} + \frac{\sigma^2}{n_1 E}\Tr(R^{*\top}_1\Sigma^{-1}_XR^{*}_1) \rp\\
    &\lesssim \frac{\sigma^2 \norm{\Sigma_X}}{n_2}\lp \frac{\Tr(\Lambda_{11})}{E} + \frac{\sigma^2}{n_1 E}\Tr(R^{*\top}_1\Sigma^{-1}_XR^{*}_1) \rp\\
    &= \frac{\sigma^2 \norm{\Sigma_X}\Tr(\Lambda_{11})}{n_2E} + \frac{\sigma^4\norm{\Sigma_X}\Tr(R^{*\top}_1\Sigma^{-1}_XR^{*}_1)}{n_1n_2E}.
\end{aligned}
\]
Note that $T_{1,1}$ is of order $O(\sigma^2 k / n_2 E + \sigma^4 k / n_1 n_2 E)$ which will be omitted as a lower order term. 

\[
\begin{aligned}
  T_{1,2} = & \lambda_1^2 \Tr  \lp 
     \widehat{R}_1^\top C^{-1}_{n_2, \lambda_1,\lambda_2} \Sigma_X C^{-1}_{n_2, \lambda_1,\lambda_2}\widehat{R}_1  \rp  \norm{\E_{\btheta^{*E+1}} \ls  \widehat{R}_1^\top \bv \bv^\top \widehat{R}_1 \rs}\\ 
    &\le \frac{\sigma^2 \Tr(\widehat{R}_1^\top \Sigma_X \widehat{R_1})}{n_2} \norm{\widehat{R}_1^\top R_1^*\Lambda_{11} R_1^{*\top}\widehat{R}_1 + \widehat{R}_1^\top R_2^*\Lambda_{22}  R_2^{*\top}\widehat{R}_1} \\
    &\lesssim  \frac{\sigma^2 \Tr(R^{*\top}_1 \Sigma_X R^*_1)}{n_2} \ls \lp 1 - \norm{R_2^{*\top}\widehat{R}_1}^2 \rp\norm{\Lambda_{11}}^2 + \norm{R_2^{*\top}\widehat{R}_1}^2 \norm{\Lambda_{22}} \rs\\
    &\lesssim \frac{\sigma^2 \norm{\Lambda_{11}}\Tr(R^{*\top}_1 \Sigma_X R^*_1)}{n_2}
\end{aligned}
\]
where in the second equality we use $\Sigma_{{\rm RM}} = R^*_1 \Lambda_{11} R^{*\top}_1 + R^*_2 \Lambda_{22} R^{*\top}_2$, and in the third inequality we use Lemma~\ref{lemma: Davis-Kahan} that $\min_{O \in \mathcal{O}(k)}\norm{\widehat{R}_1 - R^*_1 O}^2 \asymp
 \norm{R_2^{*\top}\widehat{R}_1}^2 = O(d/E + \sigma^2/n_1^2)$. $T_{1,2}$ is of order $O(\sigma^2k/n_2)$ which is one of the dominant terms in the main theorem.

\paragraph{Estimation of $T_2.$}
Similarly to the first term, the second term can be written as
\[
\begin{aligned}
T_2 &= \E_{\btheta^{*E+1}}\ls\lambda_2^2\btheta^{*E+1 \top} \mathcal{P}^\perp_{\widehat{R}_1} C^{-1}_{n_2, \lambda_1,\lambda_2} \Sigma_{X^{E+1}} C^{-1}_{n_2, \lambda_1,\lambda_2} \mathcal{P}^\perp_{\widehat{R}_1}   \btheta^{*E+1}\rs\\
&\lesssim \Tr \lp  \E_{\btheta^{*E+1}}\ls \widehat{R}_2^\top \btheta^{*E+1}\btheta^{*E+1 \top}  \widehat{R}_2\rs \lambda_2^2  \widehat{R}_2^\top C^{-1}_{n_2, \lambda_1,\lambda_2} \Sigma_X C^{-1}_{n_2, \lambda_1,\lambda_2}  \widehat{R}_2 \rp\\
&\lesssim \lambda_2^2 \lp \lp 1 - \norm{R_2^{*\top}\widehat{R}_1}^2 \rp\norm{ \Lambda_{22}}  + \norm{R_2^{*\top}\widehat{R}_1}^2 \norm{\Lambda_{11}}\rp \Tr \lp \widehat{R}_2^\top C^{-1}_{n_2, \lambda_1,\lambda_2} \Sigma C^{-1}_{n_2, \lambda_1,\lambda_2}  \widehat{R}_2 \rp
\end{aligned}
\] 

Similarly to the second term of $T_1$, 
\begin{align*}
   T_2 \lesssim \frac{\sigma^2 \norm{\Lambda_{22}}\Tr\lp  R^{*\top}_2\Sigma_X R^*_2 \rp }{n_2}.
\end{align*} Under Assumption \ref{assumption: trace assumption}, one can verify that $T_2 \lesssim T_{1,2}$.

\paragraph{Estimation of $T_3.$}
\begin{align*}
    T_3 &= \frac{1}{n_2^2} \bz_{n_2}^\top X^{E+1 }C^{-1}_{n_2, \lambda_1,\lambda_2} \Sigma_{X^{E+1}} C^{-1}_{n_2, \lambda_1,\lambda_2} X^{E+1 \top}\bz_{n_2}\\
    &= \frac{1}{n^2_2}\norm{\Sigma^{1/2}_{X^{E+1}}C^{-1}_{n_2,\lambda_1,\lambda_2}X^{E+1 \top}\bz_{n_2}}^2\\
    &\lesssim \frac{1}{n^2_2}\norm{\mathcal{P}_{R^*_1}\Sigma^{1/2}_{X}C^{-1}_{n_2,\lambda_1,\lambda_2}X^{E+1 \top}\bz_{n_2}}^2 + \frac{1}{n^2_2}\norm{\mathcal{P}_{R^*_2}\Sigma^{1/2}_{X}C^{-1}_{n_2,\lambda_1,\lambda_2}X^{E+1 \top}\bz_{n_2}}^2\\
    &\lesssim \frac{\sigma^2 \Tr(R^{*\top}_1 \Sigma_X R_1^*)}{n_2} + \frac{\sigma^2 \Tr(R^{*\top}_2 \Sigma_X R_2^*)}{n_2}\\
    &\lesssim \lp 1+ \frac{\norm{\Lambda_{11}}}{\norm{\Lambda_{22}}}\rp \frac{\sigma^2 \Tr(R^{*\top}_1 \Sigma_X R_1^*)}{n_2}
\end{align*}
where we use Assumption~\ref{assumption: trace assumption} in the last step.

\paragraph{Estimation of $T_4.$}

\[
\begin{aligned}
T_4 &= \E_{\btheta^{*E+1}}\ls\lag \lambda_1 C^{-1}_{n_2, \lambda_1,\lambda_2} \mathcal{P}_{\widehat{R}_1} \lp \bar{\btheta}^E - R^*\btheta^{*E+1} \rp,   \frac{1}{n_2}C^{-1}_{n_2, \lambda_1,\lambda_2}X^{E+1 \top}\bz_{n_2}\rag \rs \\
&= \lag \lambda_1 C^{-1}_{n_2, \lambda_1,\lambda_2} \mathcal{P}_{\widehat{R}_1} \lp \bar{\btheta}^E - R_1^*\btheta^{*} \rp,   \frac{1}{n_2}C^{-1}_{n_2, \lambda_1,\lambda_2}X^{E+1 \top}\bz_{n_2}\rag\\
&= \frac{\lambda_1}{n_2} \lp \bar{\btheta}^E - R_1^*\btheta^{*} \rp^\top \mathcal{P}_{\widehat{R}_1} C^{-2}_{n_2, \lambda_1,\lambda_2}X^{E+1 \top}\bz_{n_2}\\
&\le \norm{\mathcal{P}_{\widehat{R}_1}\lp \bar{\btheta}^E - R_1^*\btheta^{*} \rp} \norm{\frac{\lambda_1}{n_2}\mathcal{P}_{\widehat{R}_1}C^{-2}_{n_2, \lambda_1,\lambda_2}X^{E+1 \top}\bz_{n_2}}\\
&\lesssim \lp \sqrt{\frac{1}{E}\Tr(\Lambda_{11})} + \sqrt{\frac{\sigma^2\Tr(R^{*\top}_1\Sigma^{-1}_XR^{*}_1)}{n_1 E}}\rp \lp \frac{\lambda_1 \sigma}{\sqrt{n_2}}\sqrt{\Tr(\widehat{R}^{\top}_1 C^{-2}_{n_2, \lambda_1,\lambda_2} \Sigma_X C^{-2}_{n_2, \lambda_1,\lambda_2}\widehat{R}_1)} \rp\\
&\lesssim \frac{\sigma}{n_2}\sqrt{\frac{\Tr(R^{*\top}_1\Sigma_XR^{*}_1)}{n_2}}\lp \sqrt{\frac{1}{E}\Tr(\Lambda_{11})} + \sqrt{\frac{\sigma^2\Tr(R^{*\top}_1\Sigma^{-1}_XR^{*}_1)}{n_1 E}}\rp \\
&\lesssim \frac{\sigma}{n_2}\sqrt{\frac{\Tr(R^{*\top}_1\Sigma_XR^{*}_1)\Tr(\Lambda_{11})}{n_2 E}} + \frac{\sigma^2}{n_2}\sqrt{\frac{\Tr(R^{*\top}_1\Sigma_XR^{*}_1)\Tr(R^{*\top}_1\Sigma^{-1}_XR^{*}_1)}{n_1 n_2E}}
\end{aligned}
\]
where in the second inequality we plug in our choice of $\lambda_1$ in ~\eqref{eqn: lambda_1}.

\paragraph{Estimation of $T_5.$} The estimation of $T_5$ is almost the same as that of $T_4$. 
\[
\begin{aligned}
T_5 &= \E_{\btheta^{*E+1}}\ls \lag -\lambda_2 C^{-1}_{n_2, \lambda_1,\lambda_2} \mathcal{P}^\perp_{\widehat{R}_1}  R^*\btheta^{*E+1},\frac{1}{n_2}C^{-1}_{n_2, \lambda_1,\lambda_2}X^{E+1 \top}\bz_{n_2} \rag \rs\\
&= \lag -\lambda_2 C^{-1}_{n_2, \lambda_1,\lambda_2} \mathcal{P}^\perp_{\widehat{R}_1}  R_1^*\btheta^{*},\frac{1}{n_2}C^{-1}_{n_2, \lambda_1,\lambda_2}X^{E+1 \top}\bz_{n_2} \rag\\
&\lesssim \norm{\mathcal{P}^\perp_{\widehat{R}_1}  R_1^*\btheta^{*}} \norm{\frac{\lambda_2}{n_2}\mathcal{P}_{\widehat{R}_1}C^{-2}_{n_2, \lambda_1,\lambda_2}X^{E+1 \top}\bz_{n_2}}\\
&\lesssim \frac{\norm{\btheta^*}}{(\min(\Lambda_{22}) - \norm{\Lambda_{11}} ) )} \lp \sqrt{\frac{d}{E}}\norm{\Lambda_{22}} + \frac{\sigma^2}{n_1\lambda_{1}(\Sigma_X)} \rp  \frac{\sigma}{n_2}\sqrt{\frac{\Tr(R^{*\top}_2\Sigma_X R^*_2)}{n_2}}\\
&\lesssim \frac{\norm{\btheta^*}}{(\min(\Lambda_{22}) - \norm{\Lambda_{11}} ) )} \lp \frac{\sigma \norm{\Lambda_{22}}}{n_2}\sqrt{\frac{d\Tr(R^{*\top}_2\Sigma_X R^*_2)}{n_2 E}} +\frac{\sigma^3 \lambda^{-1}_1(\Sigma_X)}{n_1 n_2} \sqrt{\frac{\Tr(R^{*\top}_2\Sigma_X R^*_2)}{n_2}}\rp\\
&\lesssim  \frac{\sigma \norm{\Lambda_{22}}\norm{\btheta^*}}{n_2(\min(\Lambda_{22}) - \norm{\Lambda_{11}} ) )}\sqrt{\frac{d\Tr(R^{*\top}_2\Sigma_X R^*_2)}{n_2 E}}. 
\end{aligned}
\]

With $T_{1,2},T_2,T_3,T_5$ being dominant terms and $T_{1,1},T_4$ being the lower order terms, we obtain the final bound 
\[
\begin{aligned}
    \E_{\btheta^{*E+1}}[{\rm ER}(\widehat{\btheta}^{E+1})]
&= \frac{1}{2}(T_1+T_2+T_3 + T_4 +T_5)\\
&\lesssim \frac{\sigma^2 \norm{\Lambda_{11}}\Tr\lp  R^{*\top}_1\Sigma_X R^*_1 \rp }{n_2} + \frac{\sigma^2 \norm{\Lambda_{22}}\Tr\lp  R^{*\top}_2\Sigma_X R^*_2 \rp }{n_2}\\ &+ \frac{\sigma \norm{\Lambda_{22}}\norm{\btheta^*}}{n_2(\min(\Lambda_{22}) - \norm{\Lambda_{11}} ) )}\sqrt{\frac{d\Tr(R^{*\top}_2\Sigma_X R^*_2)}{n_2 E}}  + \xi(\sigma,n_1,n_2,d,k,E)
\end{aligned}
\]
where
\[
\begin{aligned}
   \xi(\sigma,n_1,n_2,d,k,E) &=  \frac{\sigma^2 \norm{\Sigma_X}\Tr(\Lambda_{11})}{n_2E} + \frac{\sigma^4\norm{\Sigma_X}\Tr(R^{*\top}_1\Sigma^{-1}_XR^{*}_1)}{n_1n_2E} + \frac{\sigma}{n_2}\sqrt{\frac{\Tr(R^{*\top}_1\Sigma_XR^{*}_1)\Tr(\Lambda_{11})}{n_2 E}}\\ &+ \frac{\sigma^2}{n_2}\sqrt{\frac{\Tr(R^{*\top}_1\Sigma_XR^{*}_1)\Tr(R^{*\top}_1\Sigma^{-1}_XR^{*}_1)}{n_1 n_2E}}.
\end{aligned}
\]

\subsection{Proof of Lemma~\ref{lemma: guarantee of source linear}}
For $1\le e \le E$, $\widehat{\btheta}^e$ is the OLS estimator:
\[
\begin{aligned}
    \widehat{\btheta}^e &= \lp X^{e\top}X^{e} \rp^{-1} X^{e\top}\by^e\\
    &= \lp X^{e\top}X^{e} \rp^{-1} X^{e\top} \lp X^e \btheta^{*e} + \bz^e \rp\\
    &=  \btheta^{*e} + \lp X^{e\top}X^{e} \rp^{-1} X^{e\top}\bz^e.
\end{aligned}
\] 

Let $\widehat{O} \in \mathcal{O}(k)$ be such that 
\[
\widehat{O} = \underset{O \in \mathcal{O}(k)} {\argmin}\norm{\widehat{R}_1 - R_1 O}.
\]  Then the $\ell_2$ error can be decomposed as: 
\[
\begin{aligned}
    \norm{\widehat{R}_1^\top\lp\frac{1}{E}\sum_{e=1}^E \widehat{\btheta}^e - R_1^*\btheta^* \rp} &\le \norm{ R_1^{*\top} \lp \frac{1}{E}\sum_{e=1}^E R^*\btheta^{*e} -   R_1\btheta^* \rp}\\ &+ \norm{ R_1^{*\top} \lp\frac{1}{E}\sum_{e=1}^E \lp X^{e\top}X^{e} \rp^{-1} X^{e\top}\bz^e \rp}\\ &+ \norm{\lp\widehat{R}_1^{\top} - \widehat{O}^\top R^{*\top}_1\rp \lp \frac{1}{E}\sum_{e=1}^E R^*\btheta^{*e} -   R_1\btheta^* \rp}\\ &+ \norm{\lp \widehat{R}^\top_1 - \widehat{O}^\top R_1^{*\top} \rp \lp\frac{1}{E}\sum_{e=1}^E \lp X^{e\top}X^{e} \rp^{-1} X^{e\top}\bz^e \rp}.
\end{aligned}
\]
where we use the fact that $\ell_2$ norm is orthogonal invariant. Note that 
\[
    R_1^{*\top} \lp \frac{1}{E}\sum_{e=1}^E R^*\btheta^{*e} -   R_1\btheta^* \rp \sim \mathcal{N}\lp \bzero , \frac{1}{E} \Lambda_{11}  \rp.
\] The Chernoff bound gives that with probability $1-o(1)$,
\[
\norm{ R_1^{*\top} \lp \frac{1}{E}\sum_{e=1}^E R^*\btheta^{*e} -   R_1\btheta^* \rp} \lesssim \sqrt{\frac{1}{E}\Tr(\Lambda_{11})}.
\]
Note that 
\[
R_1^{*\top} \lp\frac{1}{E}\sum_{e=1}^E \lp X^{e\top}X^{e} \rp^{-1} X^{e\top}\bz^e \rp \sim \mathcal{N}\lp \bzero , \frac{\sigma^2}{n_1E^2} \sum_{e=1}^E  R_1^{*\top}\lp \frac{X^{e\top}X^e}{n_1} \rp^{-1}R_1^*  \rp .
\] The Chernoff bound gives that with probability $1-o(1)$,
\[
    \norm{R_1^{*\top} \lp\frac{1}{E}\sum_{e=1}^E \lp X^{e\top}X^{e} \rp^{-1} X^{e\top}\bz^e \rp } \lesssim \frac{\sigma}{\sqrt{n_1 E}} \sqrt{\Tr(R^{*\top}_1\Sigma^{-1}_XR^{*}_1)}
\] 

The following lemma provides the estimation of $ \norm{\widehat{R}_1 - R_1^* \widehat{O}}$.
\begin{lemma}
\label{lemma: Davis-Kahan}
Under the conditions in Theorem~\ref{thm: linear main full}, with probability $1-o(1)$, 
\[
 \norm{\widehat{R}_1 - R_1^* \widehat{O}}  \lesssim (\min(\Lambda_{22}) - \norm{\Lambda_{11}} ) )^{-1} \lp \sqrt{\frac{d}{E}}\norm{\Lambda_{22}} + \frac{\sigma^2}{n_1\lambda_{1}(\Sigma_X)} \rp.
\]
\end{lemma}

Then the remaining two terms are of lower order.
\begin{align*}
    &\norm{\lp\widehat{R}_1^\top - \widehat{O}^\top R^\top_1\rp \lp \frac{1}{E}\sum_{e=1}^E\btheta^{*e} -   R_1\btheta^* \rp} \lesssim \norm{\widehat{R}_1 - R_1 \widehat{O}}\norm{\frac{1}{E}\sum_{e=1}^E\btheta^{*e} -   R_1\btheta^* }\\
    &\lesssim (\min(\Lambda_{22}) - \norm{\Lambda_{11}} ) )^{-1} \lp \sqrt{\frac{d}{E}}\norm{\Lambda_{22}} + \frac{\sigma^2}{n_1\lambda_{1}(\Sigma_X)} \rp \sqrt{\frac{\Tr(\Lambda_{11}+\Lambda_{22})}{E}}.
\end{align*}

Putting the terms together, we get
\begin{align*}
     &\norm{\lp \widehat{R}^\top_1 - \widehat{O}^\top R_1^{*\top} \rp \lp\frac{1}{E}\sum_{e=1}^E \lp X^{e\top}X^{e} \rp^{-1} X^{e\top}\bz^e \rp} \lesssim \norm{\widehat{R}_1 - R_1 \widehat{O}} \norm{\frac{1}{E}\sum_{e=1}^E \lp X^{e\top}X^{e} \rp^{-1} X^{e\top}\bz^e }\\
      &\lesssim (\min(\Lambda_{22}) - \norm{\Lambda_{11}} ) )^{-1} \lp \sqrt{\frac{d}{E}}\norm{\Lambda_{22}} + \frac{\sigma^2}{n_1\lambda_{1}(\Sigma_X)} \rp \sqrt{\frac{\Tr(\Sigma^{-1})}{n_1 E}}.
\end{align*}

\subsection{Proof of Lemma~\ref{lemma: Davis-Kahan}}

Let $\widehat{O} \in \mathcal{O}(k)$ be such that 
\[
\widehat{O} = \argmin_{O \in \mathcal{O}(k)} \norm{\widehat{R}_1 - R_1 O}.
\] It can be verified that $\widehat{O} = \bar{U}\bar{V}^\top$ given the SVD of $R_1^{*\top} \widehat{R}_1$ being $\bar{U} \bar{D}\bar{V}^\top$. Then Davis-Kahan Theorem (Theorem~\ref{thm:dk}) gives the bound
\begin{align*}
    \norm{\widehat{R}_1 - R_1 \widehat{O}} &= \norm{\lp I_d - R_1^*R_1^{*\top} \rp \widehat{R}_1} + \norm{R_1^{*\top} \widehat{R}_1 - \widehat{O}}\\
    &\le 2\norm{\sin \lp R_1 ,\widehat{R}_1\rp}\\
    &\le \frac{2\norm{\widehat{\Sigma}_{\widehat{\theta}}-\Sigma_{{\rm RM}}}}{\min(\Lambda_{22}) - \norm{\Lambda_{11}} )  - \norm{\widehat{\Sigma}_{\widehat{\theta}}-\Sigma_{{\rm RM}}}}.
\end{align*}
Under Assumption \ref{assumption: eigengap}, 
\begin{align*}
    \norm{\widehat{R}_1 - R_1 \widehat{O}} &\lesssim \frac{\norm{\widehat{\Sigma}_{\widehat{\theta}}-\Sigma_{{\rm RM}}}}{\min(\Lambda_{22}) - \norm{\Lambda_{11}} ) }.
\end{align*}
For readability of the proof, we define the auxiliary quantities as follows. Let 
\begin{align*}
\Delta^e &= \lp X^{e\top}X^{e} \rp^{-1} X^{e\top}\bz^e\\
\bar{\btheta}^{*E}&= \frac{1}{E} \sum_{e=1}^{E} \btheta^{*e},\\
\bar{\Delta}^E &= \frac{1}{E} \sum_{e=1}^{E} \Delta^e,\\
    \widehat{\Sigma}_{\btheta^*} &=  \frac{1}{E} \sum_{e=1}^{E} \lp \btheta^{*e} -\bar{\btheta}^{*E}\rp \lp \btheta^{*e} -\bar{\btheta}^{*E} \rp^\top, \\
    \widehat{\Sigma}^0_{\btheta^*} &=  \frac{1}{E} \sum_{e=1}^{E} \lp \btheta^{*e} -R^*_1 \btheta^*\rp \lp \btheta^{*e} -R^*_1 \btheta^* \rp^\top,\\
    \widehat{\Sigma}_{\Delta} &= \frac{1}{E} \sum_{e=1}^{E} \lp \Delta^e - \bar{\Delta}^E \rp  \lp \Delta^e - \bar{\Delta}^E \rp^\top,\\
    \widehat{\Sigma}^0_{\Delta} &= \frac{1}{E} \sum_{e=1}^{E} \Delta^e   \Delta^{e \top}.
\end{align*}
Then $\norm{\widehat{\Sigma}_{\widehat{\theta}}-\Sigma_{{\rm RM}}}$ can be decomposed as:
\[
\begin{aligned}
    \norm{\widehat{\Sigma}_{\widehat{\theta}}-\Sigma_{{\rm RM}}} &= \norm{\frac{1}{E} \sum_{e=1}^E \lp \widehat{\btheta}^{e}  -\bar{\btheta}^E\rp \lp\widehat{\btheta}^{e}  -\bar{\btheta}^E \rp^\top -\Sigma_{{\rm RM}}}\\
    &= \norm{\frac{1}{E} \sum_{e=1}^E \lp \btheta^{*e}  -\bar{\btheta}^{*E} + \Delta^e - \bar{\Delta}^{E}\rp \lp \btheta^{*e}  -\bar{\btheta}^{*E}  + \Delta^e - \bar{\Delta}^{E} \rp^\top -\Sigma_{{\rm RM}}}\\ &= \norm{ \widehat{\Sigma}_{\btheta^*}  -\Sigma_{{\rm RM}}} + \norm{\widehat{\Sigma}_{\Delta}}\\ &+  \norm{\frac{1}{E} \sum_{e=1}^E \btheta^{*e} \Delta^{e \top}-  \bar{\btheta}^{*E}\bar{\Delta}^{E\top} + \frac{1}{E} \sum_{e=1}^E \Delta^{e} \btheta^{*e \top} -  \bar{\Delta}^{E}\bar{\btheta}^{*E \top}}.
\end{aligned}
\] 
\paragraph{Estimation of $\norm{ \widehat{\Sigma}_{\btheta^*}  -\Sigma_{{\rm RM}}}.$}
We have that
\begin{align*}
    \norm{ \widehat{\Sigma}_{\btheta^*}  -\Sigma_{{\rm RM}}} &\le  \norm{ \widehat{\Sigma}^0_{\btheta^*}  -\Sigma_{{\rm RM}}} + \norm{\lp \bar{\btheta}^{*E} - R_1^*\btheta^* \rp \lp \bar{\btheta}^{*E} - R_1^*\btheta^* \rp^\top}.
\end{align*}
Since $\btheta^{e} - R^*_1 \btheta^*$ is centered Gaussian with covariance $\Sigma_{{\rm RM}}$, by the standard covariance estimation in \citep{Wain19}, with probability $1-o(1)$, 
\[
\norm{ \widehat{\Sigma}^0_{\btheta^*}  -\Sigma_{{\rm RM}}} \lesssim \sqrt{\frac{d}{E}}\norm{\Lambda_{22}}.
\] The second term upper bounded by the squared norm of the Gaussian vector $\norm{\bar{\btheta}^{*E} - R_1^*\btheta^* }$
\[
    \norm{\lp \bar{\btheta}^{*E} - R_1^*\btheta^* \rp \lp \bar{\btheta}^{*E} - R_1^*\btheta^* \rp^\top} = \norm{\bar{\btheta}^{*E} - R_1^*\btheta^* }^2 \lesssim \frac{\Tr\lp \Lambda_{11} + \Lambda_{22} \rp}{E}.
\]
\paragraph{Estimation of $\norm{\widehat{\Sigma}_{\Delta}}.$}
Similarly, $\Delta^e$ is a centered Gaussian with covariance $\sigma^2\Sigma_{X^e}^{-1}/n_1$. Then 
\begin{align*}
    \norm{\widehat{\Sigma}_{\Delta}} &\le \norm{\Sigma_{\Delta}} + \norm{\Sigma_{\Delta} -\widehat{\Sigma^0_{\Delta}}} + \norm{\bar{\Delta}^E \bar{\Delta}^{E\top}}\\
    &\lesssim \frac{\sigma^2}{n_1\lambda_{1}(\Sigma)} \lp 1 + \sqrt{\frac{d}{E}} \rp + \frac{\sigma^2\Tr\lp \Sigma_X^{-1} \rp}{n_1 E}
\end{align*}

\paragraph{Estimation of $\norm{\frac{1}{E} \sum_{e=1}^E \btheta^{*e} \Delta^{e \top}-  \bar{\btheta}^{*E}\bar{\Delta}^{E\top} + \frac{1}{E} \sum_{e=1}^E \Delta^{e} \btheta^{*e \top} -  \bar{\Delta}^{E}\bar{\btheta}^{*E \top}}.$}
Since the two parts of the sum are transpose of each other, it suffices to find the upper bound one of them only. 
Then we have that 
\begin{align*}
    \norm{\frac{1}{E} \sum_{e=1}^E \btheta^{*e} \Delta^{e \top}-  \bar{\btheta}^{*E}\bar{\Delta}^{E\top} } &\le  \norm{\frac{1}{E} \sum_{e=1}^E (\btheta^{*e} - R_1^*\btheta^*) \Delta^{e \top} } + \norm{(R_1^*\btheta^* - \bar{\btheta}^{*E})\bar{\Delta}^{E\top}}
\end{align*}
Chernoff bound gives the upper bound of the second term. With probability $1-o(1)$,
\[
\begin{aligned}
    \norm{(R_1^*\btheta^* - \bar{\btheta}^{*E})\bar{\Delta}^{E\top}} &\le \norm{R_1^*\btheta^* - \bar{\btheta}^{*E}}\norm{\bar{\Delta}^{E}}\\
    & \lesssim \sqrt{\frac{\Tr(\Lambda_{11}+\Lambda_{22})}{E} \frac{\sigma^2\Tr\lp \Sigma_X^{-1} \rp}{n_1 E}
    }
\end{aligned}
\]
Then we will apply Matrix Bernstein (Theorem~\ref{thm:bernstein}) to find the upper bound of the first term. Note that each term is bounded by
\begin{align*}
    \norm{\frac{1}{E}(\btheta^{*e} - R_1^*\btheta^*) \Delta^{e \top} } \lesssim \frac{\sigma \sqrt{\Tr(\Lambda_{11}+\Lambda_{22}) \Tr\lp \Sigma_X^{-1} \rp}}{\sqrt{n_1}E}.
\end{align*}
Ther variance proxy is 
\begin{align*}
    \norm{\mathbb{E} \ls \sum_{e=1}^E \frac{1}{E^2} (\btheta^{*e} - R_1^*\btheta^*) \Delta^{e \top}\Delta^{e}(\btheta^{*e} - R_1^*\btheta^*)^\top \rs} &\lesssim \frac{ \E_{\btheta^*} \norm{\btheta^{*e} - R_1^*\btheta^*}^2 \norm{\Delta^{e}}^2}{E} \\
    &\lesssim \frac{\sigma^2 \Tr(\Lambda_{11}+\Lambda_{22}) \Tr\lp \Sigma_X^{-1} \rp}{n_1 E}.
\end{align*}
Then, by Bernstein's inequality, 
\begin{align*}
    \norm{\frac{1}{E} \sum_{e=1}^E (\btheta^{*e} - R_1^*\btheta^*) \Delta^{e \top} } &\lesssim \sqrt{\frac{\sigma^2 \Tr(\Lambda_{11}+\Lambda_{22}) \Tr\lp \Sigma_X^{-1} \rp\log(E)}{n_1 E}} +  \frac{\sigma \sqrt{\Tr(\Lambda_{11}+\Lambda_{22}) \Tr\lp \Sigma_X^{-1} \rp}\log(E)}{\sqrt{n_1}E}\\
    &\lesssim \sqrt{\frac{\sigma^2 \Tr(\Lambda_{11}+\Lambda_{22}) \Tr\lp \Sigma_X^{-1} \rp\log(E)}{n_1 E}}.
\end{align*}

Thus, we combine the terms together and omit the  lower order terms:
\begin{align*}
    \norm{\widehat{\Sigma}_{\widehat{\theta}}-\Sigma_{{\rm RM}}} \lesssim \sqrt{\frac{d}{E}}\norm{\Lambda_{22}} + \frac{\sigma^2}{n_1\lambda_{1}(\Sigma_X)}
\end{align*}
and 
\[
\norm{\widehat{R}_1 - R_1^* \widehat{O}} \lesssim \frac{\norm{\widehat{\Sigma}_{\widehat{\theta}}-\Sigma_{{\rm RM}}}}{\min(\Lambda_{22}) - \norm{\Lambda_{11}} ) } \lesssim (\min(\Lambda_{22}) - \norm{\Lambda_{11}} ) )^{-1} \lp \sqrt{\frac{d}{E}}\norm{\Lambda_{22}} + \frac{\sigma^2}{n_1\lambda_{1}(\Sigma_X)} \rp.
\]

\section{Technical ingredients}

\begin{theorem}[Generalized Davis-Kahan theorem{~\citep{DLS21,zhong2023improved}} ]\label{thm:dk}
Consider the eigenvalue problem $N^{-1}M \bu = \lambda\bu$ where $M$ and $N$ are both Hermitian, and $N$ is positive definite. Let $X$ be the matrix that has the eigenvectors of $N^{-1}M$ as columns. Then $N^{-1}M$ is diagonalizable and can be written as
\[
	N^{-1}M = X \Lambda X^H = X_1 \Lambda_1 X_1^H + X_2 \Lambda_2 X_2^H
\]where
\[
	X^{-1} = \begin{bmatrix}
		X_1 &X_2
	\end{bmatrix}^{-1} = \begin{bmatrix}
		Y_1^H \\ Y_2^H
	\end{bmatrix}, ~~~ \Lambda = \begin{bmatrix}
		\Lambda_1 &  \\ &\Lambda_2
	\end{bmatrix}.
\]

Suppose $\delta = \min_{i} |(\Lambda_{2})_{ii} - \widehat{\lambda}|$ is the absolute separation of $\widehat{\lambda}$ from $(\Lambda_{2})_{ii}$, then for any vector $\widehat{\bu}$ we have
\[
	\norm{P\widehat{\bu} } \le \frac{\sqrt{\kappa(N)}\norm{(N^{-1}M-\widehat{\lambda} I_n)\widehat{\bu}}}{\delta}.
\]where $P = (Y^{\dagger}_2)^H(Y_2)^H = I - (X^{\dagger}_1)^H(X_1)^H$ is the orthogonal projection matrix onto the orthogonal complement of the column space of $X_1$, $\kappa(N) = \norm{N}\norm{N^{-1}}$ is the condition number of $N$ and $Y^{\dagger}_2$ is the Moore-Penrose inverse of $Y_2$.

When $N = I$ and $(\hat{\lambda}, \widehat{\bu})$ be an eigen-pair of a matrix $\widehat{M}$, we have 
\[
	\sin\theta \leq \frac{\norm{(M-\widehat{M})\widehat{\bu}}}{\delta}
\] where $\theta$ is the canonical angle between $\widehat{\bu}$ and the column space of $X_1$. In this case the theorem reduces to the classical Davis-Kahan theorem \citep{DK70}.
\end{theorem}

\begin{lemma}[\citep{DHK21}, Lemma A.6]\label{lemma: covariance concentration of subgaussian rv}
    Let $\ba_1,\ldots, \ba_n$ be i.i.d. $d-$dimensional random vectors such that $\E[\ba_i] = 0$, $\E[\ba_i\ba_i^\top] =I_d$, and $\ba_i$ is $\rho^2$-subgaussian. If $n \gtrsim \rho^4 d$, then it holds with probability $1-o(1)$ that 
    \[
    0.9 I_d \preceq \frac{1}{n} \sum_{i=1}^n \ba_i \ba_i^\top \preceq 1.1 I_d.
    \]
\end{lemma}

\begin{theorem}[Matrix Bernstein~\citep{T12}]
	Consider a finite sequence of independent random matrices $\lc Z_k \rc$. Assume that each random matrix satisfies
	\begin{align*}
		\mathbb{E}Z_k =0,~~~\norm{Z_k} \le R.
	\end{align*}
	Then for all $t \ge 0$,
	\begin{align*}
		\mathbb{P}\lp \norm{\sum_{k}Z_k} \ge t\rp \le (d_1 + d_2) \cdot \exp(-\frac{t^2/2}{\sigma^2+ Rt/3}).
	\end{align*}
	where 
	\begin{align*}
		\sigma^2 = \max \lc \norm{\sum_k \mathbb{E}Z_k^\top Z_k},\norm{\sum_k \mathbb{E}Z_k Z_k^\top} \rc.
	\end{align*}
	Then with probability at least $1-n^{-\gamma+1}$,
	\begin{align*}
		\norm{\sum_{k}Z_k} \le \sqrt{2\gamma \sigma^2\log(d_1+d_2)} + \frac{2\gamma R \log(d_1+d_2)}{3}.
	\end{align*}
\label{thm:bernstein}
\end{theorem}

\newpage
\section{Experimental Results}

We show the full results in the following tables. B stands for base feature, Y stands for the target-specific feature, S stands for the shared feature, and E stands for the environment-specific feature. We use the Adam optimizer with lr $\in$ [1e-5, 5e-5, 1e-4], batch size $\in$ [32, 64, 128]. We choose the best model based on the source validation accuracy.

\subsection{Linear Probing Results}

We fixed the source validation dataset based on random seed and use that validation dataset for hyperparameter tuning. We vary the hyperparameters of the logistic regression models: C $\in$ [1e-5, 1e-4, 1e-3, 1e-2, 1e-1], lbfgs solver, max iter = 1000. We report the results in the following \autoref{tab:lp_officehome}, \autoref{tab:lp_pacs}, \autoref{tab:lp_vlcs}, \autoref{tab:lp_terraincognita}.

\begin{table}[!htbp]
    \small
\scalebox{0.9}{
    \begin{tabular}{lllllllll}
\toprule
 &  & 0.10 & 0.20 & 0.40 & 0.60 & 0.80 & 1.00 & mean \\
$E_{t}$ & Method &  &  &  &  &  &  &  \\
\midrule
\multirow[t]{11}{*}{0} & DANN & 0.4527 ± 0.0107 & 0.5984 ± 0.0109 & 0.6527 ± 0.0090 & 0.7111 ± 0.0054 & 0.7193 ± 0.0092 & 0.7177 ± 0.0087 & 0.6420 ± 0.0090 \\
 & DIWA & 0.5835 ± 0.0132 & 0.6675 ± 0.0097 & 0.7276 ± 0.0094 & 0.7358 ± 0.0082 & 0.7720 ± 0.0118 & 0.7909 ± 0.0048 & 0.7129 ± 0.0095 \\
 & ERM & 0.5745 ± 0.0105 & 0.6848 ± 0.0046 & 0.7243 ± 0.0107 & 0.7539 ± 0.0108 & 0.7827 ± 0.0082 & 0.7926 ± 0.0085 & 0.7188 ± 0.0089 \\
 & NUC-0.01 & 0.4840 ± 0.0129 & 0.6543 ± 0.0043 & 0.7407 ± 0.0097 & 0.7934 ± 0.0136 & 0.8049 ± 0.0040 & 0.8165 ± 0.0062 & 0.7156 ± 0.0084 \\
 & NUC-0.1 & 0.4733 ± 0.0072 & 0.6782 ± 0.0111 & \textbf{0.7687 ± 0.0117} & \textbf{0.8049 ± 0.0038} & 0.8132 ± 0.0062 & 0.8099 ± 0.0077 & 0.7247 ± 0.0080 \\
 & PN-B & 0.5852 ± 0.0051 & 0.6733 ± 0.0115 & 0.7449 ± 0.0068 & 0.7613 ± 0.0065 & \textbf{0.8165 ± 0.0025} & 0.8074 ± 0.0053 & 0.7314 ± 0.0063 \\
 & PN-B-Y & 0.5770 ± 0.0076 & 0.6955 ± 0.0050 & 0.7490 ± 0.0068 & 0.7761 ± 0.0053 & 0.8074 ± 0.0094 & 0.8329 ± 0.0074 & 0.7396 ± 0.0069 \\
 & PN-B-Y-S-E & 0.5844 ± 0.0057 & 0.7119 ± 0.0114 & 0.7366 ± 0.0077 & 0.7580 ± 0.0054 & 0.8041 ± 0.0110 & \textbf{0.8337 ± 0.0055} & 0.7381 ± 0.0078 \\
 & PN-Y & 0.5877 ± 0.0080 & \textbf{0.7169 ± 0.0081} & 0.7481 ± 0.0102 & 0.7588 ± 0.0083 & 0.8066 ± 0.0075 & 0.8280 ± 0.0126 & 0.7410 ± 0.0091 \\
 & PN-Y-S & 0.5745 ± 0.0070 & 0.7062 ± 0.0095 & 0.7432 ± 0.0135 & 0.7860 ± 0.0135 & 0.8041 ± 0.0127 & 0.8222 ± 0.0080 & 0.7394 ± 0.0107 \\
 & PN-Y-S-E & \textbf{0.5901 ± 0.0098} & 0.7029 ± 0.0105 & 0.7498 ± 0.0068 & 0.7778 ± 0.0066 & 0.8082 ± 0.0028 & 0.8296 ± 0.0094 & \textbf{0.7431 ± 0.0077} \\
\cline{1-9}
\multirow[t]{11}{*}{1} & DANN & 0.4842 ± 0.0128 & 0.5986 ± 0.0077 & 0.6856 ± 0.0078 & 0.7011 ± 0.0090 & 0.7263 ± 0.0047 & 0.7318 ± 0.0072 & 0.6546 ± 0.0082 \\
 & DIWA & 0.5716 ± 0.0082 & 0.6577 ± 0.0101 & 0.7121 ± 0.0077 & 0.7487 ± 0.0067 & 0.7808 ± 0.0050 & 0.7725 ± 0.0082 & 0.7072 ± 0.0077 \\
 & ERM & 0.5538 ± 0.0071 & 0.6490 ± 0.0065 & 0.6989 ± 0.0054 & 0.7204 ± 0.0034 & 0.7432 ± 0.0038 & 0.7803 ± 0.0089 & 0.6909 ± 0.0058 \\
 & NUC-0.01 & 0.4838 ± 0.0070 & 0.6293 ± 0.0049 & 0.6879 ± 0.0055 & 0.7414 ± 0.0075 & 0.7730 ± 0.0063 & 0.7867 ± 0.0053 & 0.6837 ± 0.0061 \\
 & NUC-0.1 & 0.4824 ± 0.0101 & 0.6449 ± 0.0066 & 0.6728 ± 0.0095 & 0.7382 ± 0.0087 & 0.7744 ± 0.0061 & 0.7794 ± 0.0067 & 0.6820 ± 0.0080 \\
 & PN-B & 0.5542 ± 0.0082 & 0.6568 ± 0.0028 & 0.7281 ± 0.0062 & 0.7684 ± 0.0059 & 0.7812 ± 0.0063 & 0.7973 ± 0.0026 & 0.7143 ± 0.0053 \\
 & PN-B-Y & 0.5593 ± 0.0092 & 0.6545 ± 0.0035 & \textbf{0.7465 ± 0.0045} & 0.7808 ± 0.0092 & 0.7844 ± 0.0053 & 0.8009 ± 0.0064 & \textbf{0.7211 ± 0.0064} \\
 & PN-B-Y-S-E & 0.5712 ± 0.0066 & 0.6526 ± 0.0076 & 0.7217 ± 0.0060 & 0.7698 ± 0.0060 & \textbf{0.7863 ± 0.0009} & \textbf{0.8064 ± 0.0044} & 0.7180 ± 0.0052 \\
 & PN-Y & 0.5584 ± 0.0052 & 0.6563 ± 0.0068 & 0.7245 ± 0.0070 & \textbf{0.7817 ± 0.0058} & 0.7826 ± 0.0128 & 0.8018 ± 0.0056 & 0.7175 ± 0.0072 \\
 & PN-Y-S & \textbf{0.5744 ± 0.0113} & 0.6517 ± 0.0025 & 0.7263 ± 0.0046 & 0.7744 ± 0.0070 & 0.7826 ± 0.0071 & 0.7899 ± 0.0092 & 0.7166 ± 0.0070 \\
 & PN-Y-S-E & 0.5707 ± 0.0078 & \textbf{0.6613 ± 0.0035} & 0.7359 ± 0.0047 & 0.7661 ± 0.0034 & 0.7808 ± 0.0085 & 0.7995 ± 0.0030 & 0.7191 ± 0.0052 \\
\cline{1-9}
\multirow[t]{11}{*}{2} & DANN & 0.6649 ± 0.0056 & 0.7423 ± 0.0018 & 0.8230 ± 0.0062 & 0.8509 ± 0.0058 & 0.8667 ± 0.0036 & 0.8784 ± 0.0049 & 0.8044 ± 0.0046 \\
 & DIWA & 0.7734 ± 0.0082 & 0.8189 ± 0.0059 & 0.8649 ± 0.0049 & 0.8959 ± 0.0036 & 0.9063 ± 0.0052 & 0.9045 ± 0.0055 & 0.8607 ± 0.0055 \\
 & ERM & 0.7743 ± 0.0069 & 0.8401 ± 0.0019 & 0.8716 ± 0.0048 & 0.9113 ± 0.0038 & 0.8995 ± 0.0039 & 0.9122 ± 0.0033 & 0.8682 ± 0.0041 \\
 & NUC-0.01 & 0.7176 ± 0.0030 & 0.8320 ± 0.0049 & \textbf{0.9077 ± 0.0031} & 0.9126 ± 0.0046 & \textbf{0.9261 ± 0.0024} & \textbf{0.9302 ± 0.0028} & \textbf{0.8710 ± 0.0035} \\
 & NUC-0.1 & 0.7257 ± 0.0085 & 0.8279 ± 0.0052 & 0.8977 ± 0.0052 & \textbf{0.9198 ± 0.0032} & 0.9212 ± 0.0053 & 0.9248 ± 0.0023 & 0.8695 ± 0.0050 \\
 & PN-B & 0.7689 ± 0.0024 & 0.8365 ± 0.0037 & 0.8689 ± 0.0053 & 0.8919 ± 0.0026 & 0.9140 ± 0.0034 & 0.9203 ± 0.0038 & 0.8667 ± 0.0035 \\
 & PN-B-Y & 0.7721 ± 0.0083 & 0.8351 ± 0.0044 & 0.8613 ± 0.0039 & 0.8955 ± 0.0044 & 0.9144 ± 0.0034 & 0.9117 ± 0.0039 & 0.8650 ± 0.0047 \\
 & PN-B-Y-S-E & \textbf{0.7797 ± 0.0038} & 0.8459 ± 0.0051 & 0.8730 ± 0.0039 & 0.8914 ± 0.0049 & 0.9162 ± 0.0008 & 0.9185 ± 0.0041 & 0.8708 ± 0.0038 \\
 & PN-Y & 0.7527 ± 0.0071 & 0.8446 ± 0.0054 & 0.8770 ± 0.0042 & 0.8923 ± 0.0046 & 0.8991 ± 0.0060 & 0.9234 ± 0.0052 & 0.8649 ± 0.0054 \\
 & PN-Y-S & 0.7703 ± 0.0074 & \textbf{0.8468 ± 0.0065} & 0.8671 ± 0.0033 & 0.8955 ± 0.0034 & 0.9027 ± 0.0039 & 0.9176 ± 0.0021 & 0.8667 ± 0.0044 \\
 & PN-Y-S-E & 0.7770 ± 0.0069 & 0.8455 ± 0.0090 & 0.8829 ± 0.0074 & 0.8959 ± 0.0053 & 0.9023 ± 0.0054 & 0.9225 ± 0.0027 & 0.8710 ± 0.0061 \\
\cline{1-9}
\multirow[t]{11}{*}{3} & DANN & 0.6385 ± 0.0043 & 0.7183 ± 0.0051 & 0.7518 ± 0.0067 & 0.7638 ± 0.0044 & 0.7853 ± 0.0067 & 0.7862 ± 0.0058 & 0.7407 ± 0.0055 \\
 & DIWA & 0.7867 ± 0.0085 & 0.8000 ± 0.0034 & 0.8271 ± 0.0040 & 0.8408 ± 0.0044 & 0.8344 ± 0.0037 & 0.8560 ± 0.0073 & 0.8242 ± 0.0052 \\
 & ERM & \textbf{0.8248 ± 0.0050} & 0.8367 ± 0.0053 & 0.8619 ± 0.0044 & 0.8555 ± 0.0079 & 0.8546 ± 0.0084 & 0.8555 ± 0.0054 & 0.8482 ± 0.0061 \\
 & NUC-0.01 & 0.7156 ± 0.0052 & 0.8147 ± 0.0085 & 0.8596 ± 0.0033 & 0.8670 ± 0.0036 & 0.8656 ± 0.0049 & 0.8812 ± 0.0112 & 0.8339 ± 0.0061 \\
 & NUC-0.1 & 0.7165 ± 0.0111 & 0.8165 ± 0.0058 & 0.8587 ± 0.0017 & 0.8679 ± 0.0049 & 0.8624 ± 0.0030 & 0.8771 ± 0.0078 & 0.8332 ± 0.0057 \\
 & PN-B & 0.7894 ± 0.0077 & 0.8312 ± 0.0036 & 0.8628 ± 0.0029 & 0.8651 ± 0.0031 & 0.8693 ± 0.0062 & 0.8633 ± 0.0066 & 0.8469 ± 0.0050 \\
 & PN-B-Y & 0.7940 ± 0.0058 & 0.8252 ± 0.0023 & 0.8615 ± 0.0006 & 0.8656 ± 0.0034 & \textbf{0.8817 ± 0.0071} & 0.8647 ± 0.0039 & 0.8488 ± 0.0038 \\
 & PN-B-Y-S-E & 0.7872 ± 0.0066 & 0.8353 ± 0.0099 & 0.8587 ± 0.0034 & 0.8633 ± 0.0039 & 0.8683 ± 0.0064 & 0.8665 ± 0.0060 & 0.8466 ± 0.0060 \\
 & PN-Y & 0.7986 ± 0.0037 & 0.8422 ± 0.0074 & 0.8495 ± 0.0051 & 0.8615 ± 0.0046 & 0.8688 ± 0.0022 & 0.8720 ± 0.0028 & 0.8488 ± 0.0043 \\
 & PN-Y-S & 0.7812 ± 0.0038 & 0.8381 ± 0.0030 & \textbf{0.8679 ± 0.0060} & 0.8628 ± 0.0039 & 0.8693 ± 0.0061 & \textbf{0.8835 ± 0.0028} & 0.8505 ± 0.0043 \\
 & PN-Y-S-E & 0.7904 ± 0.0051 & \textbf{0.8454 ± 0.0051} & 0.8661 ± 0.0039 & \textbf{0.8706 ± 0.0048} & 0.8679 ± 0.0039 & 0.8697 ± 0.0059 & \textbf{0.8517 ± 0.0048} \\
\cline{1-9}
\bottomrule
\end{tabular}

}
\caption{Linear Probing results on OfficeHome dataset.}\label{tab:lp_officehome}
\end{table}

\begin{table}[!htbp]
    \small
\scalebox{0.9}{
    \begin{tabular}{lllllllll}
\toprule
 &  & 0.10 & 0.20 & 0.40 & 0.60 & 0.80 & 1.00 & mean \\
$E_{t}$ & Method &  &  &  &  &  &  &  \\
\midrule
\multirow[t]{11}{*}{0} & DANN & 0.2956 ± 0.0118 & 0.3337 ± 0.0150 & 0.3171 ± 0.0079 & 0.3698 ± 0.0179 & 0.4059 ± 0.0192 & 0.4215 ± 0.0124 & 0.3572 ± 0.0140 \\
 & DIWA & \textbf{0.9405 ± 0.0081} & \textbf{0.9346 ± 0.0033} & \textbf{0.9512 ± 0.0077} & \textbf{0.9551 ± 0.0036} & \textbf{0.9473 ± 0.0076} & \textbf{0.9444 ± 0.0065} & \textbf{0.9455 ± 0.0061} \\
 & ERM & 0.8702 ± 0.0043 & 0.8673 ± 0.0042 & 0.8849 ± 0.0068 & 0.9044 ± 0.0059 & 0.9141 ± 0.0072 & 0.9190 ± 0.0037 & 0.8933 ± 0.0053 \\
 & NUC-0.01 & 0.8429 ± 0.0052 & 0.8780 ± 0.0056 & 0.8937 ± 0.0036 & 0.9015 ± 0.0066 & 0.8966 ± 0.0070 & 0.8966 ± 0.0054 & 0.8849 ± 0.0056 \\
 & NUC-0.1 & 0.8273 ± 0.0070 & 0.8702 ± 0.0099 & 0.8839 ± 0.0081 & 0.9024 ± 0.0044 & 0.8976 ± 0.0083 & 0.8937 ± 0.0099 & 0.8792 ± 0.0079 \\
 & PN-B & 0.8810 ± 0.0077 & 0.8878 ± 0.0074 & 0.9239 ± 0.0065 & 0.9190 ± 0.0059 & 0.9307 ± 0.0071 & 0.9288 ± 0.0065 & 0.9119 ± 0.0068 \\
 & PN-B-Y & 0.8849 ± 0.0072 & 0.8868 ± 0.0076 & 0.9190 ± 0.0037 & 0.9180 ± 0.0039 & 0.9337 ± 0.0045 & 0.9346 ± 0.0075 & 0.9128 ± 0.0057 \\
 & PN-B-Y-S-E & 0.9024 ± 0.0044 & 0.9054 ± 0.0043 & 0.9220 ± 0.0046 & 0.9337 ± 0.0077 & 0.9434 ± 0.0061 & 0.9405 ± 0.0047 & 0.9246 ± 0.0053 \\
 & PN-Y & 0.9054 ± 0.0048 & 0.9005 ± 0.0092 & 0.9268 ± 0.0034 & 0.9317 ± 0.0053 & 0.9337 ± 0.0048 & 0.9268 ± 0.0031 & 0.9208 ± 0.0051 \\
 & PN-Y-S & 0.9122 ± 0.0084 & 0.8917 ± 0.0068 & 0.9220 ± 0.0074 & 0.9327 ± 0.0039 & 0.9337 ± 0.0037 & 0.9220 ± 0.0051 & 0.9190 ± 0.0059 \\
 & PN-Y-S-E & 0.9073 ± 0.0034 & 0.9024 ± 0.0067 & 0.9171 ± 0.0094 & 0.9239 ± 0.0067 & 0.9415 ± 0.0041 & 0.9210 ± 0.0086 & 0.9189 ± 0.0065 \\
\cline{1-9}
\multirow[t]{11}{*}{1} & DANN & 0.3419 ± 0.0137 & 0.3983 ± 0.0048 & 0.4256 ± 0.0048 & 0.4359 ± 0.0102 & 0.4538 ± 0.0148 & 0.4906 ± 0.0068 & 0.4244 ± 0.0092 \\
 & DIWA & 0.8906 ± 0.0098 & 0.9154 ± 0.0067 & 0.9231 ± 0.0072 & 0.9402 ± 0.0045 & 0.9402 ± 0.0038 & 0.9342 ± 0.0089 & 0.9239 ± 0.0068 \\
 & ERM & 0.8752 ± 0.0114 & 0.9111 ± 0.0065 & 0.9231 ± 0.0062 & 0.9316 ± 0.0076 & 0.9470 ± 0.0037 & \textbf{0.9581 ± 0.0086} & 0.9244 ± 0.0074 \\
 & NUC-0.01 & 0.8436 ± 0.0083 & 0.8581 ± 0.0075 & 0.8897 ± 0.0044 & 0.8974 ± 0.0045 & 0.8906 ± 0.0042 & 0.9179 ± 0.0048 & 0.8829 ± 0.0056 \\
 & NUC-0.1 & 0.8308 ± 0.0046 & 0.8624 ± 0.0044 & 0.8923 ± 0.0076 & 0.8949 ± 0.0086 & 0.9034 ± 0.0068 & 0.9145 ± 0.0030 & 0.8830 ± 0.0058 \\
 & PN-B & 0.9034 ± 0.0064 & 0.9205 ± 0.0064 & 0.9308 ± 0.0067 & 0.9256 ± 0.0026 & 0.9265 ± 0.0028 & 0.9436 ± 0.0046 & 0.9251 ± 0.0049 \\
 & PN-B-Y & \textbf{0.9222 ± 0.0056} & 0.9162 ± 0.0064 & 0.9179 ± 0.0095 & 0.9171 ± 0.0032 & 0.9350 ± 0.0031 & 0.9359 ± 0.0076 & 0.9241 ± 0.0059 \\
 & PN-B-Y-S-E & 0.9128 ± 0.0032 & 0.9248 ± 0.0072 & \textbf{0.9342 ± 0.0052} & 0.9359 ± 0.0041 & \textbf{0.9487 ± 0.0027} & 0.9487 ± 0.0043 & 0.9342 ± 0.0044 \\
 & PN-Y & 0.9068 ± 0.0104 & 0.9282 ± 0.0048 & 0.9231 ± 0.0052 & 0.9359 ± 0.0019 & 0.9308 ± 0.0078 & 0.9487 ± 0.0014 & 0.9289 ± 0.0052 \\
 & PN-Y-S & 0.9162 ± 0.0053 & \textbf{0.9342 ± 0.0017} & 0.9265 ± 0.0046 & \textbf{0.9427 ± 0.0048} & 0.9368 ± 0.0046 & 0.9556 ± 0.0052 & \textbf{0.9353 ± 0.0044} \\
 & PN-Y-S-E & 0.9077 ± 0.0074 & 0.9342 ± 0.0032 & 0.9291 ± 0.0040 & 0.9342 ± 0.0066 & 0.9479 ± 0.0039 & 0.9513 ± 0.0055 & 0.9340 ± 0.0051 \\
\cline{1-9}
\multirow[t]{11}{*}{2} & DANN & 0.6431 ± 0.0062 & 0.6311 ± 0.0140 & 0.6862 ± 0.0122 & 0.6862 ± 0.0082 & 0.7186 ± 0.0143 & 0.7138 ± 0.0088 & 0.6798 ± 0.0106 \\
 & DIWA & 0.9737 ± 0.0031 & 0.9737 ± 0.0015 & 0.9820 ± 0.0033 & 0.9880 ± 0.0000 & 0.9796 ± 0.0031 & 0.9844 ± 0.0031 & 0.9802 ± 0.0023 \\
 & ERM & 0.9533 ± 0.0035 & 0.9569 ± 0.0035 & 0.9653 ± 0.0040 & 0.9677 ± 0.0056 & 0.9677 ± 0.0052 & 0.9593 ± 0.0035 & 0.9617 ± 0.0042 \\
 & NUC-0.01 & 0.9569 ± 0.0064 & 0.9737 ± 0.0045 & 0.9749 ± 0.0055 & \textbf{0.9916 ± 0.0031} & 0.9808 ± 0.0022 & 0.9880 ± 0.0019 & 0.9776 ± 0.0039 \\
 & NUC-0.1 & 0.9545 ± 0.0024 & 0.9760 ± 0.0019 & 0.9784 ± 0.0049 & 0.9868 ± 0.0022 & 0.9808 ± 0.0035 & \textbf{0.9904 ± 0.0024} & 0.9778 ± 0.0029 \\
 & PN-B & 0.9701 ± 0.0060 & 0.9796 ± 0.0036 & \textbf{0.9796 ± 0.0041} & 0.9808 ± 0.0048 & 0.9737 ± 0.0031 & 0.9784 ± 0.0045 & 0.9770 ± 0.0043 \\
 & PN-B-Y & 0.9784 ± 0.0059 & 0.9820 ± 0.0054 & \textbf{0.9856 ± 0.0031} & 0.9832 ± 0.0029 & 0.9772 ± 0.0048 & 0.9820 ± 0.0050 & 0.9814 ± 0.0045 \\
 & PN-B-Y-S-E & \textbf{0.9796 ± 0.0041} & 0.9772 ± 0.0048 & 0.9808 ± 0.0029 & 0.9796 ± 0.0031 & 0.9784 ± 0.0052 & 0.9880 ± 0.0019 & 0.9806 ± 0.0037 \\
 & PN-Y & 0.9749 ± 0.0029 & \textbf{0.9856 ± 0.0031} & 0.9832 ± 0.0040 & 0.9760 ± 0.0033 & \textbf{0.9892 ± 0.0012} & 0.9808 ± 0.0022 & \textbf{0.9816 ± 0.0028} \\
 & PN-Y-S & 0.9725 ± 0.0024 & 0.9820 ± 0.0050 & 0.9784 ± 0.0062 & 0.9737 ± 0.0045 & 0.9784 ± 0.0024 & 0.9880 ± 0.0019 & 0.9788 ± 0.0037 \\
 & PN-Y-S-E & 0.9641 ± 0.0066 & 0.9760 ± 0.0063 & 0.9820 ± 0.0038 & \textbf{0.9904 ± 0.0024} & 0.9784 ± 0.0015 & 0.9832 ± 0.0040 & 0.9790 ± 0.0041 \\
\cline{1-9}
\multirow[t]{11}{*}{3} & DANN & 0.6132 ± 0.0038 & 0.6626 ± 0.0069 & 0.7008 ± 0.0071 & 0.6962 ± 0.0120 & 0.7298 ± 0.0049 & 0.7394 ± 0.0091 & 0.6903 ± 0.0073 \\
 & DIWA & 0.8718 ± 0.0055 & \textbf{0.8896 ± 0.0042} & \textbf{0.9170 ± 0.0032} & 0.9074 ± 0.0034 & 0.9115 ± 0.0053 & 0.9160 ± 0.0018 & 0.9022 ± 0.0039 \\
 & ERM & \textbf{0.8936 ± 0.0039} & 0.8891 ± 0.0030 & 0.9033 ± 0.0052 & \textbf{0.9206 ± 0.0057} & \textbf{0.9277 ± 0.0046} & \textbf{0.9232 ± 0.0041} & \textbf{0.9096 ± 0.0044} \\
 & NUC-0.01 & 0.6972 ± 0.0055 & 0.7659 ± 0.0091 & 0.8081 ± 0.0034 & 0.8036 ± 0.0061 & 0.8193 ± 0.0023 & 0.8249 ± 0.0060 & 0.7865 ± 0.0054 \\
 & NUC-0.1 & 0.6987 ± 0.0042 & 0.7603 ± 0.0056 & 0.7969 ± 0.0103 & 0.8107 ± 0.0059 & 0.8249 ± 0.0064 & 0.8275 ± 0.0051 & 0.7865 ± 0.0063 \\
 & PN-B & 0.8427 ± 0.0066 & 0.8718 ± 0.0097 & 0.8840 ± 0.0040 & 0.8967 ± 0.0039 & 0.9033 ± 0.0039 & 0.9013 ± 0.0061 & 0.8833 ± 0.0057 \\
 & PN-B-Y & 0.8601 ± 0.0040 & 0.8636 ± 0.0026 & 0.8947 ± 0.0049 & 0.9018 ± 0.0013 & 0.8901 ± 0.0041 & 0.9069 ± 0.0061 & 0.8862 ± 0.0038 \\
 & PN-B-Y-S-E & 0.8570 ± 0.0033 & 0.8616 ± 0.0057 & 0.8941 ± 0.0030 & 0.8987 ± 0.0035 & 0.9074 ± 0.0061 & 0.9018 ± 0.0055 & 0.8868 ± 0.0045 \\
 & PN-Y & 0.8595 ± 0.0033 & 0.8585 ± 0.0035 & 0.8926 ± 0.0029 & 0.8997 ± 0.0030 & 0.9125 ± 0.0024 & 0.9120 ± 0.0039 & 0.8891 ± 0.0032 \\
 & PN-Y-S & 0.8534 ± 0.0068 & 0.8555 ± 0.0026 & 0.8906 ± 0.0053 & 0.8911 ± 0.0027 & 0.9038 ± 0.0066 & 0.8947 ± 0.0057 & 0.8815 ± 0.0049 \\
 & PN-Y-S-E & 0.8631 ± 0.0047 & 0.8784 ± 0.0057 & 0.8835 ± 0.0049 & 0.8962 ± 0.0045 & 0.9013 ± 0.0025 & 0.8997 ± 0.0051 & 0.8870 ± 0.0046 \\
\cline{1-9}
\bottomrule
\end{tabular}

}
\caption{Linear Probing results on PACS dataset.}\label{tab:lp_pacs}
\end{table}

\begin{table}[!htbp]
    \small
\scalebox{0.9}{
    \begin{tabular}{lllllllll}
\toprule
 &  & 0.10 & 0.20 & 0.40 & 0.60 & 0.80 & 1.00 & mean \\
$E_{t}$ & Method &  &  &  &  &  &  &  \\
\midrule
\multirow[t]{11}{*}{0} & DANN & 0.9620 ± 0.0042 & 0.9761 ± 0.0065 & 0.9831 ± 0.0053 & 0.9859 ± 0.0039 & 0.9930 ± 0.0022 & 0.9930 ± 0.0031 & 0.9822 ± 0.0042 \\
 & DIWA & 0.9873 ± 0.0052 & 0.9930 ± 0.0039 & 0.9972 ± 0.0028 & \textbf{1.0000 ± 0.0000} & 0.9887 ± 0.0053 & \textbf{1.0000 ± 0.0000} & 0.9944 ± 0.0029 \\
 & ERM & 0.9901 ± 0.0036 & 0.9873 ± 0.0034 & 0.9873 ± 0.0056 & 0.9775 ± 0.0068 & 0.9831 ± 0.0036 & 0.9972 ± 0.0017 & 0.9871 ± 0.0041 \\
 & NUC-0.01 & \textbf{1.0000 ± 0.0000} & \textbf{1.0000 ± 0.0000} & \textbf{1.0000 ± 0.0000} & \textbf{1.0000 ± 0.0000} & \textbf{1.0000 ± 0.0000} & 0.9944 ± 0.0056 & 0.9991 ± 0.0009 \\
 & NUC-0.1 & \textbf{1.0000 ± 0.0000} & \textbf{1.0000 ± 0.0000} & \textbf{1.0000 ± 0.0000} & \textbf{1.0000 ± 0.0000} & \textbf{1.0000 ± 0.0000} & \textbf{1.0000 ± 0.0000} & \textbf{1.0000 ± 0.0000} \\
 & PN-B & 0.9887 ± 0.0036 & 0.9930 ± 0.0022 & 0.9944 ± 0.0026 & 0.9972 ± 0.0017 & 0.9944 ± 0.0026 & 0.9986 ± 0.0014 & 0.9944 ± 0.0024 \\
 & PN-B-Y & 0.9901 ± 0.0017 & 0.9944 ± 0.0014 & 0.9930 ± 0.0022 & 0.9972 ± 0.0017 & 0.9972 ± 0.0017 & 0.9958 ± 0.0017 & 0.9946 ± 0.0018 \\
 & PN-B-Y-S-E & 0.9958 ± 0.0017 & 0.9944 ± 0.0026 & 0.9958 ± 0.0017 & 0.9944 ± 0.0014 & 0.9901 ± 0.0028 & 0.9972 ± 0.0017 & 0.9946 ± 0.0020 \\
 & PN-Y & 0.9958 ± 0.0028 & 0.9915 ± 0.0041 & 0.9915 ± 0.0026 & 0.9901 ± 0.0017 & 0.9986 ± 0.0014 & 0.9972 ± 0.0017 & 0.9941 ± 0.0024 \\
 & PN-Y-S & 0.9901 ± 0.0036 & 0.9845 ± 0.0041 & 0.9944 ± 0.0026 & 0.9930 ± 0.0000 & 0.9986 ± 0.0014 & 0.9915 ± 0.0026 & 0.9920 ± 0.0024 \\
 & PN-Y-S-E & 0.9944 ± 0.0026 & 0.9972 ± 0.0017 & 0.9972 ± 0.0017 & 0.9873 ± 0.0014 & 0.9915 ± 0.0026 & 0.9972 ± 0.0017 & 0.9941 ± 0.0020 \\
\cline{1-9}
\multirow[t]{11}{*}{1} & DANN & 0.5429 ± 0.0039 & 0.5526 ± 0.0034 & 0.6293 ± 0.0076 & 0.6391 ± 0.0160 & 0.6429 ± 0.0081 & 0.6368 ± 0.0089 & 0.6073 ± 0.0080 \\
 & DIWA & 0.7414 ± 0.0034 & 0.7504 ± 0.0102 & 0.7722 ± 0.0057 & \textbf{0.7729 ± 0.0070} & \textbf{0.7805 ± 0.0069} & \textbf{0.7992 ± 0.0096} & 0.7694 ± 0.0071 \\
 & ERM & \textbf{0.7519 ± 0.0083} & 0.7263 ± 0.0088 & 0.7549 ± 0.0082 & 0.7586 ± 0.0101 & 0.7707 ± 0.0090 & 0.7609 ± 0.0096 & 0.7539 ± 0.0090 \\
 & NUC-0.01 & 0.7331 ± 0.0117 & 0.7677 ± 0.0088 & 0.7692 ± 0.0094 & 0.7722 ± 0.0070 & 0.7602 ± 0.0038 & 0.7654 ± 0.0104 & 0.7613 ± 0.0085 \\
 & NUC-0.1 & 0.7331 ± 0.0092 & 0.7451 ± 0.0150 & 0.7654 ± 0.0128 & 0.7609 ± 0.0054 & 0.7549 ± 0.0060 & 0.7737 ± 0.0102 & 0.7555 ± 0.0098 \\
 & PN-B & 0.7113 ± 0.0066 & 0.7331 ± 0.0087 & 0.7669 ± 0.0110 & 0.7662 ± 0.0038 & 0.7421 ± 0.0044 & 0.7519 ± 0.0067 & 0.7452 ± 0.0069 \\
 & PN-B-Y & 0.7083 ± 0.0066 & 0.7617 ± 0.0059 & 0.7699 ± 0.0084 & 0.7729 ± 0.0076 & 0.7571 ± 0.0104 & 0.7707 ± 0.0072 & 0.7568 ± 0.0077 \\
 & PN-B-Y-S-E & 0.7248 ± 0.0069 & 0.7511 ± 0.0048 & 0.7767 ± 0.0035 & 0.7564 ± 0.0099 & 0.7481 ± 0.0128 & 0.7609 ± 0.0081 & 0.7530 ± 0.0077 \\
 & PN-Y & 0.7346 ± 0.0050 & \textbf{0.7797 ± 0.0130} & \textbf{0.7992 ± 0.0116} & 0.7677 ± 0.0129 & 0.7519 ± 0.0059 & 0.7842 ± 0.0102 & \textbf{0.7695 ± 0.0098} \\
 & PN-Y-S & 0.7203 ± 0.0059 & 0.7782 ± 0.0098 & 0.7714 ± 0.0077 & 0.7669 ± 0.0102 & 0.7466 ± 0.0112 & 0.7534 ± 0.0061 & 0.7561 ± 0.0085 \\
 & PN-Y-S-E & 0.7256 ± 0.0086 & 0.7774 ± 0.0129 & 0.7774 ± 0.0076 & 0.7571 ± 0.0137 & 0.7654 ± 0.0165 & 0.7511 ± 0.0069 & 0.7590 ± 0.0110 \\
\cline{1-9}
\multirow[t]{11}{*}{2} & DANN & 0.7927 ± 0.0080 & 0.8012 ± 0.0114 & 0.7817 ± 0.0060 & 0.7976 ± 0.0056 & 0.7854 ± 0.0053 & 0.8091 ± 0.0126 & 0.7946 ± 0.0081 \\
 & DIWA & \textbf{0.8201 ± 0.0075} & 0.8073 ± 0.0068 & 0.8189 ± 0.0039 & 0.8146 ± 0.0077 & 0.8049 ± 0.0066 & 0.8104 ± 0.0080 & 0.8127 ± 0.0067 \\
 & ERM & 0.8104 ± 0.0077 & \textbf{0.8268 ± 0.0046} & 0.8293 ± 0.0048 & 0.8152 ± 0.0105 & \textbf{0.8299 ± 0.0078} & 0.8311 ± 0.0124 & 0.8238 ± 0.0080 \\
 & NUC-0.01 & 0.8067 ± 0.0100 & 0.8213 ± 0.0050 & 0.8262 ± 0.0043 & \textbf{0.8329 ± 0.0079} & 0.8140 ± 0.0027 & 0.8189 ± 0.0079 & 0.8200 ± 0.0063 \\
 & NUC-0.1 & 0.8006 ± 0.0072 & 0.8262 ± 0.0061 & \textbf{0.8354 ± 0.0042} & 0.8274 ± 0.0091 & 0.8220 ± 0.0058 & \textbf{0.8341 ± 0.0059} & \textbf{0.8243 ± 0.0064} \\
 & PN-B & 0.7945 ± 0.0067 & 0.8030 ± 0.0055 & 0.8049 ± 0.0093 & 0.8152 ± 0.0033 & 0.8128 ± 0.0059 & 0.8073 ± 0.0047 & 0.8063 ± 0.0059 \\
 & PN-B-Y & 0.7823 ± 0.0100 & 0.7848 ± 0.0101 & 0.7915 ± 0.0030 & 0.8171 ± 0.0057 & 0.8165 ± 0.0054 & 0.8104 ± 0.0095 & 0.8004 ± 0.0073 \\
 & PN-B-Y-S-E & 0.8073 ± 0.0095 & 0.7970 ± 0.0051 & 0.7957 ± 0.0041 & 0.8067 ± 0.0058 & 0.7963 ± 0.0050 & 0.8165 ± 0.0113 & 0.8033 ± 0.0068 \\
 & PN-Y & 0.7982 ± 0.0118 & 0.7933 ± 0.0049 & 0.8073 ± 0.0053 & 0.8183 ± 0.0053 & 0.8079 ± 0.0042 & 0.8055 ± 0.0089 & 0.8051 ± 0.0067 \\
 & PN-Y-S & 0.7963 ± 0.0064 & 0.7866 ± 0.0059 & 0.8067 ± 0.0059 & 0.8030 ± 0.0062 & 0.8110 ± 0.0058 & 0.8293 ± 0.0042 & 0.8055 ± 0.0057 \\
 & PN-Y-S-E & 0.7957 ± 0.0062 & 0.7939 ± 0.0104 & 0.8024 ± 0.0090 & 0.8073 ± 0.0095 & 0.8213 ± 0.0074 & 0.8244 ± 0.0101 & 0.8075 ± 0.0088 \\
\cline{1-9}
\multirow[t]{11}{*}{3} & DANN & 0.5018 ± 0.0121 & 0.5302 ± 0.0022 & 0.5846 ± 0.0061 & 0.5840 ± 0.0012 & 0.5976 ± 0.0075 & 0.5716 ± 0.0037 & 0.5616 ± 0.0055 \\
 & DIWA & 0.8491 ± 0.0095 & 0.8728 ± 0.0047 & 0.8550 ± 0.0028 & 0.8651 ± 0.0039 & 0.8550 ± 0.0028 & 0.8580 ± 0.0048 & 0.8592 ± 0.0047 \\
 & ERM & 0.8604 ± 0.0069 & 0.8728 ± 0.0129 & 0.8722 ± 0.0036 & 0.8663 ± 0.0077 & 0.8787 ± 0.0066 & 0.8722 ± 0.0049 & 0.8704 ± 0.0071 \\
 & NUC-0.01 & 0.8456 ± 0.0045 & 0.8645 ± 0.0042 & 0.8817 ± 0.0021 & 0.8828 ± 0.0093 & 0.8846 ± 0.0099 & 0.8716 ± 0.0074 & 0.8718 ± 0.0062 \\
 & NUC-0.1 & 0.8467 ± 0.0071 & 0.8686 ± 0.0050 & 0.8757 ± 0.0039 & \textbf{0.8864 ± 0.0064} & \textbf{0.8947 ± 0.0047} & 0.8751 ± 0.0068 & \textbf{0.8746 ± 0.0056} \\
 & PN-B & 0.8609 ± 0.0047 & 0.8556 ± 0.0067 & 0.8740 ± 0.0050 & 0.8763 ± 0.0071 & 0.8657 ± 0.0061 & 0.8609 ± 0.0078 & 0.8656 ± 0.0062 \\
 & PN-B-Y & 0.8621 ± 0.0052 & 0.8556 ± 0.0044 & 0.8722 ± 0.0065 & 0.8675 ± 0.0076 & 0.8651 ± 0.0123 & 0.8734 ± 0.0029 & 0.8660 ± 0.0065 \\
 & PN-B-Y-S-E & \textbf{0.8728 ± 0.0039} & 0.8550 ± 0.0049 & 0.8574 ± 0.0055 & 0.8627 ± 0.0049 & 0.8680 ± 0.0058 & 0.8651 ± 0.0061 & 0.8635 ± 0.0052 \\
 & PN-Y & 0.8521 ± 0.0064 & \textbf{0.8781 ± 0.0054} & \textbf{0.8911 ± 0.0025} & 0.8598 ± 0.0050 & 0.8722 ± 0.0044 & \textbf{0.8769 ± 0.0056} & 0.8717 ± 0.0049 \\
 & PN-Y-S & 0.8669 ± 0.0085 & 0.8686 ± 0.0050 & 0.8710 ± 0.0091 & 0.8621 ± 0.0060 & 0.8805 ± 0.0070 & 0.8669 ± 0.0068 & 0.8693 ± 0.0071 \\
 & PN-Y-S-E & 0.8633 ± 0.0037 & 0.8692 ± 0.0041 & 0.8692 ± 0.0090 & 0.8615 ± 0.0059 & 0.8633 ± 0.0081 & 0.8746 ± 0.0030 & 0.8669 ± 0.0057 \\
\cline{1-9}
\bottomrule
\end{tabular}

}
\caption{Linear Probing results on VLCS dataset.}\label{tab:lp_vlcs}
\end{table}

\begin{table}[!htbp]
    \small
\scalebox{0.9}{
    \begin{tabular}{lllllllll}
\toprule
 &  & 0.10 & 0.20 & 0.40 & 0.60 & 0.80 & 1.00 & mean \\
$E_{t}$ & Method &  &  &  &  &  &  &  \\
\midrule
\multirow[t]{11}{*}{0} & DANN & 0.6823 ± 0.0044 & 0.7198 ± 0.0051 & 0.7684 ± 0.0049 & 0.7827 ± 0.0032 & 0.7852 ± 0.0052 & 0.7886 ± 0.0087 & 0.7545 ± 0.0053 \\
 & DIWA & 0.8249 ± 0.0058 & 0.8734 ± 0.0037 & 0.9055 ± 0.0062 & 0.9291 ± 0.0025 & 0.9422 ± 0.0076 & 0.9418 ± 0.0020 & 0.9028 ± 0.0046 \\
 & ERM & 0.8346 ± 0.0055 & 0.8616 ± 0.0064 & 0.9084 ± 0.0028 & 0.9329 ± 0.0043 & 0.9367 ± 0.0032 & 0.9397 ± 0.0024 & 0.9023 ± 0.0041 \\
 & NUC-0.01 & 0.7928 ± 0.0052 & 0.8329 ± 0.0056 & 0.8730 ± 0.0059 & 0.9105 ± 0.0021 & 0.9219 ± 0.0020 & 0.9291 ± 0.0037 & 0.8767 ± 0.0041 \\
 & NUC-0.1 & 0.7975 ± 0.0110 & 0.8409 ± 0.0054 & 0.8624 ± 0.0039 & 0.9110 ± 0.0051 & 0.9139 ± 0.0036 & 0.9295 ± 0.0044 & 0.8759 ± 0.0055 \\
 & PN-B & 0.8278 ± 0.0058 & \textbf{0.8844 ± 0.0045} & \textbf{0.9211 ± 0.0011} & 0.9405 ± 0.0018 & \textbf{0.9578 ± 0.0009} & 0.9603 ± 0.0044 & 0.9153 ± 0.0031 \\
 & PN-B-Y & 0.8253 ± 0.0049 & 0.8700 ± 0.0049 & 0.9135 ± 0.0049 & 0.9473 ± 0.0033 & 0.9502 ± 0.0023 & 0.9603 ± 0.0035 & 0.9111 ± 0.0040 \\
 & PN-B-Y-S-E & 0.8295 ± 0.0042 & 0.8819 ± 0.0024 & 0.9181 ± 0.0037 & 0.9405 ± 0.0020 & 0.9532 ± 0.0018 & 0.9553 ± 0.0017 & 0.9131 ± 0.0026 \\
 & PN-Y & \textbf{0.8392 ± 0.0041} & 0.8819 ± 0.0035 & 0.9131 ± 0.0012 & 0.9540 ± 0.0026 & 0.9519 ± 0.0026 & \textbf{0.9608 ± 0.0032} & 0.9168 ± 0.0029 \\
 & PN-Y-S & 0.8287 ± 0.0044 & 0.8747 ± 0.0051 & 0.9165 ± 0.0011 & 0.9502 ± 0.0029 & 0.9506 ± 0.0041 & 0.9578 ± 0.0019 & 0.9131 ± 0.0032 \\
 & PN-Y-S-E & 0.8300 ± 0.0032 & 0.8844 ± 0.0041 & 0.9190 ± 0.0061 & \textbf{0.9586 ± 0.0047} & 0.9515 ± 0.0012 & 0.9586 ± 0.0029 & \textbf{0.9170 ± 0.0037} \\
\cline{1-9}
\multirow[t]{11}{*}{1} & DANN & 0.6875 ± 0.0026 & 0.7294 ± 0.0037 & 0.7760 ± 0.0028 & 0.7690 ± 0.0067 & 0.7727 ± 0.0052 & 0.7729 ± 0.0038 & 0.7512 ± 0.0041 \\
 & DIWA & \textbf{0.8405 ± 0.0027} & 0.8692 ± 0.0027 & 0.8893 ± 0.0016 & 0.8955 ± 0.0047 & 0.9181 ± 0.0018 & 0.9281 ± 0.0026 & 0.8901 ± 0.0027 \\
 & ERM & 0.8351 ± 0.0041 & 0.8643 ± 0.0016 & 0.8916 ± 0.0016 & 0.9064 ± 0.0022 & \textbf{0.9341 ± 0.0026} & 0.9300 ± 0.0022 & 0.8936 ± 0.0024 \\
 & NUC-0.01 & 0.7947 ± 0.0033 & 0.8246 ± 0.0038 & 0.8561 ± 0.0030 & 0.8836 ± 0.0021 & 0.8949 ± 0.0030 & 0.8998 ± 0.0016 & 0.8589 ± 0.0028 \\
 & NUC-0.1 & 0.7951 ± 0.0040 & 0.8366 ± 0.0059 & 0.8690 ± 0.0039 & 0.8852 ± 0.0045 & 0.8869 ± 0.0033 & 0.9008 ± 0.0024 & 0.8623 ± 0.0040 \\
 & PN-B & 0.8310 ± 0.0061 & \textbf{0.8713 ± 0.0044} & \textbf{0.9074 ± 0.0013} & 0.9117 ± 0.0019 & 0.9220 ± 0.0021 & 0.9347 ± 0.0030 & \textbf{0.8963 ± 0.0031} \\
 & PN-B-Y & 0.8302 ± 0.0066 & 0.8622 ± 0.0026 & 0.8996 ± 0.0037 & 0.9140 ± 0.0043 & 0.9300 ± 0.0016 & \textbf{0.9398 ± 0.0036} & 0.8960 ± 0.0037 \\
 & PN-B-Y-S-E & 0.8267 ± 0.0050 & 0.8637 ± 0.0025 & 0.8922 ± 0.0029 & 0.9156 ± 0.0031 & 0.9296 ± 0.0047 & 0.9368 ± 0.0015 & 0.8941 ± 0.0033 \\
 & PN-Y & 0.8248 ± 0.0036 & 0.8550 ± 0.0047 & 0.8895 ± 0.0021 & \textbf{0.9179 ± 0.0040} & 0.9290 ± 0.0011 & 0.9326 ± 0.0037 & 0.8915 ± 0.0032 \\
 & PN-Y-S & 0.8197 ± 0.0053 & 0.8544 ± 0.0031 & 0.8926 ± 0.0015 & 0.9117 ± 0.0037 & 0.9257 ± 0.0023 & 0.9304 ± 0.0026 & 0.8891 ± 0.0031 \\
 & PN-Y-S-E & 0.8230 ± 0.0028 & 0.8706 ± 0.0021 & 0.8834 ± 0.0033 & 0.9166 ± 0.0055 & 0.9324 ± 0.0020 & 0.9347 ± 0.0011 & 0.8935 ± 0.0028 \\
\cline{1-9}
\multirow[t]{11}{*}{2} & DANN & 0.5275 ± 0.0066 & 0.6045 ± 0.0061 & 0.6453 ± 0.0065 & 0.6403 ± 0.0081 & 0.6665 ± 0.0056 & 0.6690 ± 0.0043 & 0.6255 ± 0.0062 \\
 & DIWA & 0.7264 ± 0.0102 & 0.7909 ± 0.0046 & 0.8040 ± 0.0039 & 0.8458 ± 0.0062 & 0.8499 ± 0.0091 & 0.8872 ± 0.0045 & 0.8174 ± 0.0064 \\
 & ERM & 0.7330 ± 0.0112 & 0.7864 ± 0.0084 & 0.8317 ± 0.0043 & 0.8640 ± 0.0044 & 0.8831 ± 0.0055 & 0.8912 ± 0.0029 & 0.8316 ± 0.0061 \\
 & NUC-0.01 & 0.6448 ± 0.0048 & 0.7380 ± 0.0064 & 0.7889 ± 0.0045 & 0.8388 ± 0.0066 & 0.8589 ± 0.0071 & 0.8615 ± 0.0048 & 0.7885 ± 0.0057 \\
 & NUC-0.1 & 0.6534 ± 0.0067 & 0.7345 ± 0.0101 & 0.7950 ± 0.0103 & 0.8368 ± 0.0049 & 0.8615 ± 0.0063 & 0.8680 ± 0.0030 & 0.7915 ± 0.0069 \\
 & PN-B & \textbf{0.7521 ± 0.0081} & \textbf{0.7945 ± 0.0082} & 0.8247 ± 0.0059 & 0.8443 ± 0.0048 & 0.8675 ± 0.0017 & 0.8811 ± 0.0036 & 0.8274 ± 0.0054 \\
 & PN-B-Y & 0.7436 ± 0.0040 & 0.7884 ± 0.0069 & 0.8327 ± 0.0068 & 0.8630 ± 0.0083 & 0.8670 ± 0.0066 & 0.8922 ± 0.0054 & 0.8312 ± 0.0063 \\
 & PN-B-Y-S-E & 0.7345 ± 0.0063 & 0.7904 ± 0.0058 & 0.8277 ± 0.0027 & \textbf{0.8670 ± 0.0063} & 0.8856 ± 0.0047 & 0.8992 ± 0.0047 & 0.8341 ± 0.0051 \\
 & PN-Y & 0.7295 ± 0.0044 & 0.7748 ± 0.0089 & 0.8322 ± 0.0045 & 0.8620 ± 0.0059 & 0.8831 ± 0.0033 & \textbf{0.9103 ± 0.0029} & 0.8320 ± 0.0050 \\
 & PN-Y-S & 0.7325 ± 0.0073 & 0.7904 ± 0.0079 & \textbf{0.8368 ± 0.0106} & 0.8660 ± 0.0037 & 0.8836 ± 0.0075 & 0.9008 ± 0.0023 & \textbf{0.8350 ± 0.0065} \\
 & PN-Y-S-E & 0.7451 ± 0.0070 & 0.7788 ± 0.0039 & 0.8287 ± 0.0046 & 0.8610 ± 0.0084 & \textbf{0.8932 ± 0.0060} & 0.8957 ± 0.0044 & 0.8338 ± 0.0057 \\
\cline{1-9}
\multirow[t]{11}{*}{3} & DANN & 0.4942 ± 0.0087 & 0.5143 ± 0.0048 & 0.5014 ± 0.0034 & 0.5099 ± 0.0098 & 0.5218 ± 0.0020 & 0.5296 ± 0.0069 & 0.5118 ± 0.0059 \\
 & DIWA & 0.7156 ± 0.0030 & 0.7340 ± 0.0052 & 0.8116 ± 0.0023 & 0.8255 ± 0.0030 & 0.8616 ± 0.0036 & 0.8810 ± 0.0030 & 0.8049 ± 0.0034 \\
 & ERM & 0.7367 ± 0.0067 & 0.7554 ± 0.0093 & 0.8119 ± 0.0053 & 0.8265 ± 0.0045 & 0.8527 ± 0.0055 & 0.8779 ± 0.0010 & 0.8102 ± 0.0054 \\
 & NUC-0.01 & 0.6442 ± 0.0070 & 0.7061 ± 0.0072 & 0.7687 ± 0.0062 & 0.8122 ± 0.0052 & 0.8337 ± 0.0036 & 0.8456 ± 0.0033 & 0.7684 ± 0.0054 \\
 & NUC-0.1 & 0.6633 ± 0.0061 & 0.7139 ± 0.0055 & 0.7759 ± 0.0069 & 0.8027 ± 0.0046 & 0.8340 ± 0.0025 & 0.8442 ± 0.0053 & 0.7723 ± 0.0051 \\
 & PN-B & 0.7310 ± 0.0068 & 0.7735 ± 0.0060 & 0.8221 ± 0.0049 & 0.8398 ± 0.0062 & 0.8660 ± 0.0029 & 0.8810 ± 0.0074 & 0.8189 ± 0.0057 \\
 & PN-B-Y & 0.7241 ± 0.0069 & 0.7701 ± 0.0024 & \textbf{0.8265 ± 0.0038} & 0.8507 ± 0.0054 & 0.8724 ± 0.0058 & 0.8827 ± 0.0041 & 0.8211 ± 0.0047 \\
 & PN-B-Y-S-E & \textbf{0.7446 ± 0.0054} & 0.7721 ± 0.0037 & 0.8173 ± 0.0032 & \textbf{0.8520 ± 0.0031} & \textbf{0.8861 ± 0.0061} & 0.8857 ± 0.0056 & \textbf{0.8263 ± 0.0045} \\
 & PN-Y & 0.7357 ± 0.0037 & 0.7748 ± 0.0042 & 0.8126 ± 0.0065 & 0.8500 ± 0.0037 & 0.8680 ± 0.0046 & 0.8847 ± 0.0041 & 0.8210 ± 0.0045 \\
 & PN-Y-S & 0.7391 ± 0.0058 & 0.7568 ± 0.0030 & 0.8184 ± 0.0070 & 0.8463 ± 0.0067 & 0.8694 ± 0.0082 & 0.8816 ± 0.0013 & 0.8186 ± 0.0053 \\
 & PN-Y-S-E & 0.7432 ± 0.0031 & \textbf{0.7823 ± 0.0065} & 0.7980 ± 0.0028 & 0.8432 ± 0.0010 & 0.8782 ± 0.0049 & \textbf{0.8898 ± 0.0032} & 0.8224 ± 0.0036 \\
\cline{1-9}
\bottomrule
\end{tabular}

}
\caption{Linear Probing results on Terrain-Cognita dataset.}\label{tab:lp_terraincognita}
\end{table}

\newpage
\subsection{Target Finetuning Results}

We use the source pretrained feature encoder and classifier as the initialization for the target finetuning. We use the same hyperparameter range as the source pretraining. The best model is choosen based on the target validation accuracy. We report the results in the following \autoref{tab:tf_officehome}, \autoref{tab:tf_pacs}, \autoref{tab:tf_vlcs}, \autoref{tab:tf_terraincognita}.

\begin{table}[H]
    \small
\scalebox{0.9}{
    \begin{tabular}{lllllllll}
\toprule
 &  & 0.10 & 0.20 & 0.40 & 0.60 & 0.80 & 1.00 & mean \\
$E_{t}$ & Method &  &  &  &  &  &  &  \\
\midrule
\multirow[t]{8}{*}{0} & DANN & 0.6263 ± 0.0091 & 0.6650 ± 0.0098 & 0.7251 ± 0.0062 & 0.7588 ± 0.0051 & 0.7786 ± 0.0035 & 0.7984 ± 0.0045 & 0.7254 ± 0.0064 \\
 & DIWA & 0.6329 ± 0.0423 & 0.6872 ± 0.0194 & 0.7366 ± 0.0069 & 0.7514 ± 0.0075 & 0.7630 ± 0.0119 & 0.8148 ± 0.0054 & 0.7310 ± 0.0156 \\
 & ERM & 0.7407 ± 0.0079 & 0.7786 ± 0.0054 & 0.8000 ± 0.0042 & 0.8272 ± 0.0034 & 0.8362 ± 0.0040 & 0.8593 ± 0.0033 & 0.8070 ± 0.0047 \\
 & NUC-0.01 & 0.4535 ± 0.0069 & 0.6049 ± 0.0125 & 0.7103 ± 0.0065 & 0.7646 ± 0.0066 & 0.7770 ± 0.0067 & 0.8148 ± 0.0072 & 0.6875 ± 0.0077 \\
 & NUC-0.1 & 0.4461 ± 0.0051 & 0.6058 ± 0.0129 & 0.7029 ± 0.0073 & 0.7556 ± 0.0067 & 0.7852 ± 0.0067 & 0.8173 ± 0.0108 & 0.6855 ± 0.0083 \\
 & PN-Y & 0.7588 ± 0.0100 & 0.7893 ± 0.0086 & 0.8082 ± 0.0059 & \textbf{0.8305 ± 0.0030} & \textbf{0.8519 ± 0.0037} & \textbf{0.8675 ± 0.0038} & 0.8177 ± 0.0058 \\
 & PN-Y-S & \textbf{0.7621 ± 0.0035} & \textbf{0.7926 ± 0.0059} & \textbf{0.8239 ± 0.0049} & 0.8247 ± 0.0050 & 0.8502 ± 0.0046 & 0.8568 ± 0.0015 & \textbf{0.8184 ± 0.0043} \\
 & PN-Y-S-E & 0.7605 ± 0.0082 & 0.7819 ± 0.0062 & 0.8140 ± 0.0072 & 0.8296 ± 0.0053 & 0.8510 ± 0.0044 & 0.8601 ± 0.0041 & 0.8162 ± 0.0059 \\
\cline{1-9}
\multirow[t]{8}{*}{1} & DANN & 0.4641 ± 0.0030 & 0.6105 ± 0.0124 & 0.7382 ± 0.0036 & 0.7922 ± 0.0048 & 0.8110 ± 0.0079 & 0.8352 ± 0.0049 & 0.7085 ± 0.0061 \\
 & DIWA & 0.6229 ± 0.0045 & \textbf{0.7121 ± 0.0170} & 0.7510 ± 0.0160 & 0.8018 ± 0.0122 & 0.8110 ± 0.0114 & 0.8384 ± 0.0063 & 0.7562 ± 0.0113 \\
 & ERM & \textbf{0.6526 ± 0.0100} & 0.7062 ± 0.0036 & \textbf{0.7629 ± 0.0072} & 0.7931 ± 0.0095 & 0.8220 ± 0.0073 & 0.8279 ± 0.0039 & 0.7608 ± 0.0069 \\
 & NUC-0.01 & 0.4288 ± 0.0057 & 0.5922 ± 0.0107 & 0.7231 ± 0.0064 & 0.7844 ± 0.0085 & 0.8165 ± 0.0018 & \textbf{0.8426 ± 0.0063} & 0.6979 ± 0.0066 \\
 & NUC-0.1 & 0.4279 ± 0.0051 & 0.5931 ± 0.0103 & 0.7199 ± 0.0067 & 0.7945 ± 0.0039 & 0.8151 ± 0.0081 & 0.8375 ± 0.0069 & 0.6980 ± 0.0068 \\
 & PN-Y & 0.6526 ± 0.0028 & 0.7053 ± 0.0100 & 0.7616 ± 0.0041 & \textbf{0.8087 ± 0.0043} & \textbf{0.8247 ± 0.0006} & 0.8412 ± 0.0020 & \textbf{0.7657 ± 0.0040} \\
 & PN-Y-S & 0.6293 ± 0.0051 & 0.6783 ± 0.0107 & 0.7442 ± 0.0033 & 0.7858 ± 0.0047 & 0.8014 ± 0.0051 & 0.8183 ± 0.0020 & 0.7429 ± 0.0052 \\
 & PN-Y-S-E & 0.6302 ± 0.0062 & 0.6870 ± 0.0089 & 0.7368 ± 0.0067 & 0.7785 ± 0.0043 & 0.8046 ± 0.0028 & 0.8183 ± 0.0025 & 0.7426 ± 0.0052 \\
\cline{1-9}
\multirow[t]{8}{*}{2} & DANN & 0.7770 ± 0.0091 & 0.8450 ± 0.0053 & 0.8986 ± 0.0050 & 0.9099 ± 0.0043 & 0.9252 ± 0.0033 & 0.9365 ± 0.0024 & 0.8821 ± 0.0049 \\
 & DIWA & 0.7856 ± 0.0131 & 0.8568 ± 0.0081 & 0.8878 ± 0.0066 & 0.9072 ± 0.0058 & 0.9207 ± 0.0086 & 0.9275 ± 0.0061 & 0.8809 ± 0.0080 \\
 & ERM & \textbf{0.8572 ± 0.0201} & \textbf{0.8851 ± 0.0150} & \textbf{0.9068 ± 0.0082} & \textbf{0.9221 ± 0.0074} & \textbf{0.9342 ± 0.0011} & 0.9360 ± 0.0039 & \textbf{0.9069 ± 0.0093} \\
 & NUC-0.01 & 0.6923 ± 0.0082 & 0.8059 ± 0.0063 & 0.8820 ± 0.0031 & 0.9113 ± 0.0057 & 0.9324 ± 0.0042 & 0.9315 ± 0.0009 & 0.8592 ± 0.0047 \\
 & NUC-0.1 & 0.6995 ± 0.0058 & 0.8054 ± 0.0062 & 0.8838 ± 0.0036 & 0.9122 ± 0.0038 & 0.9342 ± 0.0037 & 0.9356 ± 0.0023 & 0.8618 ± 0.0042 \\
 & PN-Y & 0.8266 ± 0.0124 & 0.8563 ± 0.0108 & 0.9023 ± 0.0049 & 0.9149 ± 0.0029 & 0.9324 ± 0.0031 & \textbf{0.9419 ± 0.0030} & 0.8957 ± 0.0062 \\
 & PN-Y-S & 0.8176 ± 0.0117 & 0.8437 ± 0.0080 & 0.8919 ± 0.0064 & 0.8991 ± 0.0034 & 0.9207 ± 0.0028 & 0.9279 ± 0.0020 & 0.8835 ± 0.0057 \\
 & PN-Y-S-E & 0.8288 ± 0.0205 & 0.8455 ± 0.0054 & 0.8842 ± 0.0031 & 0.9063 ± 0.0034 & 0.9185 ± 0.0039 & 0.9270 ± 0.0034 & 0.8851 ± 0.0066 \\
\cline{1-9}
\multirow[t]{8}{*}{3} & DANN & 0.7440 ± 0.0097 & 0.8128 ± 0.0080 & 0.8339 ± 0.0034 & 0.8477 ± 0.0039 & 0.8463 ± 0.0036 & 0.8537 ± 0.0017 & 0.8231 ± 0.0050 \\
 & DIWA & 0.7972 ± 0.0120 & 0.8179 ± 0.0106 & 0.8399 ± 0.0090 & 0.8151 ± 0.0147 & 0.8349 ± 0.0143 & 0.8339 ± 0.0095 & 0.8232 ± 0.0117 \\
 & ERM & 0.8335 ± 0.0067 & 0.8541 ± 0.0033 & 0.8596 ± 0.0020 & 0.8578 ± 0.0025 & 0.8624 ± 0.0065 & 0.8638 ± 0.0030 & 0.8552 ± 0.0040 \\
 & NUC-0.01 & 0.6445 ± 0.0087 & 0.7583 ± 0.0058 & 0.8266 ± 0.0037 & 0.8381 ± 0.0080 & 0.8495 ± 0.0064 & 0.8596 ± 0.0051 & 0.7961 ± 0.0063 \\
 & NUC-0.1 & 0.6450 ± 0.0098 & 0.7592 ± 0.0032 & 0.8174 ± 0.0041 & 0.8408 ± 0.0059 & 0.8431 ± 0.0084 & 0.8472 ± 0.0083 & 0.7921 ± 0.0066 \\
 & PN-Y & 0.8564 ± 0.0065 & 0.8651 ± 0.0048 & 0.8780 ± 0.0045 & 0.8789 ± 0.0011 & 0.8798 ± 0.0021 & 0.8743 ± 0.0031 & 0.8721 ± 0.0037 \\
 & PN-Y-S & \textbf{0.8596 ± 0.0038} & \textbf{0.8697 ± 0.0052} & \textbf{0.8803 ± 0.0032} & 0.8771 ± 0.0039 & \textbf{0.8849 ± 0.0028} & 0.8775 ± 0.0041 & \textbf{0.8748 ± 0.0038} \\
 & PN-Y-S-E & 0.8537 ± 0.0071 & 0.8647 ± 0.0031 & 0.8771 ± 0.0037 & \textbf{0.8803 ± 0.0034} & 0.8826 ± 0.0029 & \textbf{0.8817 ± 0.0041} & 0.8733 ± 0.0041 \\
\cline{1-9}
\bottomrule
\end{tabular}

}
\caption{Target Finetuning results on OfficeHome dataset.}\label{tab:tf_officehome}
\end{table}

\begin{table}[!htbp]
    \small
\scalebox{0.9}{
    \begin{tabular}{lllllllll}
\toprule
 &  & 0.10 & 0.20 & 0.40 & 0.60 & 0.80 & 1.00 & mean \\
$E_{t}$ & Method &  &  &  &  &  &  &  \\
\midrule
\multirow[t]{8}{*}{0} & DANN & 0.7132 ± 0.0216 & 0.8166 ± 0.0138 & 0.8644 ± 0.0079 & 0.8946 ± 0.0113 & 0.9063 ± 0.0036 & 0.9112 ± 0.0064 & 0.8511 ± 0.0108 \\
 & DIWA & \textbf{0.9454 ± 0.0061} & \textbf{0.9463 ± 0.0067} & \textbf{0.9541 ± 0.0040} & 0.9580 ± 0.0057 & 0.9571 ± 0.0066 & 0.9649 ± 0.0056 & 0.9543 ± 0.0058 \\
 & ERM & 0.9190 ± 0.0025 & 0.9356 ± 0.0036 & 0.9483 ± 0.0025 & 0.9620 ± 0.0028 & 0.9590 ± 0.0043 & 0.9512 ± 0.0034 & 0.9459 ± 0.0032 \\
 & NUC-0.01 & 0.8878 ± 0.0072 & 0.9102 ± 0.0045 & 0.9307 ± 0.0066 & 0.9483 ± 0.0025 & 0.9571 ± 0.0028 & 0.9541 ± 0.0037 & 0.9314 ± 0.0046 \\
 & NUC-0.1 & 0.8820 ± 0.0062 & 0.9141 ± 0.0045 & 0.9395 ± 0.0080 & 0.9502 ± 0.0024 & 0.9473 ± 0.0056 & 0.9551 ± 0.0036 & 0.9314 ± 0.0051 \\
 & PN-Y & 0.9395 ± 0.0025 & 0.9463 ± 0.0038 & 0.9541 ± 0.0020 & 0.9600 ± 0.0032 & 0.9639 ± 0.0043 & 0.9698 ± 0.0028 & \textbf{0.9556 ± 0.0031} \\
 & PN-Y-S & 0.9298 ± 0.0045 & 0.9415 ± 0.0034 & 0.9532 ± 0.0033 & \textbf{0.9639 ± 0.0029} & \textbf{0.9659 ± 0.0031} & \textbf{0.9707 ± 0.0027} & 0.9541 ± 0.0033 \\
 & PN-Y-S-E & 0.9366 ± 0.0034 & 0.9346 ± 0.0072 & 0.9493 ± 0.0020 & 0.9590 ± 0.0040 & 0.9600 ± 0.0036 & 0.9620 ± 0.0024 & 0.9502 ± 0.0038 \\
\cline{1-9}
\multirow[t]{8}{*}{1} & DANN & 0.7812 ± 0.0120 & 0.8573 ± 0.0156 & 0.9043 ± 0.0118 & 0.9299 ± 0.0072 & 0.9419 ± 0.0074 & 0.9513 ± 0.0035 & 0.8943 ± 0.0096 \\
 & DIWA & \textbf{0.9342 ± 0.0058} & \textbf{0.9453 ± 0.0053} & \textbf{0.9667 ± 0.0025} & \textbf{0.9701 ± 0.0030} & \textbf{0.9735 ± 0.0016} & \textbf{0.9726 ± 0.0022} & \textbf{0.9604 ± 0.0034} \\
 & ERM & 0.8940 ± 0.0041 & 0.9051 ± 0.0031 & 0.9350 ± 0.0041 & 0.9453 ± 0.0034 & 0.9538 ± 0.0028 & 0.9538 ± 0.0021 & 0.9312 ± 0.0033 \\
 & NUC-0.01 & 0.8299 ± 0.0133 & 0.8846 ± 0.0081 & 0.9179 ± 0.0028 & 0.9274 ± 0.0056 & 0.9453 ± 0.0031 & 0.9538 ± 0.0055 & 0.9098 ± 0.0064 \\
 & NUC-0.1 & 0.8291 ± 0.0158 & 0.8829 ± 0.0068 & 0.9179 ± 0.0048 & 0.9376 ± 0.0064 & 0.9402 ± 0.0049 & 0.9547 ± 0.0052 & 0.9104 ± 0.0073 \\
 & PN-Y & 0.9197 ± 0.0051 & 0.9325 ± 0.0096 & 0.9530 ± 0.0014 & 0.9590 ± 0.0010 & 0.9632 ± 0.0017 & 0.9658 ± 0.0019 & 0.9489 ± 0.0035 \\
 & PN-Y-S & 0.9051 ± 0.0102 & 0.9214 ± 0.0100 & 0.9479 ± 0.0021 & 0.9496 ± 0.0016 & 0.9564 ± 0.0031 & 0.9607 ± 0.0034 & 0.9402 ± 0.0051 \\
 & PN-Y-S-E & 0.9051 ± 0.0083 & 0.9256 ± 0.0067 & 0.9479 ± 0.0034 & 0.9521 ± 0.0021 & 0.9556 ± 0.0029 & 0.9581 ± 0.0025 & 0.9407 ± 0.0043 \\
\cline{1-9}
\multirow[t]{8}{*}{2} & DANN & 0.8970 ± 0.0179 & 0.9497 ± 0.0094 & 0.9760 ± 0.0042 & 0.9808 ± 0.0022 & 0.9796 ± 0.0031 & 0.9808 ± 0.0069 & 0.9607 ± 0.0073 \\
 & DIWA & \textbf{0.9928 ± 0.0022} & \textbf{0.9928 ± 0.0035} & \textbf{0.9976 ± 0.0015} & \textbf{0.9988 ± 0.0012} & \textbf{0.9988 ± 0.0012} & \textbf{0.9976 ± 0.0015} & \textbf{0.9964 ± 0.0018} \\
 & ERM & 0.9868 ± 0.0035 & 0.9892 ± 0.0035 & 0.9952 ± 0.0012 & 0.9940 ± 0.0033 & 0.9952 ± 0.0029 & 0.9964 ± 0.0015 & 0.9928 ± 0.0026 \\
 & NUC-0.01 & 0.9665 ± 0.0031 & 0.9760 ± 0.0054 & 0.9880 ± 0.0027 & 0.9856 ± 0.0041 & 0.9892 ± 0.0029 & 0.9868 ± 0.0029 & 0.9820 ± 0.0035 \\
 & NUC-0.1 & 0.9665 ± 0.0031 & 0.9784 ± 0.0041 & 0.9880 ± 0.0027 & 0.9856 ± 0.0015 & 0.9880 ± 0.0019 & 0.9868 ± 0.0044 & 0.9822 ± 0.0029 \\
 & PN-Y & 0.9880 ± 0.0000 & 0.9880 ± 0.0000 & 0.9940 ± 0.0019 & 0.9916 ± 0.0015 & 0.9904 ± 0.0015 & 0.9916 ± 0.0015 & 0.9906 ± 0.0010 \\
 & PN-Y-S & 0.9844 ± 0.0024 & \textbf{0.9928 ± 0.0022} & 0.9892 ± 0.0022 & 0.9940 ± 0.0019 & 0.9928 ± 0.0012 & 0.9964 ± 0.0024 & 0.9916 ± 0.0021 \\
 & PN-Y-S-E & 0.9868 ± 0.0029 & 0.9916 ± 0.0024 & 0.9928 ± 0.0012 & 0.9952 ± 0.0022 & 0.9940 ± 0.0000 & 0.9928 ± 0.0012 & 0.9922 ± 0.0017 \\
\cline{1-9}
\multirow[t]{8}{*}{3} & DANN & 0.8336 ± 0.0041 & 0.8804 ± 0.0049 & 0.9150 ± 0.0061 & 0.9272 ± 0.0040 & 0.9298 ± 0.0029 & 0.9496 ± 0.0010 & 0.9059 ± 0.0038 \\
 & DIWA & \textbf{0.9323 ± 0.0041} & \textbf{0.9298 ± 0.0048} & \textbf{0.9399 ± 0.0035} & \textbf{0.9537 ± 0.0023} & \textbf{0.9580 ± 0.0043} & 0.9552 ± 0.0033 & \textbf{0.9448 ± 0.0037} \\
 & ERM & 0.8758 ± 0.0099 & 0.9135 ± 0.0029 & 0.9328 ± 0.0061 & 0.9486 ± 0.0055 & 0.9506 ± 0.0013 & \textbf{0.9557 ± 0.0013} & 0.9295 ± 0.0045 \\
 & NUC-0.01 & 0.7847 ± 0.0119 & 0.8687 ± 0.0076 & 0.9053 ± 0.0067 & 0.9170 ± 0.0037 & 0.9288 ± 0.0048 & 0.9338 ± 0.0065 & 0.8897 ± 0.0069 \\
 & NUC-0.1 & 0.7817 ± 0.0111 & 0.8687 ± 0.0021 & 0.8952 ± 0.0038 & 0.9150 ± 0.0046 & 0.9277 ± 0.0036 & 0.9450 ± 0.0034 & 0.8889 ± 0.0047 \\
 & PN-Y & 0.9043 ± 0.0013 & 0.9216 ± 0.0028 & 0.9257 ± 0.0019 & 0.9389 ± 0.0055 & 0.9486 ± 0.0026 & 0.9476 ± 0.0037 & 0.9311 ± 0.0030 \\
 & PN-Y-S & 0.8957 ± 0.0030 & 0.9099 ± 0.0022 & 0.9277 ± 0.0025 & 0.9384 ± 0.0062 & 0.9471 ± 0.0031 & 0.9466 ± 0.0031 & 0.9276 ± 0.0033 \\
 & PN-Y-S-E & 0.8967 ± 0.0039 & 0.9104 ± 0.0046 & 0.9308 ± 0.0035 & 0.9379 ± 0.0047 & 0.9461 ± 0.0035 & 0.9496 ± 0.0052 & 0.9286 ± 0.0043 \\
\cline{1-9}
\bottomrule
\end{tabular}

}
\caption{Target Finetuning results on PACS dataset.}\label{tab:tf_pacs}
\end{table}

\begin{table}[!htbp]
    \small
\scalebox{0.9}{
    \begin{tabular}{lllllllll}
\toprule
 &  & 0.10 & 0.20 & 0.40 & 0.60 & 0.80 & 1.00 & mean \\
$E_{t}$ & Method &  &  &  &  &  &  &  \\
\midrule
\multirow[t]{8}{*}{0} & DANN & \textbf{1.0000 ± 0.0000} & \textbf{1.0000 ± 0.0000} & \textbf{1.0000 ± 0.0000} & \textbf{1.0000 ± 0.0000} & \textbf{1.0000 ± 0.0000} & \textbf{1.0000 ± 0.0000} & \textbf{1.0000 ± 0.0000} \\
 & DIWA & \textbf{1.0000 ± 0.0000} & \textbf{1.0000 ± 0.0000} & \textbf{1.0000 ± 0.0000} & \textbf{1.0000 ± 0.0000} & \textbf{1.0000 ± 0.0000} & \textbf{1.0000 ± 0.0000} & \textbf{1.0000 ± 0.0000} \\
 & ERM & 0.9915 ± 0.0026 & \textbf{1.0000 ± 0.0000} & \textbf{1.0000 ± 0.0000} & \textbf{1.0000 ± 0.0000} & \textbf{1.0000 ± 0.0000} & \textbf{1.0000 ± 0.0000} & 0.9986 ± 0.0004 \\
 & NUC-0.01 & 0.9930 ± 0.0022 & 0.9915 ± 0.0026 & 0.9972 ± 0.0017 & 0.9986 ± 0.0014 & 0.9958 ± 0.0017 & \textbf{1.0000 ± 0.0000} & 0.9960 ± 0.0016 \\
 & NUC-0.1 & 0.9930 ± 0.0022 & 0.9930 ± 0.0022 & 0.9972 ± 0.0017 & 0.9986 ± 0.0014 & 0.9958 ± 0.0017 & \textbf{1.0000 ± 0.0000} & 0.9962 ± 0.0016 \\
 & PN-Y & 0.9958 ± 0.0028 & \textbf{1.0000 ± 0.0000} & \textbf{1.0000 ± 0.0000} & \textbf{1.0000 ± 0.0000} & \textbf{1.0000 ± 0.0000} & \textbf{1.0000 ± 0.0000} & 0.9993 ± 0.0005 \\
 & PN-Y-S & 0.9958 ± 0.0017 & \textbf{1.0000 ± 0.0000} & \textbf{1.0000 ± 0.0000} & \textbf{1.0000 ± 0.0000} & \textbf{1.0000 ± 0.0000} & \textbf{1.0000 ± 0.0000} & 0.9993 ± 0.0003 \\
 & PN-Y-S-E & 0.9972 ± 0.0017 & \textbf{1.0000 ± 0.0000} & \textbf{1.0000 ± 0.0000} & \textbf{1.0000 ± 0.0000} & \textbf{1.0000 ± 0.0000} & \textbf{1.0000 ± 0.0000} & 0.9995 ± 0.0003 \\
\cline{1-9}
\multirow[t]{8}{*}{1} & DANN & 0.7053 ± 0.0080 & 0.7331 ± 0.0043 & 0.7586 ± 0.0111 & 0.7812 ± 0.0048 & 0.7699 ± 0.0070 & 0.7782 ± 0.0044 & 0.7544 ± 0.0066 \\
 & DIWA & \textbf{0.7692 ± 0.0091} & \textbf{0.7617 ± 0.0061} & \textbf{0.7842 ± 0.0085} & \textbf{0.7872 ± 0.0061} & \textbf{0.7857 ± 0.0036} & \textbf{0.8000 ± 0.0036} & \textbf{0.7813 ± 0.0062} \\
 & ERM & 0.7391 ± 0.0044 & 0.7474 ± 0.0087 & 0.7647 ± 0.0065 & 0.7850 ± 0.0018 & 0.7797 ± 0.0055 & 0.7699 ± 0.0069 & 0.7643 ± 0.0056 \\
 & NUC-0.01 & 0.7511 ± 0.0061 & 0.7504 ± 0.0079 & 0.7835 ± 0.0073 & 0.7729 ± 0.0082 & 0.7797 ± 0.0059 & 0.7977 ± 0.0048 & 0.7726 ± 0.0067 \\
 & NUC-0.1 & 0.7496 ± 0.0031 & 0.7436 ± 0.0036 & 0.7752 ± 0.0093 & 0.7850 ± 0.0084 & 0.7812 ± 0.0077 & 0.7782 ± 0.0043 & 0.7688 ± 0.0061 \\
 & PN-Y & 0.7481 ± 0.0082 & 0.7541 ± 0.0057 & 0.7564 ± 0.0047 & 0.7805 ± 0.0035 & 0.7699 ± 0.0038 & 0.7759 ± 0.0081 & 0.7642 ± 0.0057 \\
 & PN-Y-S & 0.7226 ± 0.0083 & 0.7421 ± 0.0149 & 0.7556 ± 0.0063 & 0.7489 ± 0.0090 & 0.7504 ± 0.0042 & 0.7692 ± 0.0068 & 0.7481 ± 0.0083 \\
 & PN-Y-S-E & 0.7331 ± 0.0145 & 0.7083 ± 0.0051 & 0.7346 ± 0.0064 & 0.7534 ± 0.0059 & 0.7609 ± 0.0076 & 0.7647 ± 0.0107 & 0.7425 ± 0.0083 \\
\cline{1-9}
\multirow[t]{8}{*}{2} & DANN & 0.7476 ± 0.0078 & 0.7829 ± 0.0081 & 0.7994 ± 0.0066 & 0.8067 ± 0.0077 & 0.8238 ± 0.0059 & 0.8274 ± 0.0100 & 0.7980 ± 0.0077 \\
 & DIWA & 0.7835 ± 0.0065 & 0.8110 ± 0.0050 & \textbf{0.8262 ± 0.0066} & 0.8335 ± 0.0048 & 0.8372 ± 0.0079 & 0.8378 ± 0.0054 & 0.8215 ± 0.0060 \\
 & ERM & 0.7811 ± 0.0045 & 0.7933 ± 0.0054 & 0.8128 ± 0.0039 & 0.8110 ± 0.0062 & 0.8122 ± 0.0073 & 0.8323 ± 0.0067 & 0.8071 ± 0.0057 \\
 & NUC-0.01 & 0.7567 ± 0.0072 & 0.7884 ± 0.0080 & 0.8134 ± 0.0092 & 0.8262 ± 0.0039 & \textbf{0.8409 ± 0.0060} & 0.8256 ± 0.0066 & 0.8085 ± 0.0068 \\
 & NUC-0.1 & 0.7622 ± 0.0026 & 0.7841 ± 0.0077 & 0.8134 ± 0.0083 & 0.8244 ± 0.0080 & 0.8256 ± 0.0050 & 0.8171 ± 0.0064 & 0.8045 ± 0.0063 \\
 & PN-Y & \textbf{0.8055 ± 0.0077} & \textbf{0.8226 ± 0.0039} & 0.8244 ± 0.0027 & \textbf{0.8396 ± 0.0071} & 0.8366 ± 0.0055 & 0.8329 ± 0.0024 & \textbf{0.8269 ± 0.0049} \\
 & PN-Y-S & 0.8018 ± 0.0055 & 0.8165 ± 0.0046 & 0.8091 ± 0.0055 & 0.8250 ± 0.0028 & 0.8274 ± 0.0021 & \textbf{0.8433 ± 0.0052} & 0.8205 ± 0.0043 \\
 & PN-Y-S-E & 0.7890 ± 0.0095 & 0.8061 ± 0.0061 & 0.8220 ± 0.0064 & 0.8293 ± 0.0054 & 0.8323 ± 0.0033 & 0.8366 ± 0.0053 & 0.8192 ± 0.0060 \\
\cline{1-9}
\multirow[t]{8}{*}{3} & DANN & 0.7811 ± 0.0070 & 0.8000 ± 0.0033 & 0.8178 ± 0.0072 & 0.8373 ± 0.0032 & 0.8396 ± 0.0049 & 0.8432 ± 0.0057 & 0.8198 ± 0.0052 \\
 & DIWA & 0.8556 ± 0.0030 & 0.8598 ± 0.0035 & 0.8627 ± 0.0040 & 0.8704 ± 0.0046 & 0.8734 ± 0.0029 & 0.8686 ± 0.0062 & 0.8651 ± 0.0040 \\
 & ERM & 0.8698 ± 0.0047 & 0.8586 ± 0.0068 & 0.8757 ± 0.0055 & 0.8746 ± 0.0054 & 0.8882 ± 0.0011 & 0.8799 ± 0.0057 & 0.8745 ± 0.0049 \\
 & NUC-0.01 & 0.8432 ± 0.0032 & 0.8598 ± 0.0042 & 0.8675 ± 0.0057 & 0.8704 ± 0.0052 & 0.8805 ± 0.0050 & 0.8846 ± 0.0034 & 0.8677 ± 0.0045 \\
 & NUC-0.1 & 0.8414 ± 0.0035 & 0.8592 ± 0.0035 & 0.8627 ± 0.0035 & 0.8669 ± 0.0026 & 0.8751 ± 0.0025 & 0.8746 ± 0.0080 & 0.8633 ± 0.0039 \\
 & PN-Y & 0.8734 ± 0.0030 & 0.8793 ± 0.0060 & 0.8882 ± 0.0047 & \textbf{0.8953 ± 0.0038} & 0.8941 ± 0.0050 & \textbf{0.9012 ± 0.0043} & 0.8886 ± 0.0045 \\
 & PN-Y-S & \textbf{0.8781 ± 0.0036} & 0.8811 ± 0.0024 & \textbf{0.8893 ± 0.0033} & 0.8905 ± 0.0026 & \textbf{0.9041 ± 0.0018} & 0.8964 ± 0.0036 & \textbf{0.8899 ± 0.0029} \\
 & PN-Y-S-E & 0.8722 ± 0.0059 & \textbf{0.8899 ± 0.0047} & 0.8846 ± 0.0050 & 0.8893 ± 0.0020 & 0.8964 ± 0.0030 & 0.8935 ± 0.0028 & 0.8877 ± 0.0039 \\
\cline{1-9}
\bottomrule
\end{tabular}

}
\caption{Target Finetuning results on VLCS dataset.}\label{tab:tf_vlcs}
\end{table}

\begin{table}[!htbp]
    \small
\scalebox{0.9}{
    \begin{tabular}{lllllllll}
\toprule
 &  & 0.10 & 0.20 & 0.40 & 0.60 & 0.80 & 1.00 & mean \\
$E_{t}$ & Method &  &  &  &  &  &  &  \\
\midrule
\multirow[t]{8}{*}{0} & DANN & 0.8076 ± 0.0082 & 0.8650 ± 0.0055 & 0.9169 ± 0.0041 & 0.9549 ± 0.0023 & 0.9675 ± 0.0025 & 0.9764 ± 0.0012 & 0.9147 ± 0.0040 \\
 & DIWA & 0.8793 ± 0.0043 & 0.8979 ± 0.0040 & 0.9401 ± 0.0045 & 0.9570 ± 0.0022 & \textbf{0.9726 ± 0.0013} & 0.9726 ± 0.0042 & 0.9366 ± 0.0034 \\
 & ERM & 0.8696 ± 0.0055 & 0.8937 ± 0.0014 & 0.9409 ± 0.0043 & 0.9515 ± 0.0038 & 0.9671 ± 0.0035 & \textbf{0.9806 ± 0.0022} & 0.9339 ± 0.0035 \\
 & NUC-0.01 & 0.7793 ± 0.0040 & 0.8426 ± 0.0074 & 0.9143 ± 0.0020 & 0.9460 ± 0.0045 & 0.9591 ± 0.0032 & 0.9700 ± 0.0023 & 0.9019 ± 0.0039 \\
 & NUC-0.1 & 0.7772 ± 0.0031 & 0.8494 ± 0.0028 & 0.9177 ± 0.0030 & 0.9464 ± 0.0046 & 0.9553 ± 0.0031 & 0.9709 ± 0.0016 & 0.9028 ± 0.0030 \\
 & PN-Y & \textbf{0.8835 ± 0.0046} & \textbf{0.9093 ± 0.0030} & 0.9414 ± 0.0049 & \textbf{0.9608 ± 0.0036} & 0.9679 ± 0.0026 & 0.9751 ± 0.0016 & \textbf{0.9397 ± 0.0034} \\
 & PN-Y-S & 0.8772 ± 0.0045 & 0.9051 ± 0.0037 & 0.9376 ± 0.0017 & 0.9553 ± 0.0020 & 0.9658 ± 0.0017 & 0.9730 ± 0.0014 & 0.9357 ± 0.0025 \\
 & PN-Y-S-E & 0.8787 ± 0.0020 & 0.9013 ± 0.0023 & \textbf{0.9430 ± 0.0024} & 0.9527 ± 0.0033 & 0.9679 ± 0.0039 & 0.9759 ± 0.0027 & 0.9366 ± 0.0028 \\
\cline{1-9}
\multirow[t]{8}{*}{1} & DANN & 0.8361 ± 0.0061 & 0.8949 ± 0.0032 & 0.9329 ± 0.0041 & 0.9556 ± 0.0020 & 0.9649 ± 0.0014 & 0.9747 ± 0.0008 & 0.9265 ± 0.0029 \\
 & DIWA & \textbf{0.8797 ± 0.0051} & \textbf{0.9203 ± 0.0054} & \textbf{0.9446 ± 0.0031} & \textbf{0.9575 ± 0.0051} & \textbf{0.9731 ± 0.0010} & \textbf{0.9764 ± 0.0009} & \textbf{0.9419 ± 0.0034} \\
 & ERM & 0.8333 ± 0.0071 & 0.8947 ± 0.0045 & 0.9331 ± 0.0035 & 0.9515 ± 0.0027 & 0.9653 ± 0.0028 & 0.9749 ± 0.0020 & 0.9255 ± 0.0038 \\
 & NUC-0.01 & 0.8070 ± 0.0062 & 0.8793 ± 0.0031 & 0.9265 ± 0.0042 & 0.9466 ± 0.0022 & 0.9620 ± 0.0020 & 0.9698 ± 0.0014 & 0.9152 ± 0.0032 \\
 & NUC-0.1 & 0.8154 ± 0.0032 & 0.8723 ± 0.0039 & 0.9331 ± 0.0036 & 0.9458 ± 0.0028 & 0.9624 ± 0.0015 & 0.9713 ± 0.0006 & 0.9167 ± 0.0026 \\
 & PN-Y & 0.8591 ± 0.0037 & 0.9021 ± 0.0038 & 0.9312 ± 0.0023 & 0.9487 ± 0.0011 & 0.9620 ± 0.0012 & 0.9745 ± 0.0018 & 0.9296 ± 0.0023 \\
 & PN-Y-S & 0.8472 ± 0.0050 & 0.8943 ± 0.0028 & 0.9370 ± 0.0025 & 0.9520 ± 0.0014 & 0.9651 ± 0.0020 & 0.9749 ± 0.0003 & 0.9284 ± 0.0023 \\
 & PN-Y-S-E & 0.8509 ± 0.0030 & 0.9004 ± 0.0035 & 0.9312 ± 0.0027 & 0.9472 ± 0.0025 & 0.9645 ± 0.0012 & 0.9731 ± 0.0020 & 0.9279 ± 0.0025 \\
\cline{1-9}
\multirow[t]{8}{*}{2} & DANN & 0.6816 ± 0.0133 & 0.7673 ± 0.0072 & 0.8529 ± 0.0056 & 0.8922 ± 0.0072 & 0.9179 ± 0.0061 & 0.9441 ± 0.0042 & 0.8427 ± 0.0073 \\
 & DIWA & \textbf{0.7778 ± 0.0104} & \textbf{0.8292 ± 0.0056} & \textbf{0.8826 ± 0.0065} & \textbf{0.9128 ± 0.0043} & \textbf{0.9370 ± 0.0052} & \textbf{0.9471 ± 0.0029} & \textbf{0.8811 ± 0.0058} \\
 & ERM & 0.7526 ± 0.0056 & 0.8096 ± 0.0052 & 0.8579 ± 0.0062 & 0.8997 ± 0.0062 & 0.9179 ± 0.0045 & 0.9194 ± 0.0035 & 0.8595 ± 0.0052 \\
 & NUC-0.01 & 0.6222 ± 0.0102 & 0.7375 ± 0.0126 & 0.8489 ± 0.0054 & 0.8977 ± 0.0058 & 0.9224 ± 0.0033 & 0.9390 ± 0.0043 & 0.8280 ± 0.0069 \\
 & NUC-0.1 & 0.6121 ± 0.0090 & 0.7365 ± 0.0083 & 0.8489 ± 0.0065 & 0.8987 ± 0.0019 & 0.9249 ± 0.0037 & 0.9395 ± 0.0055 & 0.8268 ± 0.0058 \\
 & PN-Y & 0.7647 ± 0.0045 & 0.8227 ± 0.0063 & 0.8665 ± 0.0052 & 0.8922 ± 0.0041 & 0.9123 ± 0.0043 & 0.9295 ± 0.0035 & 0.8647 ± 0.0047 \\
 & PN-Y-S & 0.7446 ± 0.0049 & 0.8050 ± 0.0031 & 0.8584 ± 0.0074 & 0.8856 ± 0.0049 & 0.9098 ± 0.0041 & 0.9285 ± 0.0028 & 0.8553 ± 0.0046 \\
 & PN-Y-S-E & 0.7446 ± 0.0031 & 0.7953 ± 0.0049 & 0.8559 ± 0.0042 & 0.8872 ± 0.0033 & 0.9159 ± 0.0048 & 0.9300 ± 0.0045 & 0.8548 ± 0.0041 \\
\cline{1-9}
\multirow[t]{8}{*}{3} & DANN & 0.6983 ± 0.0047 & 0.7969 ± 0.0062 & 0.8714 ± 0.0037 & 0.9034 ± 0.0022 & \textbf{0.9296 ± 0.0045} & 0.9364 ± 0.0019 & 0.8560 ± 0.0039 \\
 & DIWA & \textbf{0.7806 ± 0.0067} & \textbf{0.8289 ± 0.0090} & \textbf{0.8806 ± 0.0041} & \textbf{0.9065 ± 0.0050} & 0.9259 ± 0.0039 & 0.9276 ± 0.0022 & \textbf{0.8750 ± 0.0052} \\
 & ERM & 0.7534 ± 0.0062 & 0.8207 ± 0.0056 & 0.8704 ± 0.0047 & 0.8980 ± 0.0063 & 0.9231 ± 0.0027 & 0.9320 ± 0.0028 & 0.8663 ± 0.0047 \\
 & NUC-0.01 & 0.6476 ± 0.0060 & 0.7769 ± 0.0117 & 0.8571 ± 0.0085 & 0.8980 ± 0.0058 & 0.9184 ± 0.0042 & 0.9361 ± 0.0022 & 0.8390 ± 0.0064 \\
 & NUC-0.1 & 0.6374 ± 0.0029 & 0.7718 ± 0.0109 & 0.8585 ± 0.0083 & 0.8993 ± 0.0053 & 0.9214 ± 0.0029 & 0.9374 ± 0.0014 & 0.8376 ± 0.0053 \\
 & PN-Y & 0.7673 ± 0.0037 & 0.8289 ± 0.0077 & 0.8721 ± 0.0088 & 0.8952 ± 0.0054 & 0.9252 ± 0.0032 & 0.9384 ± 0.0035 & 0.8712 ± 0.0054 \\
 & PN-Y-S & 0.7486 ± 0.0041 & 0.8167 ± 0.0094 & 0.8714 ± 0.0057 & 0.8993 ± 0.0047 & 0.9241 ± 0.0020 & 0.9374 ± 0.0021 & 0.8663 ± 0.0047 \\
 & PN-Y-S-E & 0.7524 ± 0.0066 & 0.8185 ± 0.0070 & 0.8673 ± 0.0059 & 0.8973 ± 0.0049 & 0.9238 ± 0.0012 & \textbf{0.9408 ± 0.0022} & 0.8667 ± 0.0046 \\
\cline{1-9}
\bottomrule
\end{tabular}

}
\caption{Target Finetuning results on TerrainCognita dataset.}\label{tab:tf_terraincognita}
\end{table}

    

\clearpage

\subsection{Learning Rate Decay Results}

\begin{figure}[H]
    \small
    \begin{center}
    \scalebox{0.9}{
        \includegraphics{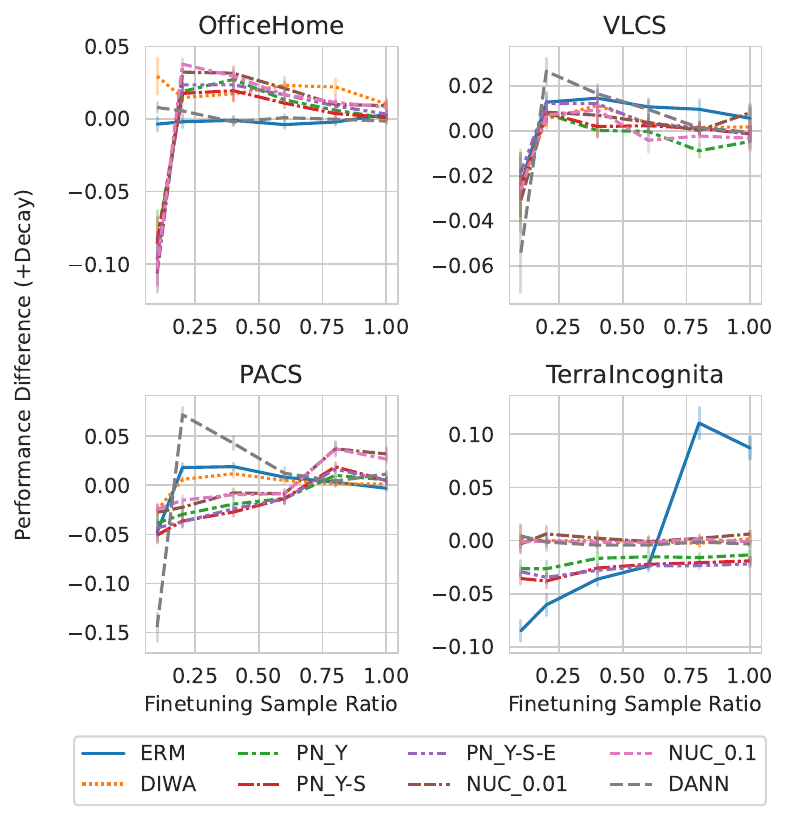}
    }
    \end{center}
    
    \caption{Feature Encoder Learning Rate Decay Results: We show that learning rate decay does not may improve the performance of the model on large FSL regimes but improves the performance on small FSL regimes.}\label{fig:lr_decay}
\end{figure}

We test the learning rate decay for the feature encoder, which is commonly used in domain adaptation methods. We use the same setting as the target finetuning experiment, but we decay the learning rate by 0.1. We report the results in \autoref{fig:lr_decay}.

\newpage
\section{Experiment Details}

Our code is available \href{https://github.com/Xiang-Pan/ProjectionNet}{https://github.com/Xiang-Pan/ProjectionNet}.

\subsection{Linear synthetic data}\label{subsec:linear_synthetic}

We have $E=999$ source environments and a target environment indexed by $1000$.  The source and target sample sizes are $n_1=50$, $n_2=2000$ respectively, with the ambient input dimension $d=10$, content feature dimension $k=6$, and variance of noise $\sigma=0.01$.

\paragraph{Generation of $X^{e}$} 
For $e\in[E]$, $X^{e} \in \reals^{n_1 \times d}$ and for $e = E+1$, $X^{E+1} \in \reals^{n_2 \times d}$. To guarantee Assumption \ref{assumption: trace assumption}, we generate $X^e$ via its SVD: $X^e = U^{e} S^{e} V^{e\top}$ where $U^{e}$ and $V^{e}$ are random orthogonal matrices, and the singular values $S^{e}$ are sampled from the mixture of Gaussian distributions. Specifically, the top $k$ singular values follow $\mathcal{N}(5, 1)$ and the rest $d-k$ follow $\mathcal{N}(0, 1)$.

\paragraph{Generation of $\btheta^{*e}$}
For $e \in [E+1]$, $\btheta^{*e}$'s are i.i.d. samples of the meta distribution (\eqref{eqn: meta distribution}) with the parameter choice $\btheta^{*} = 6\cdot \mathbf{1}_{k}, \Lambda_{11}= 0.1 \cdot I_{k}
, \Lambda_{22}= 3 \cdot I_{d-k}$.

\paragraph{Generation of $\by^{e}$} 
For $e\in [E+1]$, $\by^{e}$'s are generated via 
\[
\by^{e} = X^{e}R^{*}\btheta^{*e} + \bz^e
\] where $R^*$ is some random $d \times d$ orthogonal matrix and $\bz^{e}$'s are i.i.d. samples of $\mathcal{N}(0,\sigma^2 I_{n_1/n_2})$ independent of $X^{e}$'s.

\begin{figure}[h!]
\subfloat[Effect of $\lambda_1$]{\includegraphics[width=0.5\columnwidth]{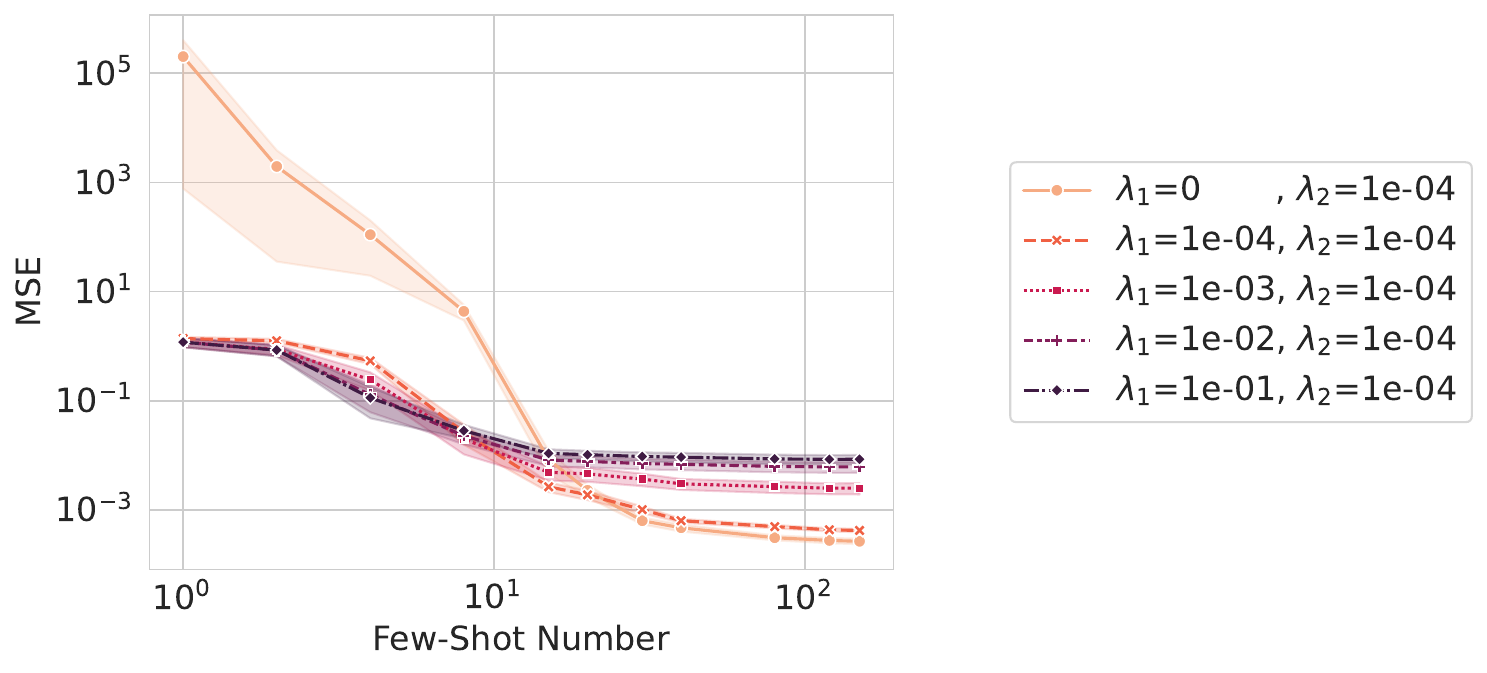}}
\hfill
\subfloat[Effect of $\lambda_2$]{\includegraphics[width=0.5\columnwidth]{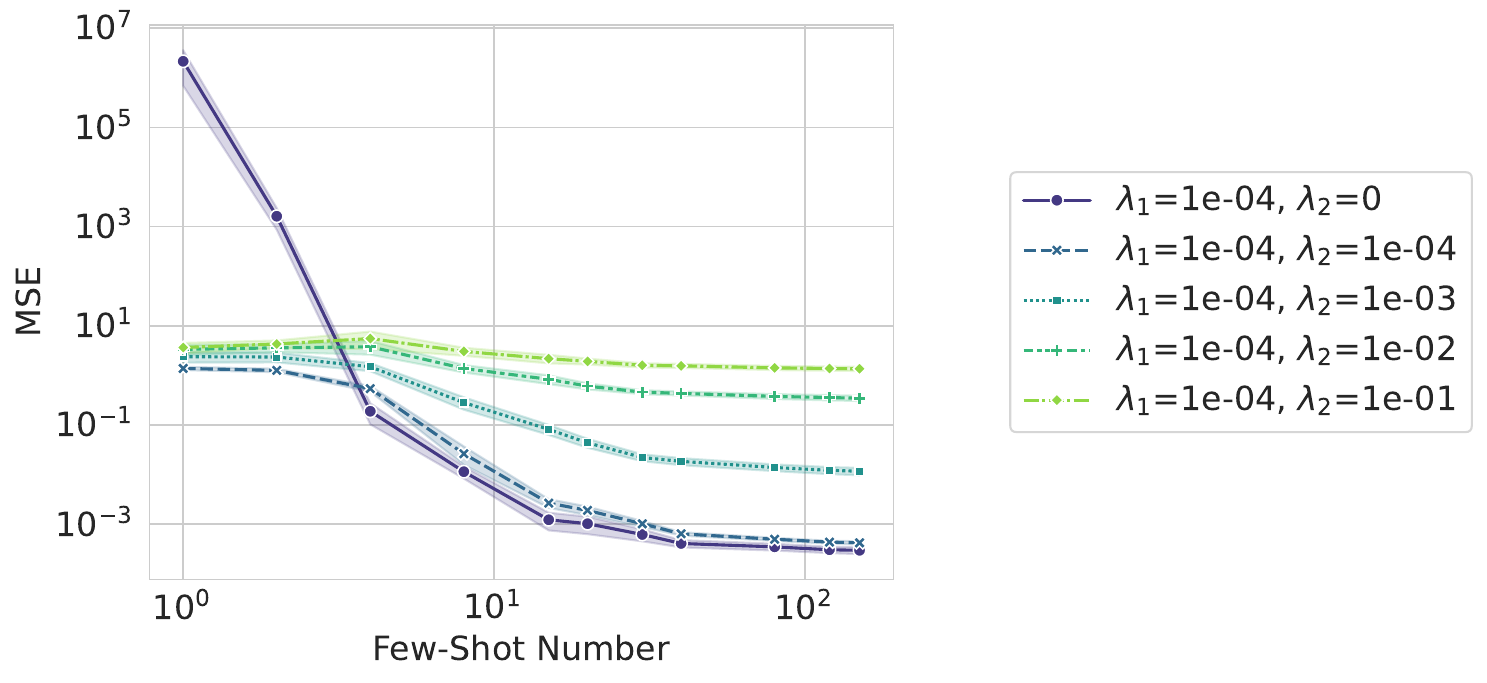}}
\caption{The effect of $(\lambda_1, \lambda_2)$ and on the performance of the target task: No regularization leads to an unstable solution and overfitting when FSL is small, but large regularization prevent the full adaptation to the target task when FSL is large.}
\label{fig:linear_lambda2}
\end{figure}


\subsection{Real data}
\subsubsection{Training Configuration}

We use ResNet50~\cite{he2016deep} as the backbone network and use the SGD optimizer to train the model, the training setting is the same as~\citep{deepdg}.

We use $\text{lr}_{\text{decay}} = 0.1, \text{lr}_\text{Feature}=\text{lr}_\text{decay} * \text{lr}_\text{Classification}$. 
We vary the learning rate decay from $0$ (frozen) to $1$ (not decayed) to control the ($\lambda_1$ and $\lambda_2$) regularization strength. In the source training stage, we use the source validation set to select the best model; In the target training stage, we use the target validation set to select the best model. 

\textbf{Computation Resource}: All the experiments can be done with A40, RTX 8000 GPU or A100 GPU, 32GB memory, and 16 CPU 2.9GHz cores (Intel Cascade Lake Platinum 8268 chips).


\end{document}